\documentclass[journal]{IEEEtran}
\usepackage{amsmath,amsfonts,amssymb,epsfig}
\usepackage{multirow,hhline,color,booktabs}
\usepackage{xcolor,soul}
\usepackage{enumitem}
\usepackage{graphicx}
\usepackage[export]{adjustbox}
\usepackage{comment}
\usepackage{url}
\usepackage{tikz}
\usepackage[percent]{overpic}
\usepackage{algorithm}
\usepackage{algpseudocode}
\usepackage{setspace}

\newcommand{\ru} {\rule{0mm}{3mm}}
\newcommand{\red}[1]{{\color{red} #1}}

\newcommand{\DL} {D_{\lambda}}
\newcommand{\DR} {D_{\rho}}
\newcommand{\DS} {D_{S}}

\newcommand{\git}{{\tt {\url{https://github.com/giu-guarino/rho-PNN}}}}
\begin{document}
\title{Zero-Shot Hyperspectral Pansharpening Using Hysteresis-Based Tuning for Spectral Quality Control}

\author{Giuseppe~Guarino,~\IEEEmembership{Student~Member,~IEEE,}
		Matteo~Ciotola,~\IEEEmembership{Member,~IEEE,}		
        Gemine~Vivone,~\IEEEmembership{Senior~Member,~IEEE,}
        Giovanni~Poggi,~\IEEEmembership{Member,~IEEE,}
        and~Giuseppe~Scarpa,~\IEEEmembership{Senior~Member,~IEEE}

\thanks{This work was supported by the Italian Space Agency (ASI) under Grant ``Space It Up!'', Spoke3, CUP E63C24000220006
and by the national recovery and resilience plan (NRRP), Mission 4 Component 2 Investment 1.4 - call for tender no. 3138 of 16 December 2021, rectified by Decree n. 3175 of 18 December 2021 of Italian Ministry of University and Research funded by the European Union – NextGenerationEU; Project code CN\_00000033, Concession Decree no. 1034 of 17 June 2022 adopted by the Italian Ministry of University and Research, codice unico di progetto (CUP) B83C22002930006, Project title ``National Biodiversity Future Center - NBFC".}
%Master I53D24000060005.
\thanks{Giuseppe Guarino, Matteo Ciotola and Giovanni Poggi are with the Department of Electrical Engineering and Information Technology, University Federico II, 80125 Napoli, Italy
(e-mail: giuseppe.guarino2@unina.it, matteo.ciotola@unina.it and poggi@unina.it).}
\thanks{Gemine Vivone is with the National Research Council, Institute of Methodologies for Environmental Analysis, CNR-IMAA, 85050 Tito, Italy, and also with the National Biodiversity Future Center (NBFC), 90133 Palermo, Italy
(e-mail: gemine.vivone@imaa.cnr.it).}
\thanks{Giuseppe Scarpa is with the Department of Engineering, University Parthenope, 80143 Napoli, Italy
(e-mail: giuseppe.scarpa@uniparthenope.it).}
}

\maketitle

\begin{abstract}
Hyperspectral pansharpening has received much attention in recent years due to technological and methodological advances that open the door to new application scenarios.
However, research on this topic is only now gaining momentum.
The most popular methods are still borrowed from the more mature field of multispectral pansharpening and often overlook the unique challenges posed by hyperspectral data fusion, such as
{\it   i)} the very large number of bands,
{\it  ii)} the overwhelming noise in selected spectral ranges,
{\it iii)} the significant spectral mismatch between panchromatic and hyperspectral components,
{\it  iv)} a typically high resolution ratio.
Imprecise data modeling especially affects spectral fidelity.
Even state-of-the-art methods perform well in certain spectral ranges and much worse in others, failing to ensure consistent quality across all bands, with the risk of generating unreliable results.
Here, we propose a hyperspectral pansharpening method that explicitly addresses this problem and ensures uniform spectral quality.
To this end, a single lightweight neural network is used, with weights that adapt on the fly to each band.
During fine-tuning, the spatial loss is turned on and off to ensure a fast convergence of the spectral loss to the desired level, according to a hysteresis-like dynamic.
Furthermore, the spatial loss itself is appropriately redefined to account for nonlinear dependencies between panchromatic and spectral bands.
Overall, the proposed method is fully unsupervised, with no prior training on external data, flexible, and low-complexity.
Experiments on a recently published benchmarking toolbox show that it ensures excellent sharpening quality, competitive with the state-of-the-art, consistently across all bands.
The software code and the full set of results are shared online on \git.

%\begin{IEEEkeywords}
%Hyperspectral images, image fusion, optical imaging, resolution enhancement, super-resolution.
%\end{IEEEkeywords}

\end{abstract}


\IEEEpeerreviewmaketitle

\section{Introduction}
\label{sec:intro}

Hyperspectral (HS) remote sensing imagery is a key resource for monitoring and managing such diverse environments as
forests \cite{Goodenough2003, Dalponte2013},
agricultural areas \cite{Jia2022, Reddy2024},
oceans \cite{Dumke2019, Kang2023},
coastal areas \cite{Brando2003, Gao2022},
inland waters and wetlands \cite{Hestir2015, Su2023},
snow and ice \cite{Nolin2000, Han2021},
atmosphere \cite{Di2022, Wang2024},
mineral resources \cite{Siebels2020, Lorenz2021,  Simoes2014},
urban areas \cite{Roessner2001, Huang2008}.
Due to the technological constraints of the sensors and the very high number of spectral bands acquired,
HS images have a rather low spatial resolution.
A cost-effective solution to overcome this limitation is to equip the flight system with an additional camera to acquire,
in synchrony with the HS image, a single high-resolution panchromatic (PAN) band.
Then, using a fusion process known as pansharpening
the PAN-HS pair is merged into a single high-resolution HS datacube.
This data fusion process improves spatial resolution while preserving spectral coherence,
thereby boosting the performance of downstream tasks such as classification \cite{Wu2024} and land use/cover mapping \cite{Lin2015}.
%Algorithms designed to improve the spatial resolution of remotely sensed images have already been shown to significantly boost performance in downstream tasks such as classification \cite{Wu2024} and land use/land cover mapping \cite{Lin2015}. It is therefore reasonable to expect that enhanced HS data could offer substantial benefits in applications where HS information is crucial but often underexploited due to limited spatial resolution, for example, in air pollution estimation \cite{Mazza2025} or water quality monitoring \cite{Goyens2022}. 
Examples of satellite-borne missions that feature the joint acquisition of PAN and HS images are
PRISMA ({\em PRecursore IperSpettrale della Missione Applicativa}) by  the Italian Space Agency (ASI), and
EO-1/ALI (Earth Observing-1/Advanced Land Imaging) by the U.S. National Aeronautics and Space Administration (NASA).
In both cases, the spatial resolution is 30m for the HS component and much better for the PAN image, 5m for PRISMA and 10m for EO-1/ALI.
Needless to say, the PRISMA case is certainly more interesting, as it promises to provide high-resolution data,
but also more challenging since it requires the pansharpening algorithm to handle a resolution ratio of 6
between the available HS data in input and the desired output.

Pansharpening is not a new topic.
It has been the object of intense research in the last decades
with reference to multispectral (MS) images,
where only a limited number of spectral bands (from 4 to 8) are available in the visible to near-infrared spectrum.
A large number of methods have been proposed for MS pansharpening \cite{Vivone2020, Deng2022}, with very different approaches such as
component substitution \cite{Aiazzi2007,Garzelli2008,Choi2011,Lolli2017},
multiresolution analysis \cite{Otazu2005,Aiazzi2006, Restaino2016},
variational optimization \cite{Palsson2014, Vivone2015, Palsson2020}, \cite{Yu2021}
and machine/deep learning \cite{Masi2016,Yang2017,Scarpa2018,Ciotola2023a,Ciotola2024a}.
Only recently the attention is shifting towards the HS case,
as testified by the growing number of research papers (see \cite{Ciotola2024} for a review) and by open challenges such as \cite{Vivone2023}.

Clearly, MS and HS pansharpening share a common formulation
which may suggest that PAN-MS fusion methods can be readily extended to operate also in the PAN-HS case.
Indeed,
early methods for HS pansharpening \cite{Capo07} were simple adaptations from the MS case, based on
Bayesian models \cite{Zhang2009, Simoes2015, Wei2015, Wei2015a},
matrix factorization \cite{Berne2010, Moel09, Huang2013},
variational optimization \cite{Kawakami2011},
component substitution and multiresolution analysis \cite{Vivone2014b}
(see \cite{Loncan2015} for a detailed review).
However, this approach is overly simplistic,
since the HS case has numerous peculiar features that make it particularly challenging:
\begin{itemize}[noitemsep,leftmargin=12pt]
\item   PAN and MS images span comparable spectral ranges, and the PAN bandwidth covers most MS bands.
        The HS image, instead, has a much larger spectral extension and most of the HS bands are not covered by the PAN.
        Therefore, their spatial structure cannot be reliably predicted from it.
\item   HS bands are much narrower than MS ones, in the order of 10nm.
        This is why, to harvest more energy, larger ground cells are necessary.
        Even so, the acquired data are rather noisy, with an extremely variable quality,
        and some groups of bands, in spectral ranges where the sunlight is weaker, must be discarded altogether.
\item   Sometimes, the PAN-MS resolution ratio is rather large (e.g., six for PRISMA),
        hence the pansharpening algorithm is called upon to recover a large amount of missing information.
\item   HS images are huge, hence the computational complexity is a serious issue, whatever the optimization strategy adopted.
        This is especially troublesome for supervised machine learning methods that require large training sets covering the whole data space.
\end{itemize}

\vspace{3mm} \noindent
In latest years, a few model-based methods specifically designed for HS pansharpening were proposed, relying on
guided filtering \cite{Qu17},
variational optimization \cite{Adde17,Huan17},
saliency-based component substitution \cite{Dong20}.
However, the deep learning (DL) revolution has by now spread to the field of remote sensing \cite{Zhu2017} and,
since the seminal 2016 work of Masi {\it et al.} \cite{Masi2016}, deep learning is a {\em de facto} standard for pansharpening. %\cite{Yang2017, Deng2022, Ciotola2023a} \cite{He2019a, heng2020,Bandara2022,Guarino2023}
Focusing on the HS case,
in \cite{He2019a} and \cite{He2020} dedicated convolutional neural networks (CNNs) were designed to strengthen the spectral prediction capability.
In \cite{Zheng2020} a dual-attention residual network was proposed, with a deep HS image prior module for HS upsampling.
In \cite{He2022, He2022a} the problem of scale and resolution variability was tackled by means of a CNN with arbitrary-scale attention modules \cite{He2022a}.
An overcomplete residual network that learns high-level features using constrained receptive fields was proposed in \cite{Bandara2022}.
Multibranch network architectures were also explored in several recent works \cite{Guan2022, Wu22, Qu22a}.
Other methods adopted classical pansharpening paradigms and used DL modules to estimate the model parameters \cite{Dong22} or to process the extracted features \cite{Dong22a}.
All these methods rely on supervised learning and obviate the absence of a real ground truth (GT) by training on reduced-resolution simulated images.
The performance is indeed very good on reduced-resolution test data but drops significantly when moving to the real target of the process, the full-resolution images.
In other words, models trained in the subsampled domain do not generalize well to the full-resolution domain.
Recently, to overcome these problems, a paradigm shift towards full-resolution training is taking place
for both MS \cite{Seo2020, Ciotola2022, Ciotola2023} and HS \cite{Nie22,Guarino2023, Guarino2023a} pansharpening,
with novel learning strategies and suitably defined loss functions.
Focusing on the HS case,
the RE-RANet method proposed in \cite{Nie22}
leverages the ratio image as latent variable and derives spatial and spectral information of the fused image from the full-resolution PAN and HS components.
The R-PNN model \cite{Guarino2023} performs pansharpening sequentially for each individual band using an unsupervised target-adaptive solution.
A divide-and-conquer strategy is used by PCA-Z-PNN, where the HS stack is split in two coherent sub-stacks which are transformed by principal component analysis and pansharpened individually by the MS-oriented Z-PNN \cite{Ciotola2022} model.
A common trait of these three methods is the use of two-term losses that suitably balance spectral and spatial quality.

Aiming to provide an objective overview of recent progress in the field and to support further research,
we have recently developed a benchmarking toolbox\footnote{\url{https://github.com/matciotola/hyperspectral_pansharpening_toolbox}} for HS pansharpening \cite{Ciotola2024}.
A large number of state-of-the-art (SotA) methods, both model-based and data-driven, have been reimplemented using a common development environment
and evaluated under uniform conditions on a large and representative set of publicly available images.
In \cite{Ciotola2024} we performed a comparative analysis of the state of the art.
However, the toolbox can be used to perform many more experiments and investigate specific aspects of pansharpening,
such as the strengths and weaknesses of different approaches.
In particular, here we focus on the issue of non-uniform quality along the spectral dimension, a problem largely neglected in the literature.
Indeed, we observed that nearly all methods perform well in some spectral intervals and much worse in others, providing unreliable results in some spectral bands.
Of course, spectral richness is a key distinctive feature of HS images, but it becomes meaningless in the absence of adequate guarantees on the accuracy of the observed data.
Only under these conditions, these images can be smoothly used by the end user.
It should be noted that a major cause of this undesirable behavior is the lack of a quality assessment protocol specifically tailored to the hyperspectral case.
The widely used performance metrics are simple adaptations of those designed for the very different MS case and, as such,
they fail to capture subtle but important differences.

Motivated by the above considerations, in this work we propose a novel DL-based HS pansharpening method
which is able to keep the spectral distortion under control in each single band, so as to ensure uniform (and top) quality along the whole spectral range.
To this end, we build upon the recently proposed \cite{Guarino2023} Rolling Pansharpening Neural Network (Rolling PNN or R-PNN) method whose structure perfectly fits our need.
R-PNN performs a sequential band-wise pansharpening using a single lightweight neural network which is adapted on the fly to the target band, with no prior training but for the first band.
This architecture allows us to address individually each band's needs.
First, the desired level of spectral distortion is estimated based on the band features.
Then a band-adaptive optimization schedule is enacted to reach this target level,
where the spatial loss is carefully balanced with the spectral loss and switched on and off according to a hysteresis-like schedule in order to avoid undesired local minima.
In addition, the spatial loss itself is re-defined to account for inverted PAN-HS correlations that can occur in specific spectral ranges.
Experimental results on the benchmarking toolbox \cite{Ciotola2024} prove the effectiveness of the proposed solution,
which is competitive with the SotA on real full-resolution images in terms of global indexes and much superior in terms of spectral quality consistency.

In summary, the proposed method inherits the appealing properties of R-PNN and adds further strengths:
\begin{itemize}[noitemsep,leftmargin=12pt]
\item[a.]   Fully unsupervised and zero-shot, with no need for prior training, no cross-dataset generalization issues.
\item[b.]   Highly flexible: the number of spectral bands does not need to be fixed.
\item[c.]   Uniform distortion along the whole spectral range, with target level defined by the user through a single and easily interpretable global parameter.
\item[d.]   SotA performance on real full-resolution images, especially in terms of spectral quality.
\item[e.]   Scalable complexity, which allows to trade-off computation for quality based on application needs.
\end{itemize}

In the rest of the paper,
Section~\ref{sec:background} provides some necessary background concepts,
Section~\ref{sec:method} describes the proposed solution,
Section~\ref{sec:results} presents and comments comparative experimental results,
and Section~\ref{sec:conclusions} draws conclusions and outlines future perspectives.

\section{Background}
\label{sec:background}

The solution proposed in this paper moves from our recent Rolling PNN method \cite{Guarino2023},
which we briefly summarize here, highlighting its distinctive features.
% with particular attention to spectral fidelity issues, which are of primary importance when dealing with HS imagery.
Then we analyze the issue of spectral fidelity for current state of the art methods.

\subsection{Rolling PNN}
\label{sec:related}

\begin{figure}
\centering
\includegraphics[width=0.32\textwidth]{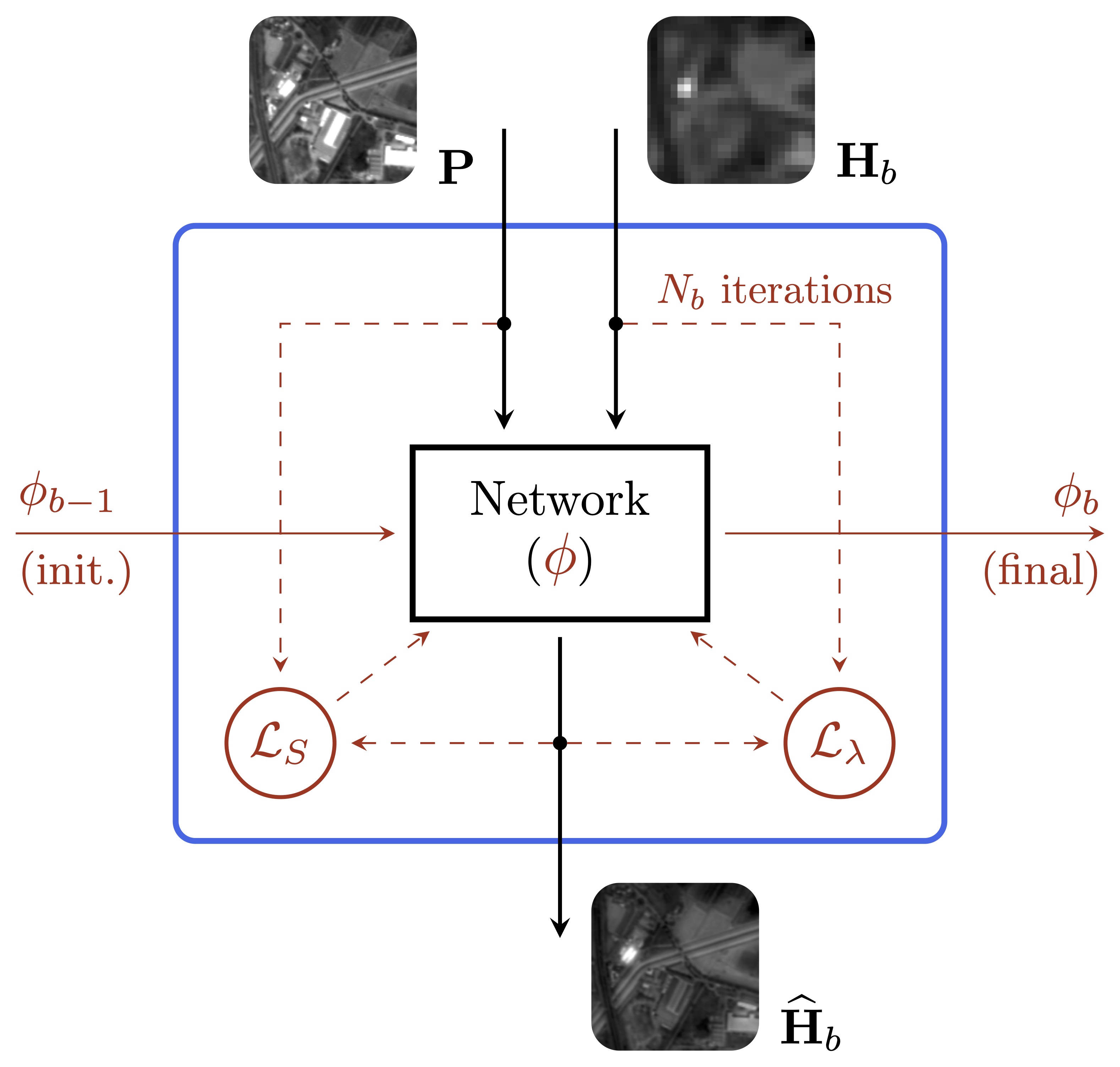}

(a)

\vspace{3mm}
\includegraphics[width=0.46\textwidth]{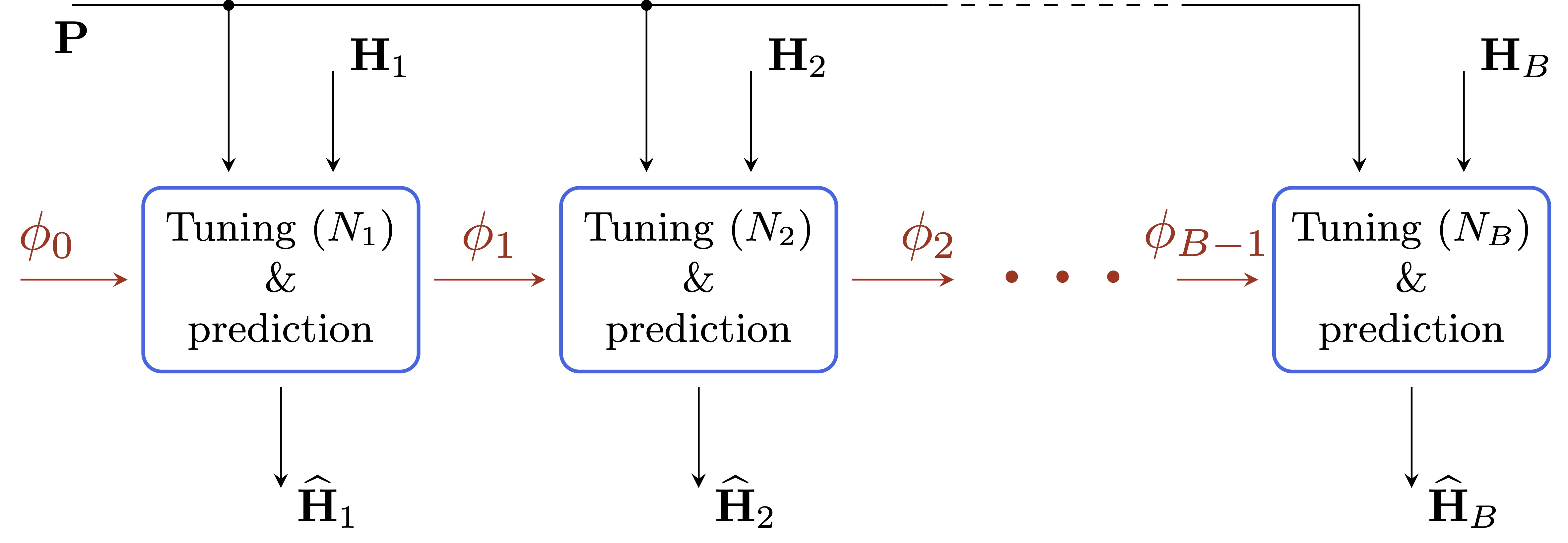}

(b)
\caption{R-PNN flowchart.
Each module (a) inherits the initial weights, $\phi_{b-1}$, from the previous module, fine-tunes them on the current band, and passes the updated version, $\phi_b$, to the next module, as depicted in (b).
The $b$-th band is pansharpened with its own tuned parameters $\phi_b$. $N_b$ is the number of tuning iterations assigned to band $b$.
}
\label{fig:rolling}
\end{figure}

\renewcommand{\H}{\mathbf{H}}
\newcommand{\X}{\mathbf{X}}
\newcommand{\Y}{\mathbf{Y}}
\renewcommand{\P}{\mathbf{P}}
\newcommand{\Q}{\mathbf{Q}}
\newcommand{\C}{\mathbf{I}}
\newcommand{\wt}[1]{\widetilde{\mathbf{#1}}}
\newcommand{\wh}[1]{\widehat{\mathbf{#1}}}
\def\L{{\cal L}}
\begin{table}
\caption{Main symbols}
\footnotesize
\centering
\setlength{\tabcolsep}{2pt}
\begin{tabular}{cll}%{p{2cm}p{3cm}p{1.5cm}}%{p{8.4cm}}
\hline
\textbf{\ru Symbol} & \textbf{Dimensions}    & \textbf{Meaning}\\\hline
%\red{HS} && Hyperspectral \\
%\red{PAN} & & Panchromatic \\
$w, h$                     & Scalars \ru                & HS width and height, respectively \\
$W, H$                     & Scalars                    & PAN width and height, respectively \\
$R$                        & Scalar                     & Resolution ratio ($R=H/h = W/w$) \\
$B$                        & Scalar                     & \# of HS bands \\
$\H$                       & $[w{\times}h{\times}B]$    & Original HS image \\
$\P$                       & $[W{\times}H]$             & Original PAN image \\
$\wt{H}$                   & $[W{\times}H{\times}B]$    & $R{\times}R$ upscaled version of $\H$ \\
$\wh{H}$                   & $[W{\times}H{\times}B]$    & Pansharpened version of $\H$ \\
$\mathbf{D}$               & $[W{\times}H{\times}B]$    & Estimated details from $\P$ and $\wt{H}$\\
$\X_b$                     & $[*{\times}*]$             & $b$-th band of $\X$, $[*{\times}*{\times}B]$\\
$\L_\lambda$, $\L_S$, $\L$ & Scalars                    & Spectral, spatial and overall loss functions \\
%$\wt{X}$ or $\wh{X}$ & $[W{\times}H{\times}\ast]$ & Upscaling or pansharpening of any $\X$ \\
\hline
\end{tabular}
\label{tab:symbols}
\end{table}

Compared to the more familiar case of pansharpening of MS images,
in the HS case the generalization of deep learning models is usually more challenging for several reasons:
less training data, richer spectral information, low correlation between PAN and spectral bands outside the visible spectrum,
larger resolution ratios,
and variability of the number of bands due to several technical reasons.
Starting from the above observations and inspired by \cite{Scarpa2018, Ciotola2022},
we proposed R-PNN \cite{Guarino2023},
a CNN-based band-wise pansharpening solution
where model parameters are not pre-trained but generated on the fly on the same target image at inference time.
Unlike previous works, suited to the MS case,
in R-PNN the PAN is fused with only one band at a time, rather than all of them, and the tuned model is passed from one band to another for targeted adaption.
The overall band-wise pansharpening process is summarized in Fig.~\ref{fig:rolling},
with the main symbols used here and in the reminder of the paper described in Tab.~\ref{tab:symbols}.

In mathematical terms,
the network of Fig.~\ref{fig:rolling}~(a) performs the mapping $\wh{H}_b = f(\H_b,\P;\phi)$, where $\phi$ is a set of trainable parameters.
For each band, $b$ (except band 1 starting from pretrained parameters),
a tuning process starts from the initial parameters, $\phi^{(0)}=\phi_{b-1}$, inherited from the previous band,
and runs for $N_b$ iterations to get the tuned parameters $\phi_b=\phi^{(N_b)}$.
$N_b$ is proportional (up to a bounding value) to the spectral distance between bands $b$ and $b-1$.
In fact, HS datasets often present spectral gaps due to discarded bands, and longer training helps making up for reduced band similarity.

Each tuning iteration is carried out on the same components $\H_b$ and $\P$ to be fused,
thanks to an unsupervised multitask loss which balances spectral and spatial consistencies of the fused image $\wh{H}_b$:
\begin{equation}
    \L = \L_\lambda\left(\wh{H}_b,\H_b \right) + \beta \L_S\left(\wh{H}_b,\P\right),
\label{eq:loss}
\end{equation}
with $\beta$ a balancing hyperparameter.
The spectral consistency is given by
\begin{equation}
    \L_\lambda \left(\wh{H}_b,\H_b \right) = \|\wh{H}_b^{(\downarrow)} -\H_b \|_1
    \label{eq:LL}
\end{equation}
where $\wh{H}_b^{(\downarrow)}$ is the downscaled (lowpass filtered and decimated) version of $\wh{H}_b$.
The spatial consistency, instead, is given by the average {\em local} correlation between $\widehat{\H}_b$ and $\P$.
Given the correlation coefficient $\rho(s)$, computed on a small $\sigma{\times}\sigma$ neighborhood of pixel $s$,
the loss is defined as
\begin{equation}
    \L_S\left(\wh{H}_b,\P\right) = \langle [\rho^{\rm max}(s)-\rho(s)]_+ \rangle
    \label{eq:LS}
\end{equation}
where $\langle x \rangle$ indicates spatial average, $[x]_+ = \max(0,x)$,
and $\rho^{\rm max}(s)$ is a rough estimate of the expected local correlation, computed on reduced resolution data.
Therefore, this spatial loss aims at maximizing $\rho(s)$, but only as long as it does not exceed a reasonable target value, $\rho^{\rm max}(s)$.

\newcommand{\sca}{0.8}

\begin{figure*}[!h]
\small
\centering
\begin{tabular}{rr}
\includegraphics[width=0.46\textwidth]{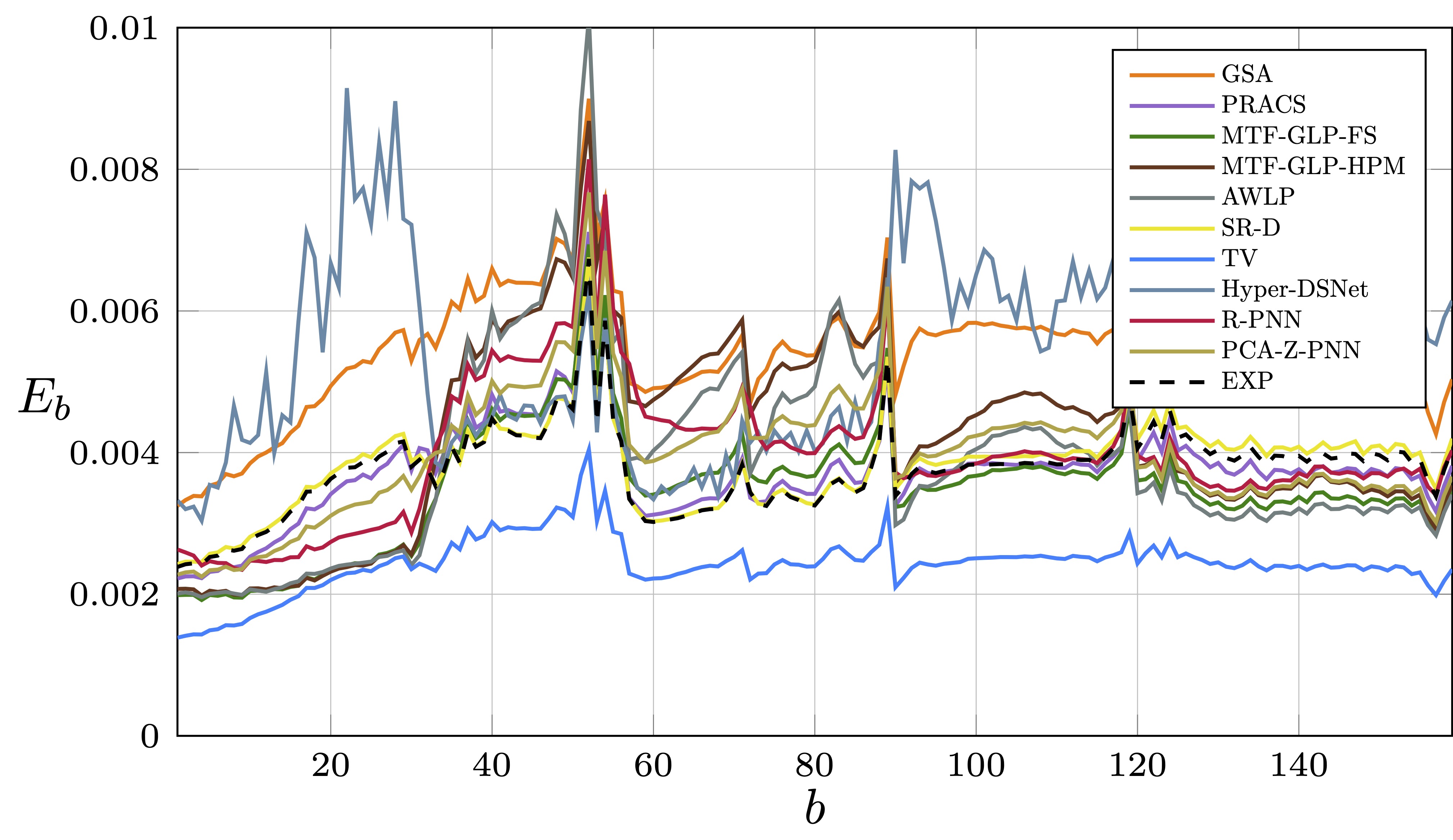}
&
\includegraphics[width=0.46\textwidth]{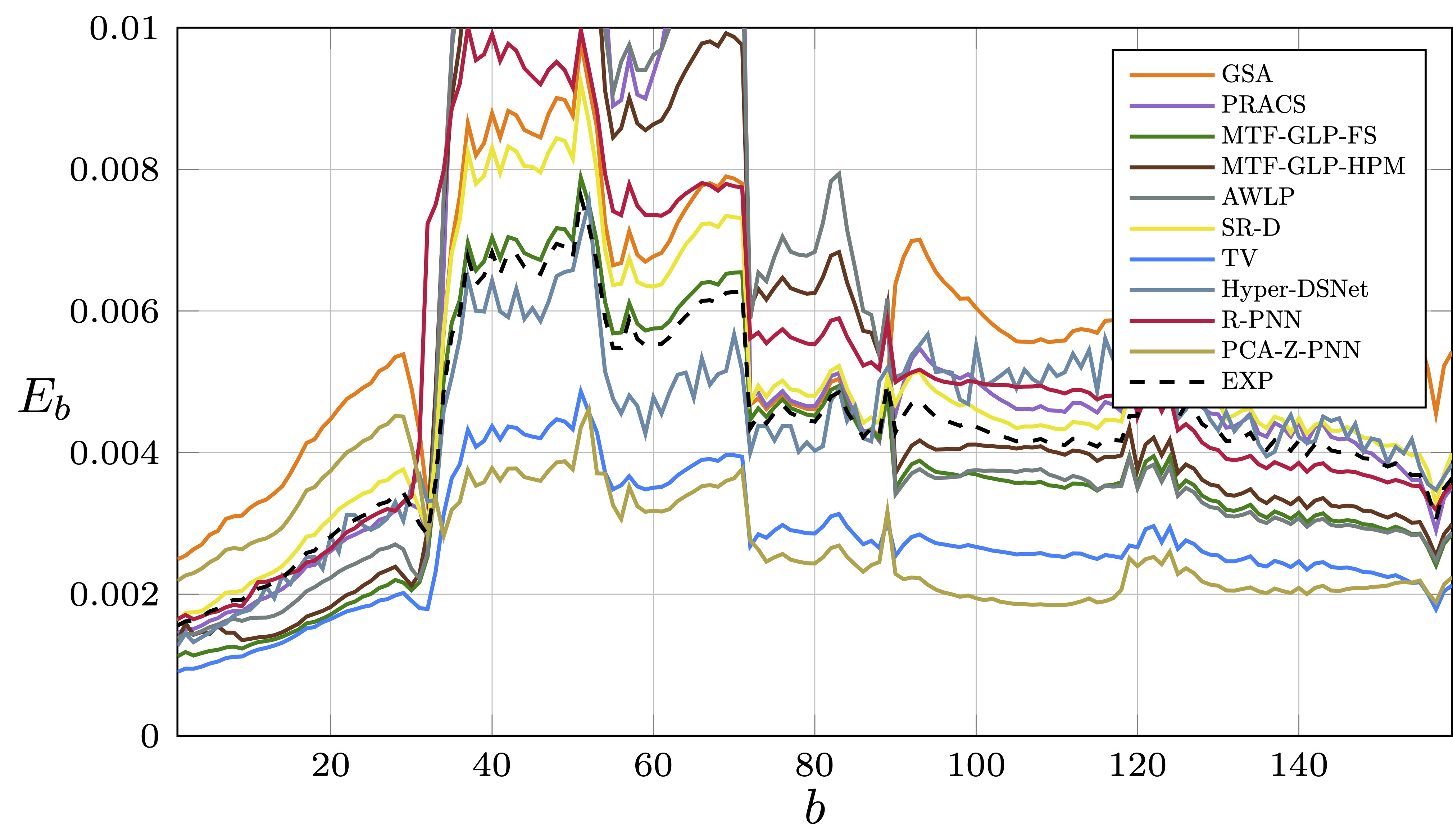} \\
\multicolumn{1}{c}{(a)} & \multicolumn{1}{c}{(b)} \\[2mm]
\includegraphics[width=0.442\textwidth]{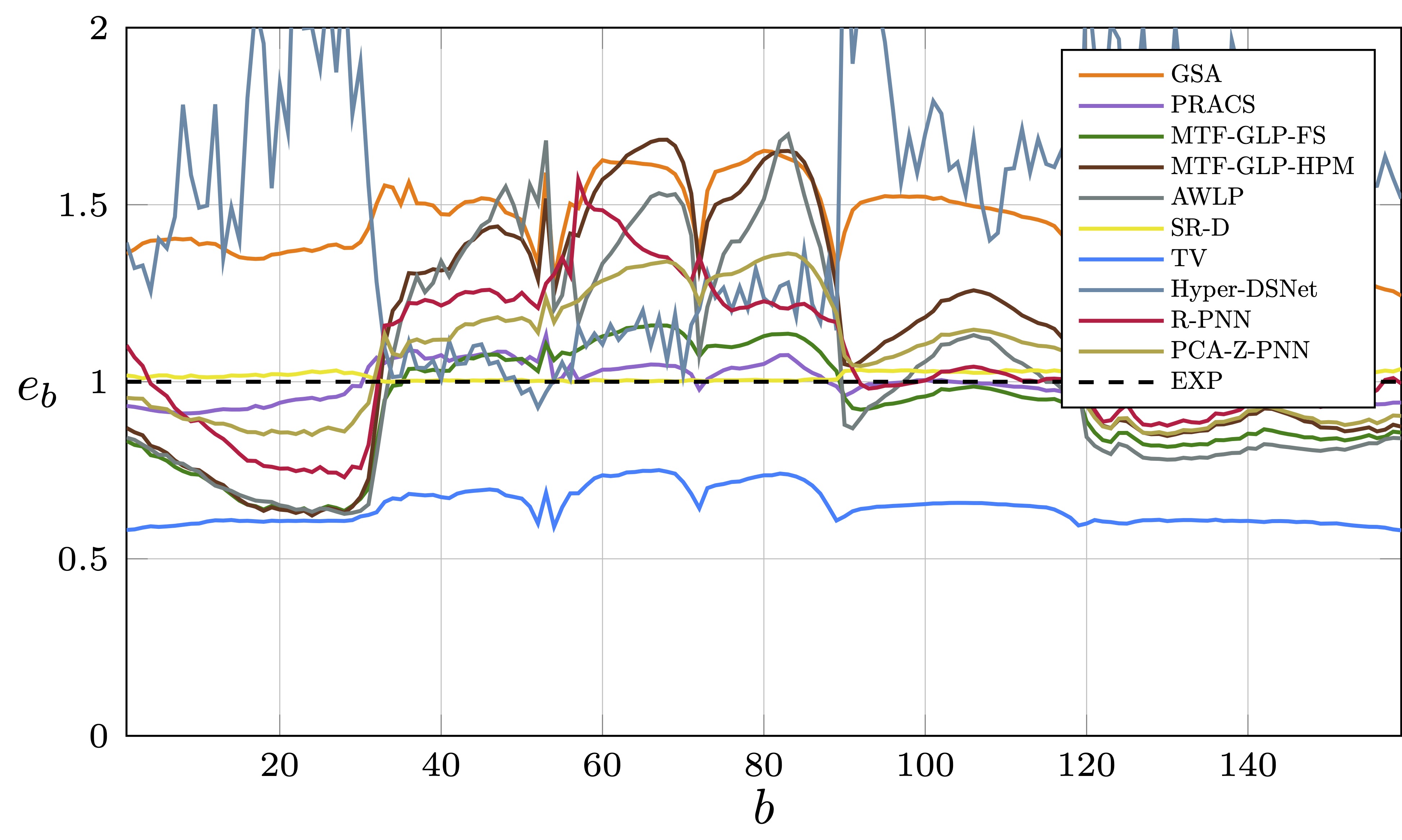}
&
\includegraphics[width=0.442\textwidth]{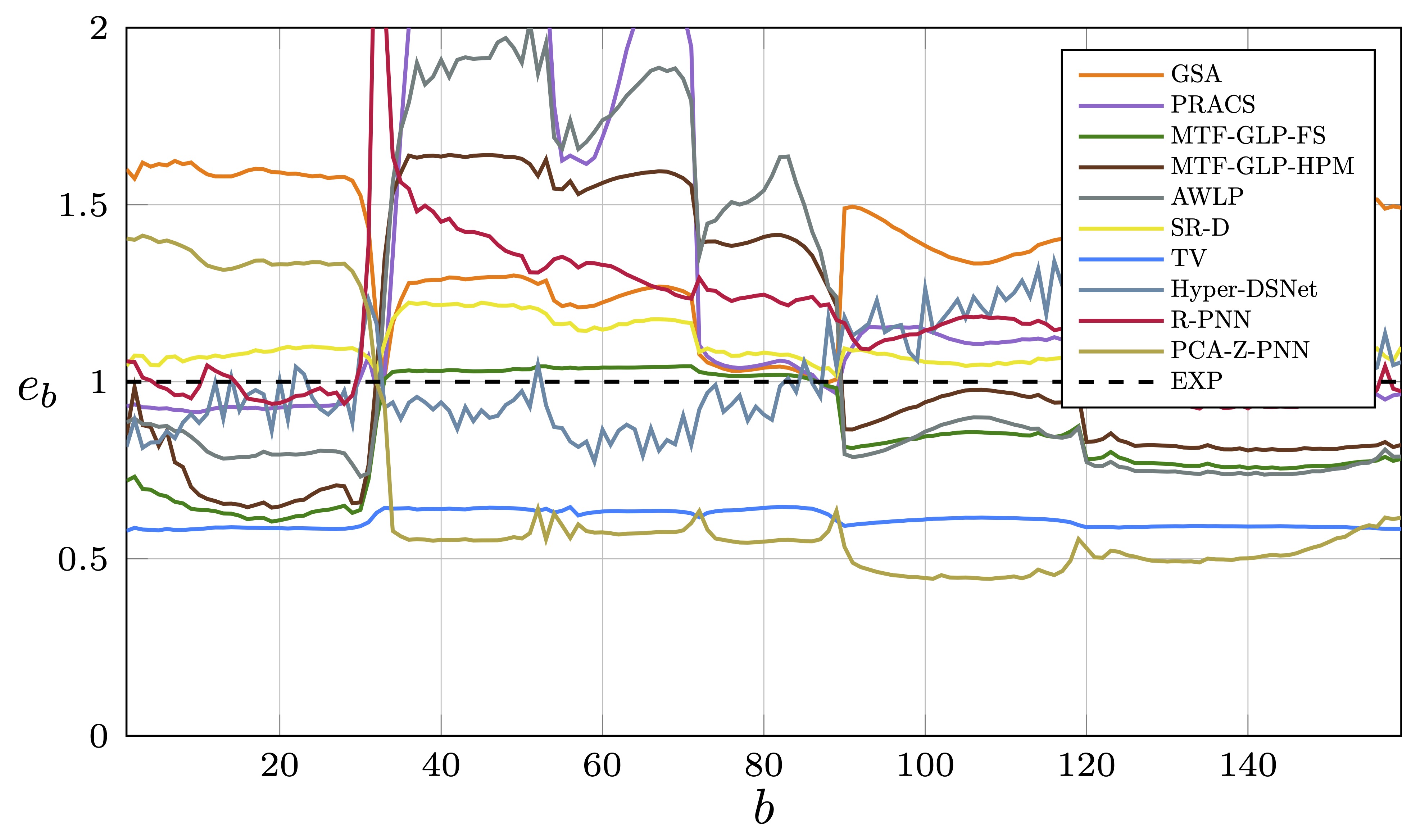}\\
\multicolumn{1}{c}{(c)} & \multicolumn{1}{c}{(d)}
\end{tabular}
\caption{Reprojection spectral error curves, $E_b$, for the PRISMA images Cagliari (a) and Udine (b), and corresponding normalized curves $e_b$ (c) and (d).}
\label{fig:mae}
\end{figure*}

\subsection{Analysis of spectral distortion}
For all methods of the benchmarking toolbox \cite{Ciotola2024}, and for two different PRISMA test images,
we computed the band-wise reprojection error\footnote{Although this quantity is often referred to as spectral distortion,
here we prefer to recall the presence of a downscaling step using the term reprojection.}
\begin{equation}
    E_b = \|\wh{\H}_b^{(\downarrow)} - \H_b \|_1
    \label{eq:mae}
\end{equation}
obtained by comparing the downscaled (smoothed and decimated) version of the pansharpened image with the input HS image.
This is the same distance used in R-PNN (\ref{eq:LL}) but other metrics provide similar results.
In Fig.~\ref{fig:mae}~(a)-(b) we plot results only for the methods characterized by the best full-resolution spectral quality.

The error is clearly band-dependent.
Almost all methods follow a common pattern, certainly related also to the different levels and properties of noise in different bands.
Even the simple EXP interpolator, which is a 23-tap polynomial approximation of the ideal interpolator, follows the same dynamics.
The average level of each curve provides information about the global distortion incurred by the method
(remember that we are neglecting spatial quality in this analysis)
but the detailed behavior of the curves shows that numbers vary significantly from band to band, and that the average distortion may be a poor indicator of the actual distortion observed in certain frequency ranges.
Under this point of view, a uniform level of distortion would be certainly an appreciable feature.

To simplify the analysis of results let us consider a normalized version of the same data.
Since the EXP interpolator works exclusively on $\H$, without adding any high-frequency spatial detail coming from the PAN,
it can be regarded as a method-independent ``spectral reference''.
Therefore, we use its reprojection error $E^{\rm EXP}_b$ to relativize the profiles of Fig.~\ref{fig:mae} by defining a normalized error $e_b$ as
\begin{equation}
    e_b = E_b \,/\, E^{\rm EXP}_b
    \label{eq:relative}
\end{equation}
The resulting curves are plot in Fig.~\ref{fig:mae}~(c)-(d).
Spectral error profiles are now easier to analyze.
As ``virtuous'' example we can mention TV which, for both test images,
keeps the normalized reprojection error low (about 60\% of EXP) and almost constant over the whole spectral range.
In contrast, other methods show significant imbalances between different spectral ranges.
This may be partly due to spatial detail injection rules that do not properly account for the spectral-spatial interdependence.
For example, several methods achieve very low spectral distortions in the visible range (first $\sim$30 bands) with a sharp performance impairment beyond the near infrared limit,
possibly due to the inappropriate injection of spatial details in a range where the correlation between PAN and HS bands decreases.
On the opposite side we mention the case of PCA-Z-PNN,
which follows a {\em divide-and-conquer} approach where the HS bands are split in two groups and separately processed.
This split is clearly reflected in the almost piece-wise uniform $e_b$ curve (see Fig.~\ref{fig:mae}~(d)), with a relatively large distortion in the visible range and a much better behavior beyond that range.

Overall, this preliminary analysis motivates us to work on a pansharpening method that adapts
to the different characteristics exhibited by HS bands in different spectral ranges and provides nearly uniform quality across the spectrum.
R-PNN, with its band-wise optimization procedure, is an excellent candidate to achieve this goal.

\section{Proposed Method}
\label{sec:method}

\newcommand{\markgp}[1]{\noindent \red{[concept: #1] \newline\noindent}}
\newcommand{\LE}{\L_{\lambda/E}}
\newcommand{\gLOW}{\gamma_{_{\rm LOW}}}
\newcommand{\gHIGH}{\gamma_{_{\rm HIGH}}}

\subsection{Correlation-based Spatial Loss for HS Images}

In our proposal, we keep the two-component loss formulation (\ref{eq:loss}) previously adopted \cite{Ciotola2023} in unsupervised pansharpening
and use the same spectral component of \cite{Guarino2023}, based on the $\ell_1$-norm (\ref{eq:LL}),
which shows good convergence properties and outperforms other norms, such as $\ell_2$ or ERGAS, in terms of perceptual quality.
In any case, the exact form of the spectral loss is not really critical
as it is always obtained by comparing the pansharpened product $\wh{\H}$ with a clear reference,
the interpolated HS image $\wt{\H}$ or (after decimation) the low-resolution HS itself.
We focus instead on the spatial loss term,
which is more elusive as it is intimately related to the problem of assessing spatial quality without a reference.
In R-PNN we used the correlation-based loss (\ref{eq:LS}) originally proposed in \cite{Ciotola2022} for MS pansharpening.
The idea is to push the pansharpened band to have a high local correlation with the PAN, so as to exhibit the same spatial layout.
Indeed, the correlation coefficient measures how well one image can be linearly predicted from another.
Since pansharpened bands and PAN share the same spatial structure, they can be expected to be strongly correlated.
On the other hand, each band has unique spectral properties that must be carefully preserved.
Therefore, to
prevent the injection of alien details from the PAN into the spectral band,
the correlation itself, $\rho$, is capped by its estimate, $\rho^{\rm max}(s)$, computed on reduced resolution data.

This correlation-based loss proved extremely effective for MS pansharpening, ensuring results characterized by very high spatial fidelity \cite{Ciotola2022,Ciotola2023}.
Its extension to the HS case, however, is more controversial.
In fact, the basic assumption is that $\rho(s)$ is generally high and certainly positive.
While this is more than reasonable for the MS case, it becomes largely questionable in the HS case
because many spectral bands fall outside the visible range (i.e., the PAN range) and exhibit a lower correlation with it,
and sometimes the so-called inversion phenomenon \cite{Thomas2008} with regions having strong but negative correlation.

To clarify this point, we carried out a correlation analysis on a sample PRISMA image (Kansas).
Lacking the ideal pansharpened datacube, we estimate the local correlation on reduced resolution data,
using the canonical resolution downgrade of the PAN to align it with the available HS datacube.
We select two sample spectral bands which represent opposite behaviors,
one in the visible spectrum, thus highly correlated with the PAN, the other outside this range and with much lower correlation.
In Fig.~\ref{fig:corr} (top row) we show the downgraded PAN $\P^{(\downarrow)}$, the two spectral bands, $\H_v$ (from the visible spectrum) and $\H_o$ (outside),
together with the corresponding correlation maps, $\rho_v=\rho_{\P^{(\downarrow)}\H_v}$ and $\rho_o=\rho_{\P^{(\downarrow)}\H_o}$, respectively.
The rightmost image is a quantized representation of $\rho_o$ used to simplify its interpretation.
On the bottom row of the figure, the same items are shown for a meaningful close-up of the image (green box).

\definecolor{cA}{rgb}{0.0, 0.1, 0.6}
\definecolor{cB}{rgb}{0.0, 0.8, 1.0}
\definecolor{cC}{rgb}{1.0, 1.0, 1.0}
\definecolor{cD}{rgb}{1.0, 0.8, 0.0}
\definecolor{cE}{rgb}{0.6, 0.1, 0.0}
\newcommand{\legenda}[2]{{\tikz{\node[#1,fill,draw=black,rectangle,minimum width = 18pt, minimum height= 5pt](l) at (0,0){};
				\node[right of=l, anchor=west, xshift = -18pt]{#2};}}}
\newcommand{\whitelegenda}[1]{{\tikz{\node[draw,rectangle,minimum width = 18pt, minimum height= 5pt](l) at (0,0){};
				\node[right of=l, anchor=west, xshift = -18pt]{#1};}}}
\newcommand{\scalebarme}[1]{%
    \begin{Overpic}{%
    #1%
    }%
    \put (74,35){\tikz{\path[draw=green,line width=1pt] (0,0) rectangle (0.6cm,0.6cm);}}%
    \end{Overpic}%
    }

\newcommand{\png}[1]{\includegraphics[width=0.3\columnwidth]{./figures/png/#1}}
\begin{figure*}
\setlength{\tabcolsep}{1.5pt}
\centering
\begin{tabular}{c@{\rule{15pt}{0pt}}cc@{\rule{15pt}{0pt}}ccc}
PAN: $\P^{(\downarrow)}$ & $\H_v$                & $\rho_v$, $\langle \rho_v\rangle = 0.95$ & $\H_o$                & $\rho_o$, $\langle \rho_o\rangle = 0.33$ & $\rho_o$ (quantized)   \\
\scalebarme{\png{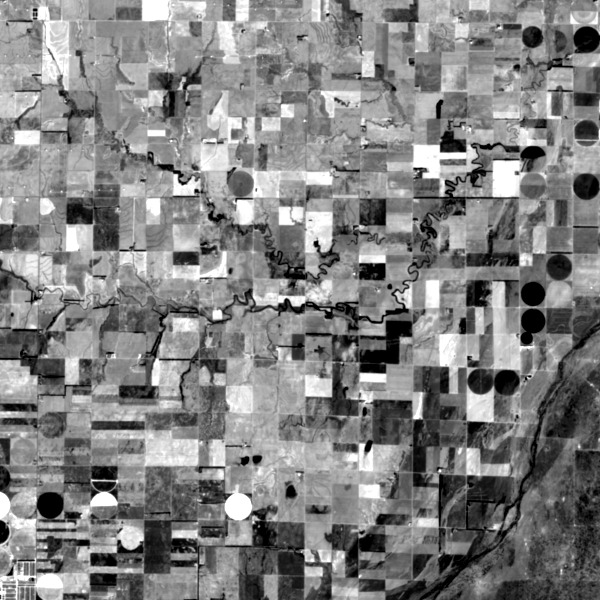}}     & \scalebarme{\png{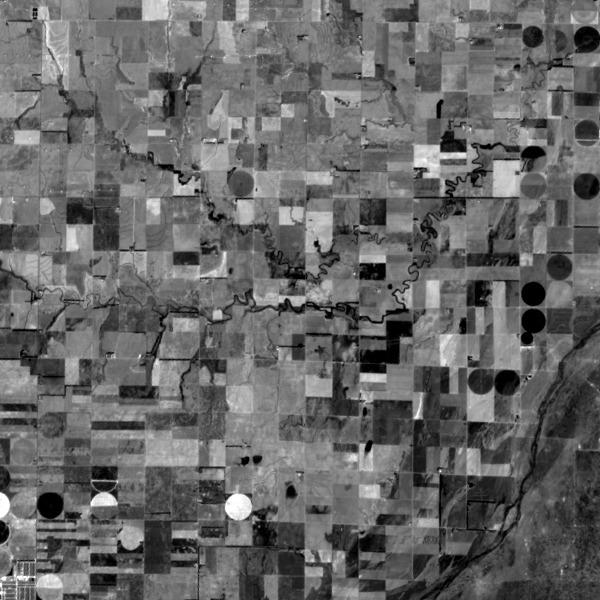}} & \scalebarme{\png{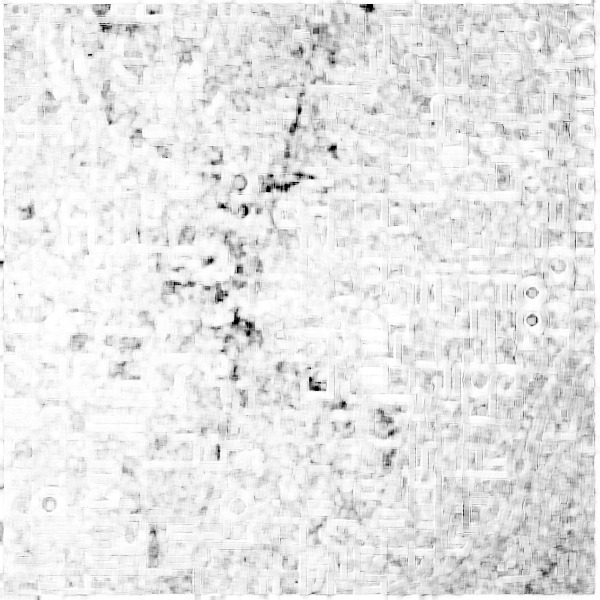}}                    & \scalebarme{\png{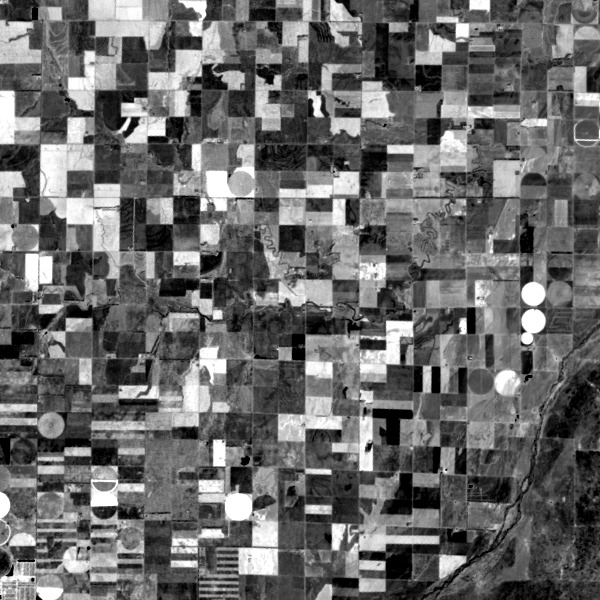}} & \scalebarme{\png{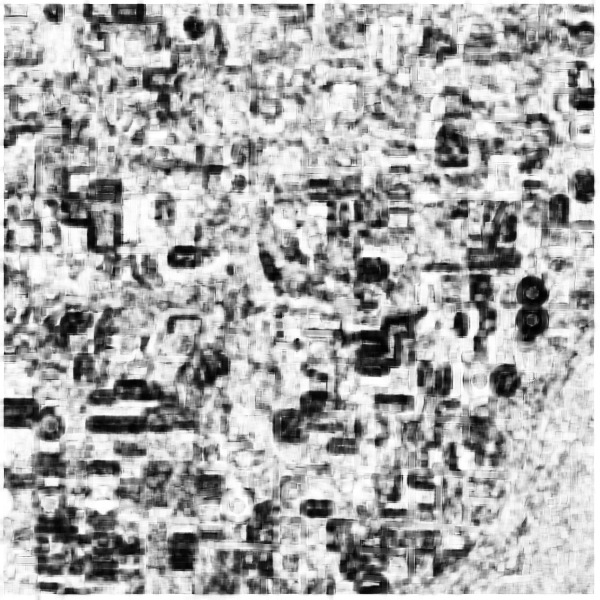}}                    & \scalebarme{\png{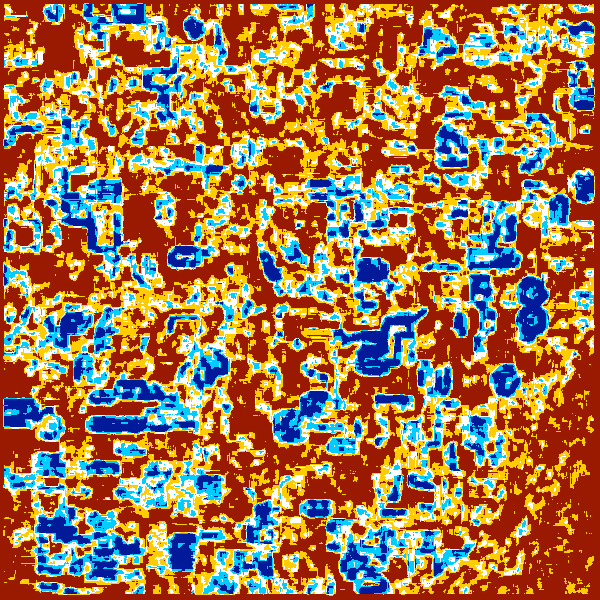}} \\
\png{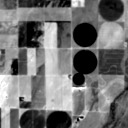}                & \png{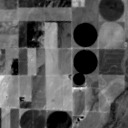}            & \png{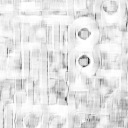}                               & \png{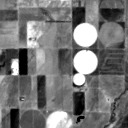}            & \png{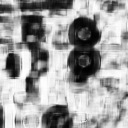}                               & \png{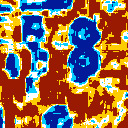}            \\
%\multicolumn{3}{l}{\legenda{cE}{\footnotesize strong correlation: $0.6<\rho_o\leq 1$}}
%		& \multicolumn{3}{l}{\legenda{cA}{\footnotesize strong (negative) correlation: $-1 \leq \rho_o < -0.6$}}\\[-2mm]
%\multicolumn{3}{l}{\legenda{cD}{\footnotesize moderate correlation: $0.2 <\rho_o\leq0.6$}}
%		& \multicolumn{3}{l}{\legenda{cB}{\footnotesize moderate (negative) correlation: $-0.6\leq\rho_o<-0.2$}}\\[-2mm]
%\multicolumn{6}{l}{\legenda{cC}{\footnotesize weak correlation: $-0.2 \leq\rho_o \leq 0.2$}}
\multicolumn{6}{l}{
\legenda{cA}{\footnotesize $-1.0 < \rho_o \leq -0.6$}\hspace{4mm}
\legenda{cB}{\footnotesize $-0.6 < \rho_o \leq -0.2$}\hspace{4mm}
\legenda{cC}{\footnotesize $-0.2 < \rho_o \leq  0.2$}\hspace{4mm}
\legenda{cD}{\footnotesize $ 0.2 < \rho_o \leq  0.6$}\hspace{4mm}
\legenda{cE}{\footnotesize $ 0.6 < \rho_o \leq  1.0$}
}
\end{tabular}
\caption{PAN-HS local correlation analysis.
First row, from left to right: PAN image shifted at the HS spatial resolution ($\P^{(\downarrow)}$),
two sample HS bands paired with the corresponding map of local correlation with the PAN, from visible ($\H_v, \rho_v$)
and NIR-SWIR ($\H_o, \rho_o$) spectral ranges, and (rightmost) the quantized version of the correlation map $\rho_o$.
Second row: close-up on a meaningful detail (highlighted with a green box).}
\label{fig:corr}
\end{figure*}

A few observations are in order.
The sample band from the visible spectrum shows a very high correlation with the PAN, 0.95 on the average and never negative.
This is a common behavior in the visible spectrum, more pronounced in the green range and a little less towards the coastal and red/infrared limits.
In this situation, maximizing the local correlation, without any upper bound, makes perfect sense.
Outside the visible spectrum, however, the correlation field has a much higher dynamics, and our sample band has an average correlation of just 0.33 with the PAN.
Such a low value, however, does not signal lack of dependence.
The spatial layouts keep being mostly the same, as easily appreciated in the close-up image, but several regions exhibit a strong {\em negative} correlation.
This is the case, for example, of the circularly shaped elements clearly visible in the close-up, which are very dark in the PAN, and very bright in $\H_o$.
In addition, intermediate correlation values, between -0.6 and 0.6, occur at the boundaries between positively and negatively correlated regions just because of the finite size of the estimation window (6$\times$6 in our experiment).
This latter problem impacts also on the estimation of the bounding map, $\rho^{\rm max}$, further reducing its effective resolution and making it scarcely reliable.
An algorithm that disregards these phenomena may generate visible pansharpening artifacts,
such as unnatural color variations and exaggerated contrasts.
Moreover, conflicts may arise between the spectral and spatial terms of the loss, hindering the training or fine-tuning phases.

Based on the above observations,
we decided to define a spatial loss term that maximizes the {\em absolute value} of the correlation with the PAN,
so as to promote structural similarity irrespective of the correlation polarity.
\begin{equation}
    \L_S\left(\wh{H}_b,\P\right) = \left\langle 1- |\rho(s)| \right\rangle
    \label{eq:LSnew}
\end{equation}
Note that we have also removed the bounding map $\rho^{\rm max}$ due to its limited reliability.
In fact, a suitable hysteresis-based optimization schedule will prevent the injection of inappropriate PAN details in the pansharpened image.

Armed with our new loss, comprising the original spectral term (\ref{eq:LL}) and the new spatial term (\ref{eq:LSnew}),
we can now pursue our original goal, that is, devising a pansharpening algorithm that provides a uniform quality across the whole spectrum.
More precisely,
considering that the various bands experience very different noise intensities,
we aim at achieving a uniform {\em relative} quality,
measured by the error normalized with respect to the EXP ideal interpolator (\ref{eq:relative}).

\subsection{Network Architecture}
\label{sec:network}

As said before, we rely on rolling pansharpening \cite{Guarino2023} as it allows us to operate band-wise.
With this strategy, only one low-resolution interpolated spectral band is given in input, together with the PAN, and only one high-resolution spectral band is provided in output.
Therefore, any pansharpening network can be adapted to the purpose by setting appropriately the depth of the input and output layers.
The network must be trained from scratch for the first band and then refined for each new band by target-adaptive tuning.
Of course, for an HS image, this latter process must be repeated many times,
which calls for the intrinsic requisite of a lightweight network.

We have implemented and tested several low-complexity solutions from the MS pansharpening literature
and verified experimentally that network structure and capacity have a minor impact on performance provided some minimal features are guaranteed:
{\it   i)} no less than three convolutional layers;
{\it  ii)} a global receptive field of a reasonable size;
{\it iii)} use of a global skip connection (residual net).
Given the above indications,
we decided to simply keep the network used in R-PNN, a single-band residual version of PNN \cite{Scarpa2018}, which meets all these criteria.
Tab.~\ref{tab:PNN} summarizes the network layers and connections.
\begin{table}
\centering
\caption{Network layers and topology.}
\footnotesize
\label{tab:PNN}
\begin{tabular}{cccc}
\hline
Layer             & Size                   & Kernel       & Outputs                    \\ \hline
Input             & $W{\times}H{\times}2$  & -            & $(\wt{\H}_b,\P)$           \\
Conv+ReLU         & $W{\times}H{\times}48$ & $7{\times}7$ & -                          \\
Conv+ReLU         & $W{\times}H{\times}32$ & $7{\times}7$ & -                          \\
Conv              & $W{\times}H{\times}1$  & $5{\times}5$ & $\mathbf{D}_b$             \\
Output (add gate) & $W{\times}H{\times}1$  & -            & $\wt{\H}_b + \mathbf{D}_b$ \\
\hline
\end{tabular}
\end{table}

\subsection{Hysteresis-based Training Procedure}

\newcommand{\trajectNoHyst}[1]{\includegraphics[width=0.40\textwidth]{./figures/lossTraject_no_hyst_B#1.jpg}}
\begin{figure}
\setlength{\tabcolsep}{1pt}
\centering
\begin{tabular}{c}
\trajectNoHyst{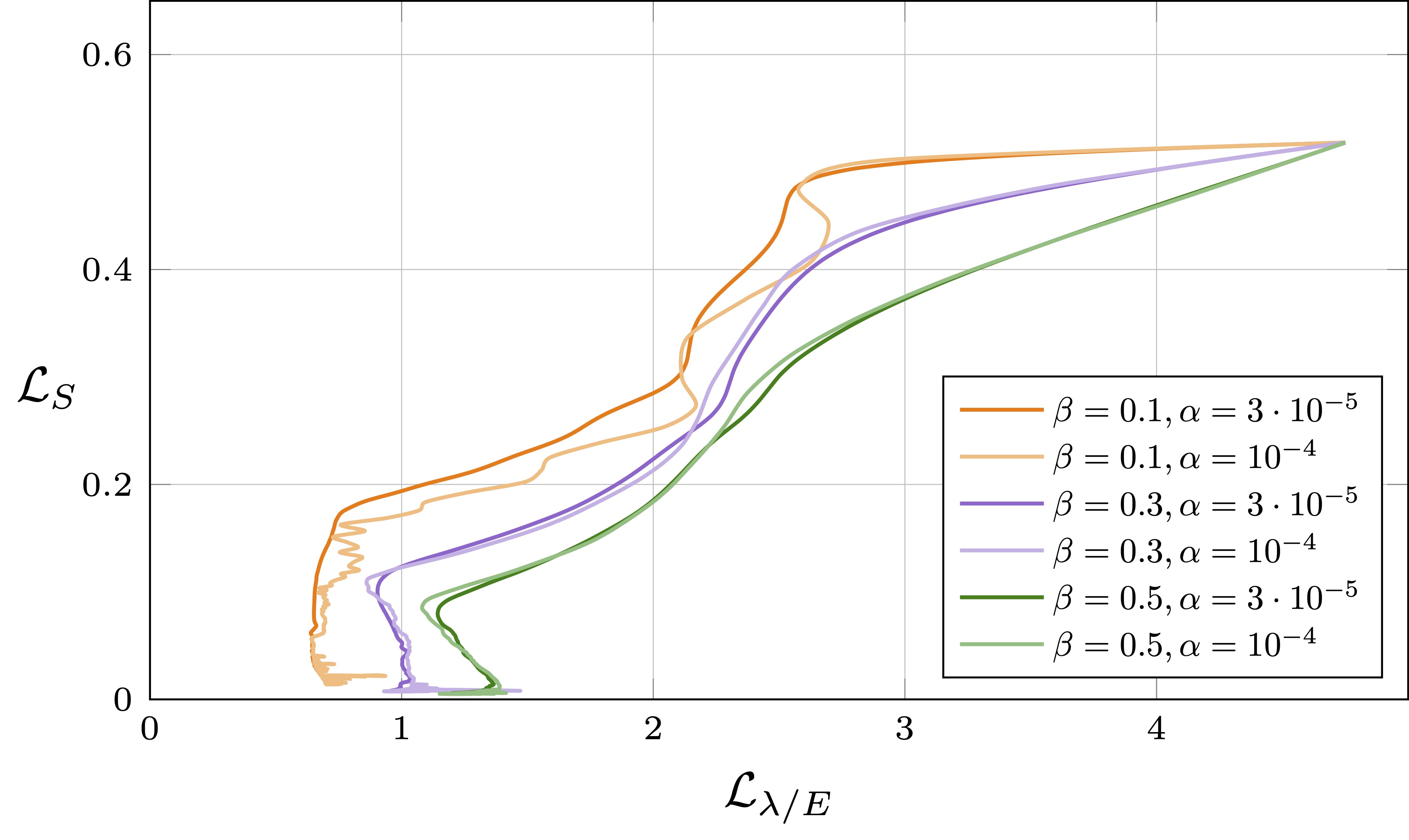}\\
\trajectNoHyst{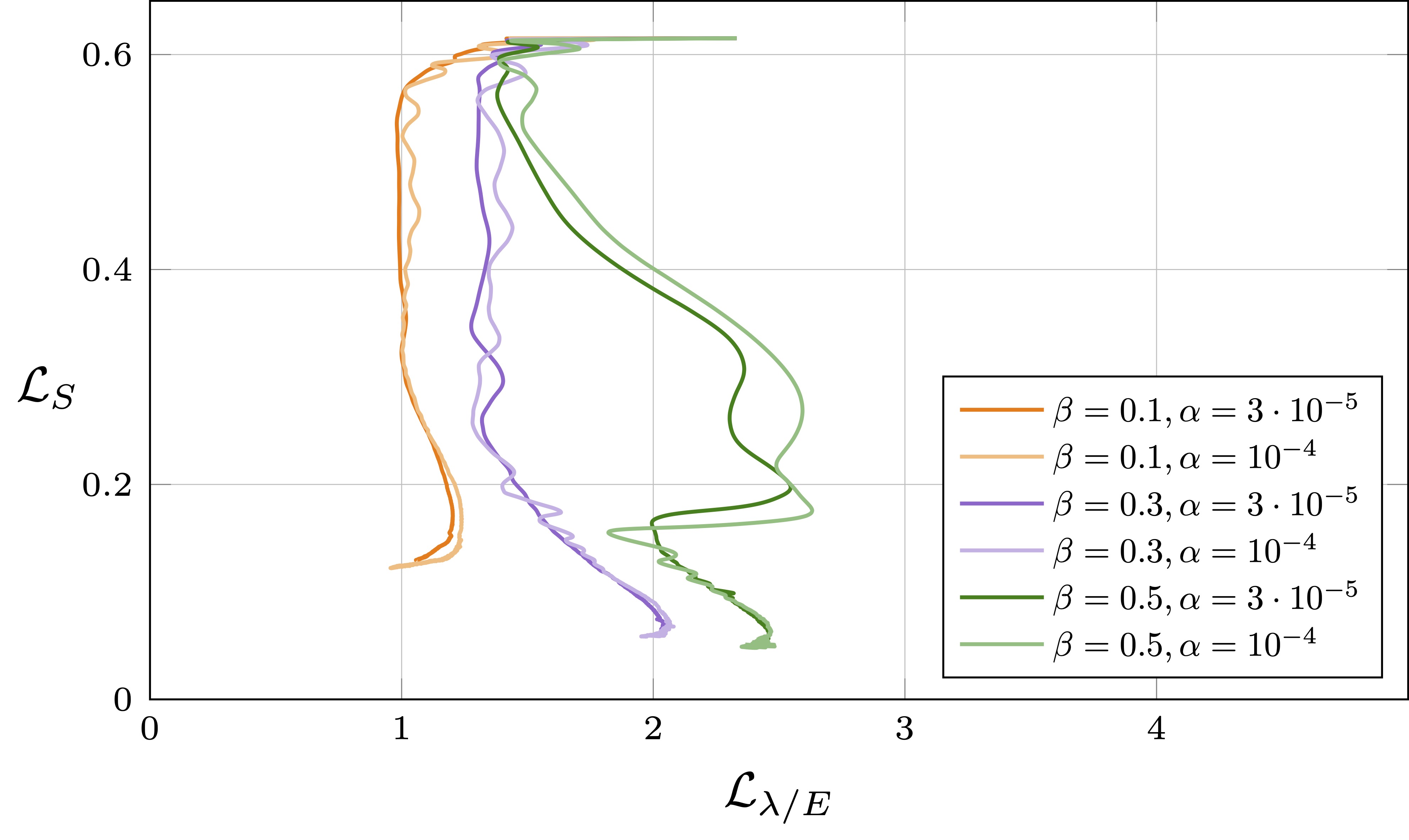}
\end{tabular}
\caption{
Loss trajectories in the $(\LE,\L_S)$ plane when tuning models with various choices of the $\alpha$ and $\beta$ parameters.
Tuning starts from random initial weights and proceeds for a large number (1000) of iterations.
Sample bands from the visible range (top), and outside it (bottom).}
\label{fig:trajectoryNoHyst}
\end{figure}

\newcommand{\traject}[1]{\includegraphics[width=0.32\textwidth]{./figures/lossTraject_#1.jpg}}
\begin{figure*}
\setlength{\tabcolsep}{1pt}
\centering
\begin{tabular}{ccc}
\traject{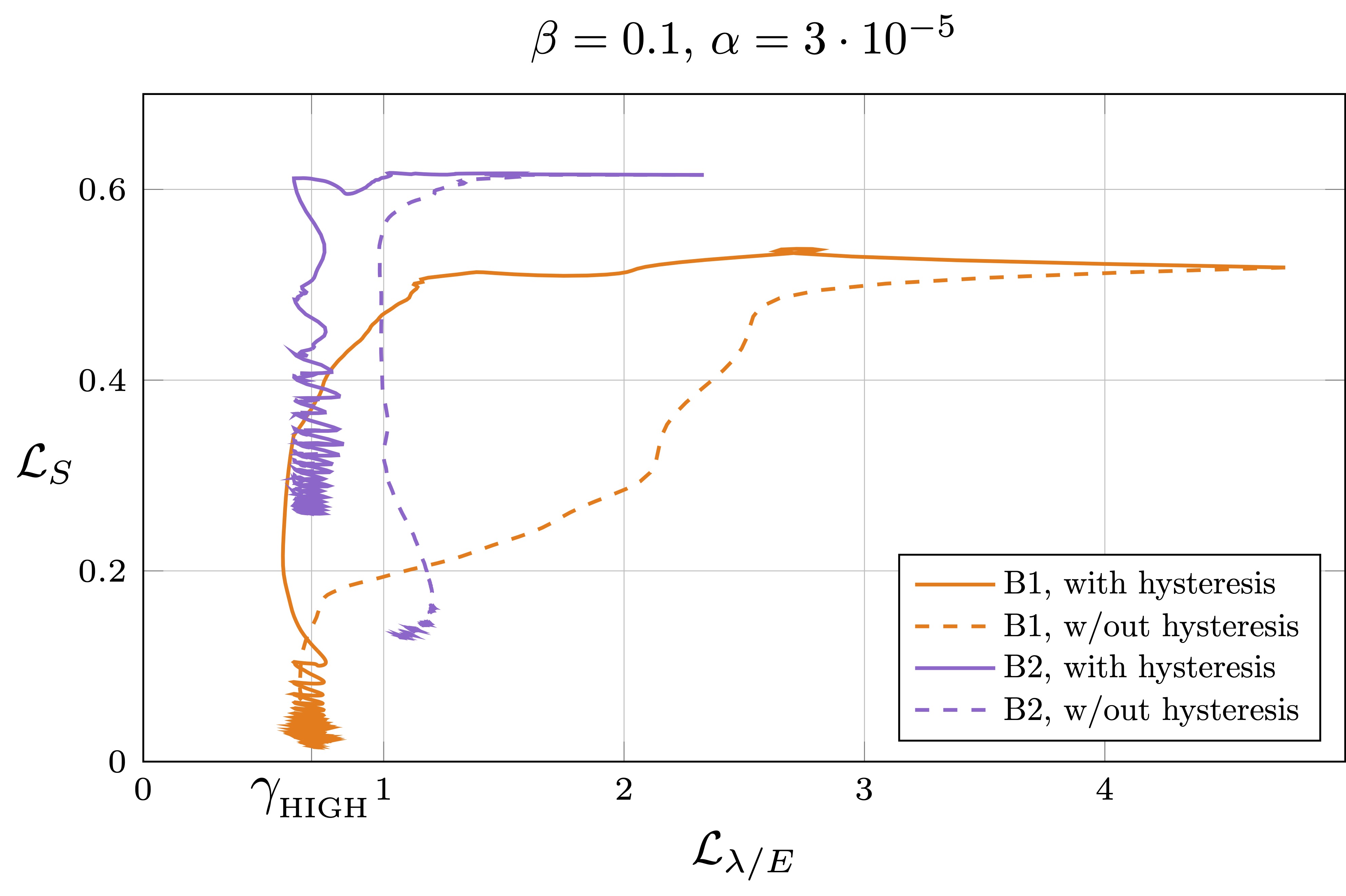} & \traject{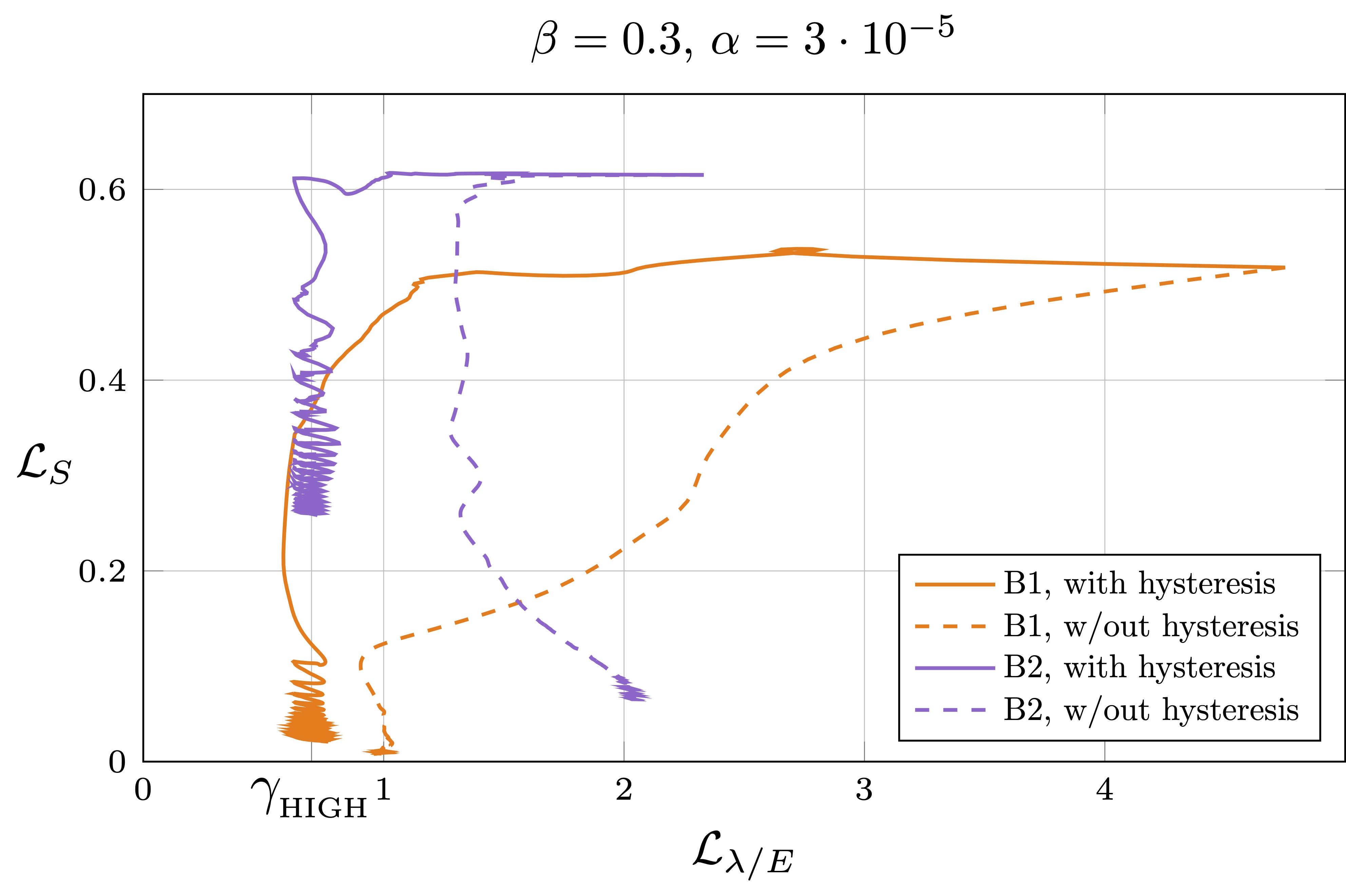} & \traject{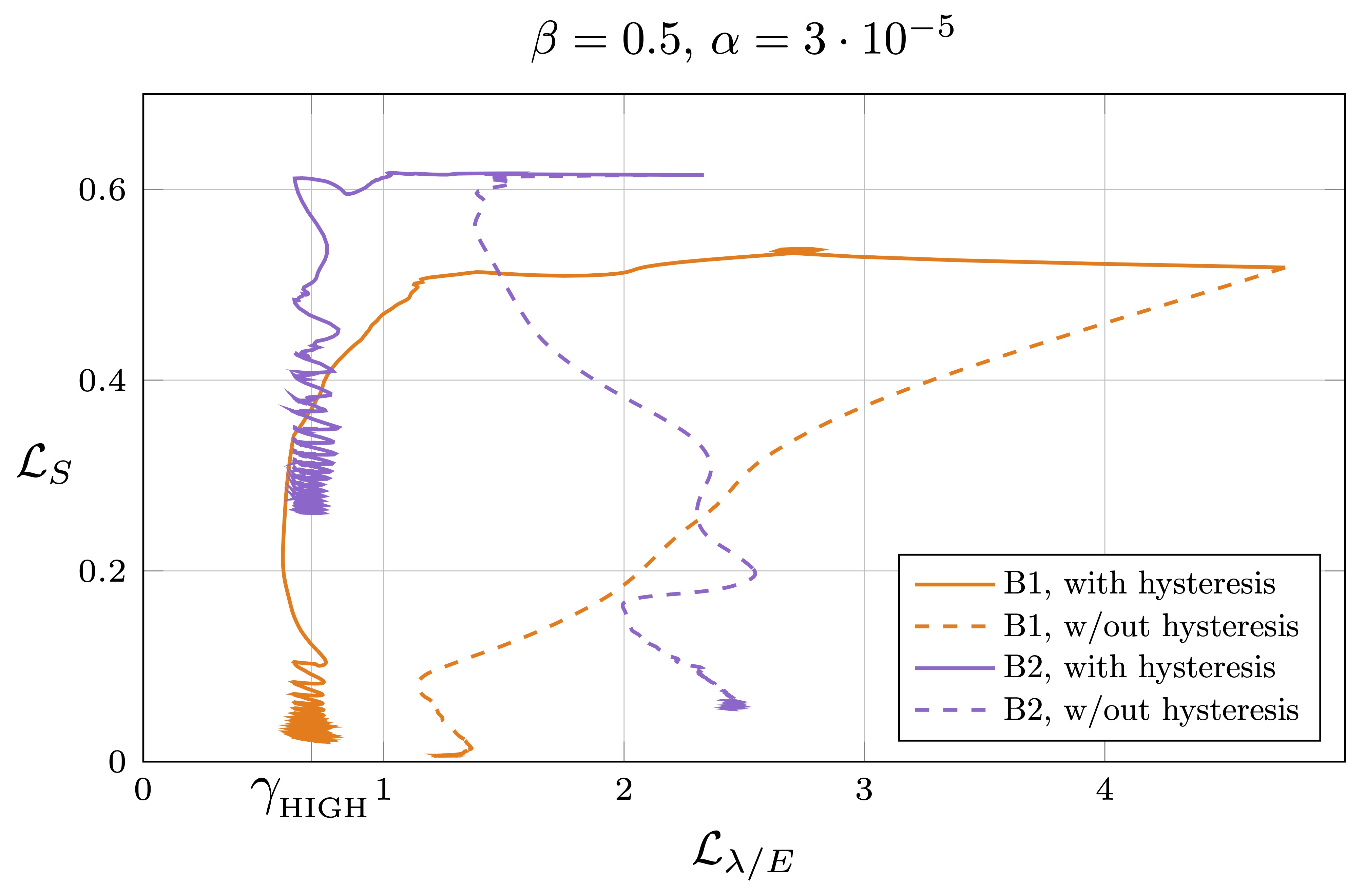} \\
%\traject{lr5e5_beta01} & \traject{lr5e5_beta03} & \traject{lr5e5_beta05} \\
\traject{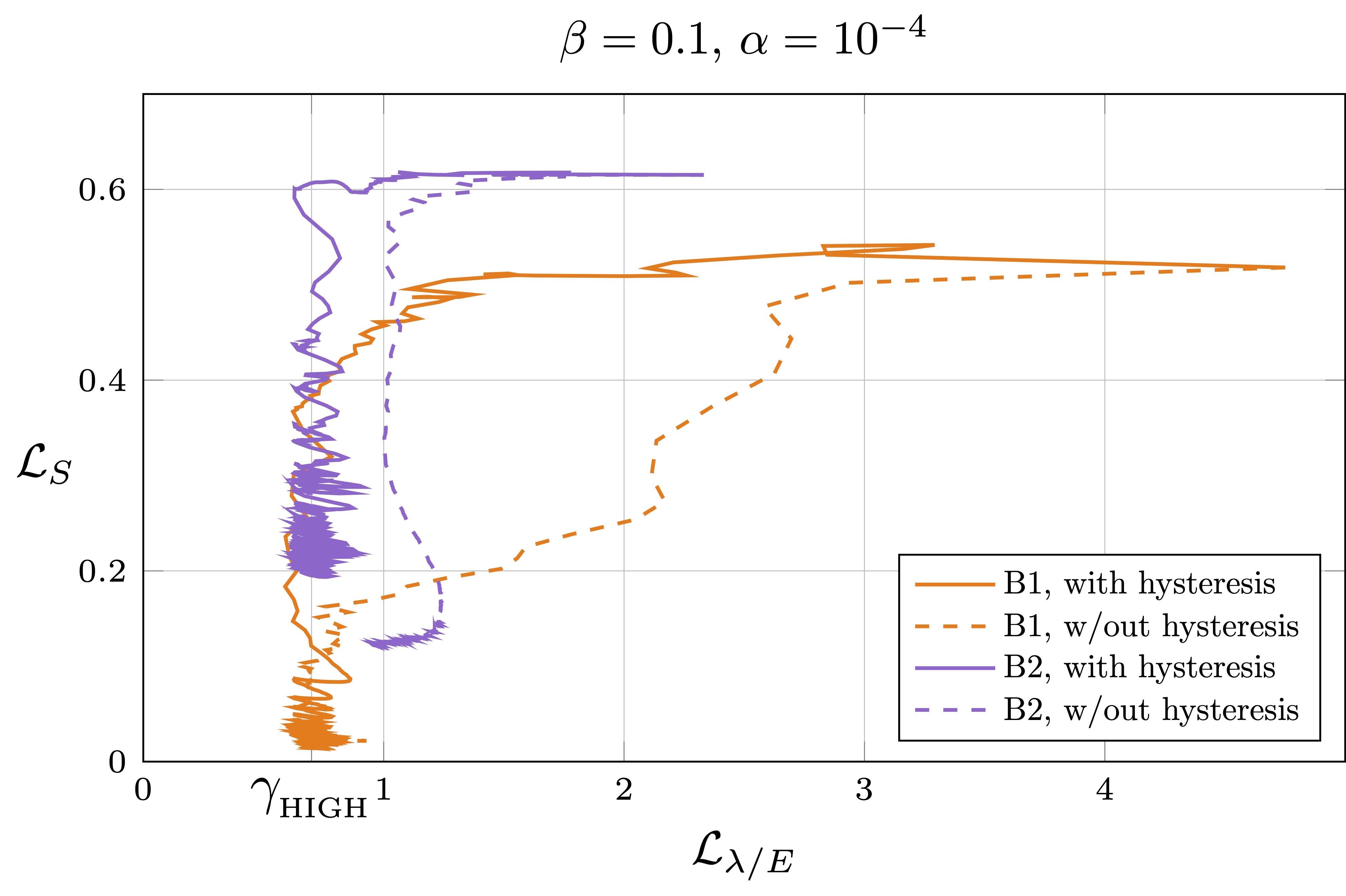} & \traject{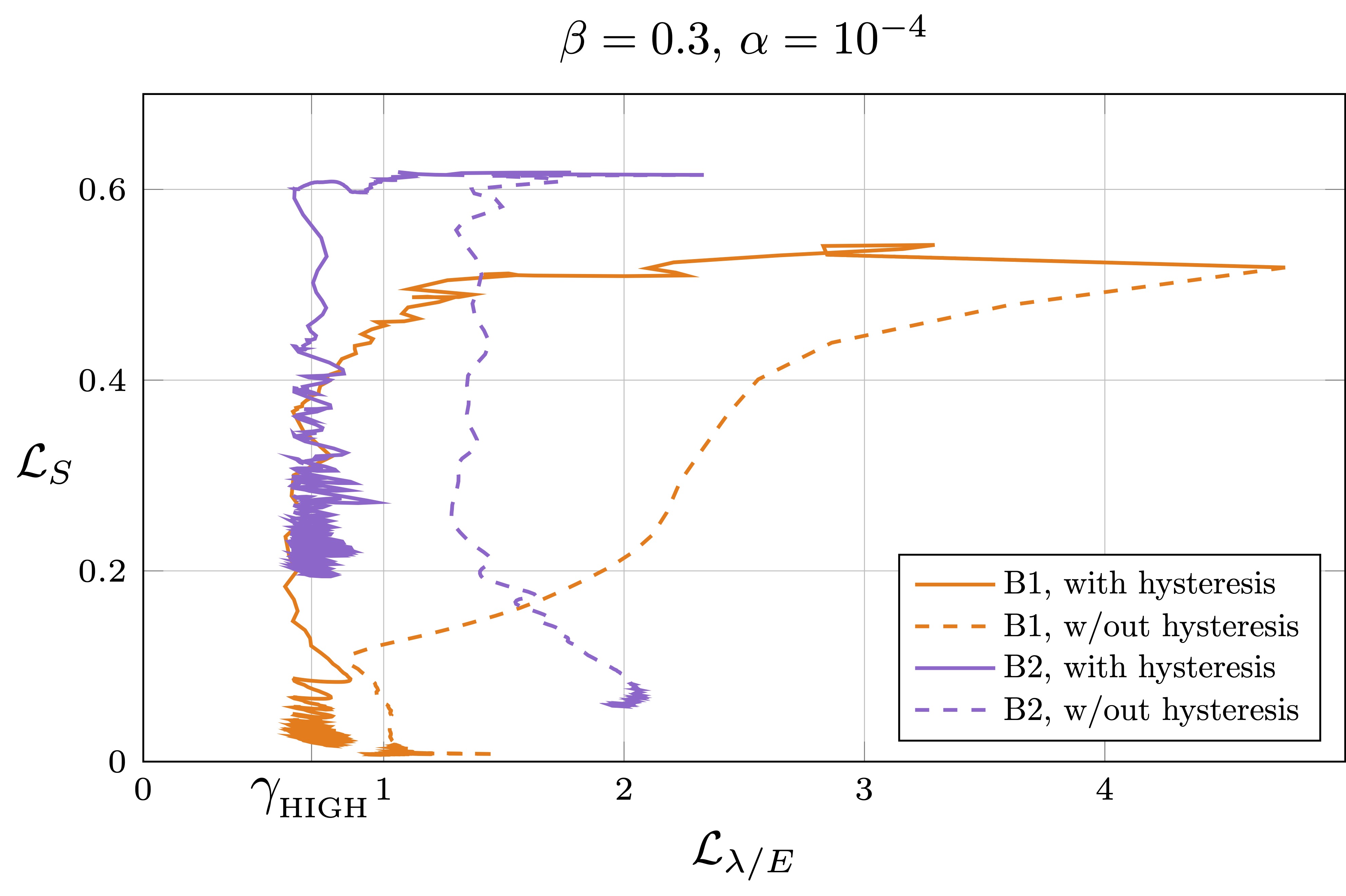} & \traject{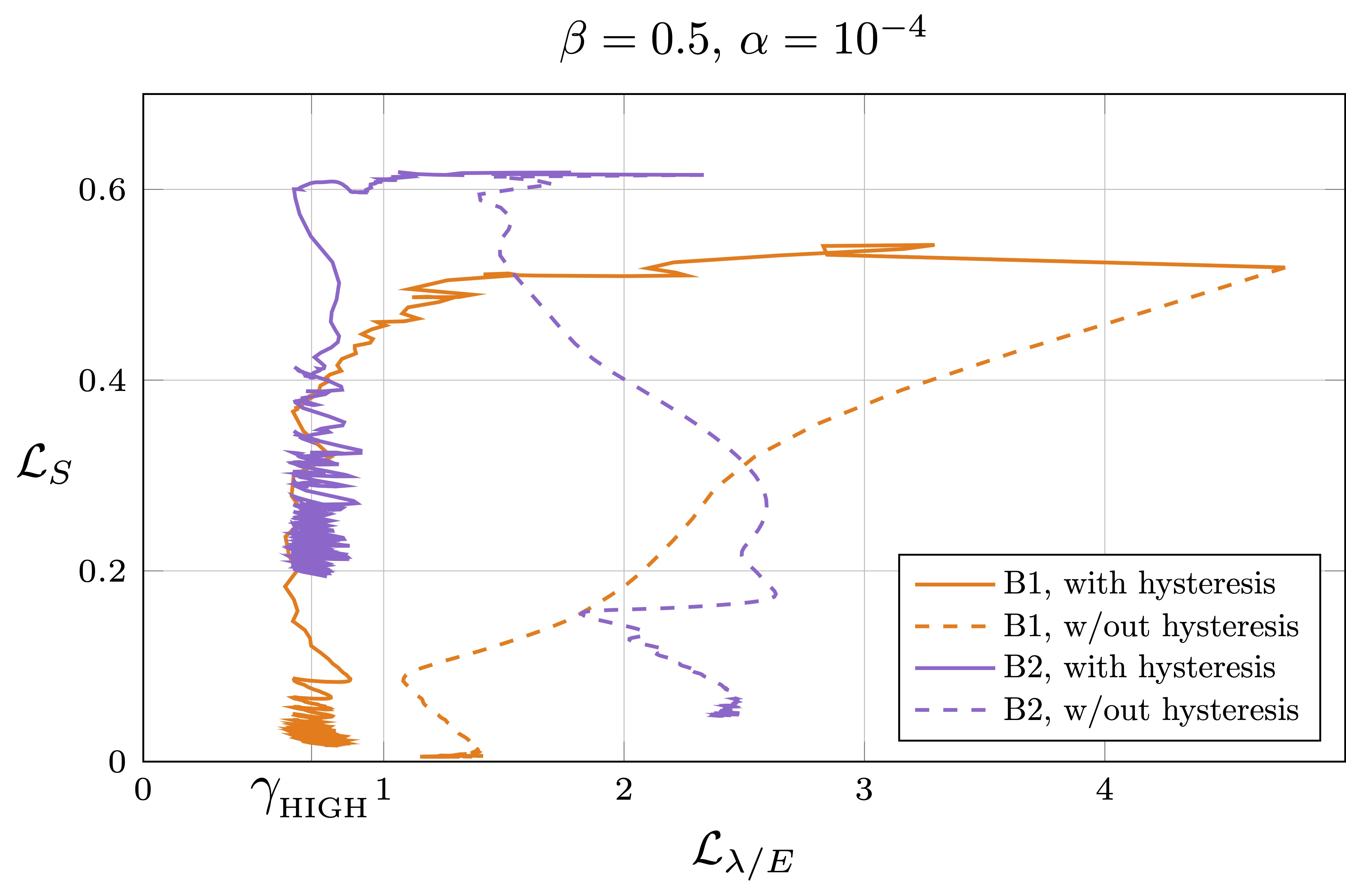}
\end{tabular}
\caption{
Loss trajectories in the $(\LE,\L_S)$ plane when tuning models with various choices of the $\alpha$ and $\beta$ parameters.
Tuning starts from random initial weights and proceeds for a large number (1000) of iterations.
In each chart,
the curves refer to
two different bands, B1 from the visible spectrum (violet) and B2 outside it (orange), and
two different tuning schedules, the proposed histeresis-like strategy (solid lines) and flat tuning (dashed lines).
}
\label{fig:trajectory}
\end{figure*}

Once fixed the network architecture, we can focus on optimizing network training and fine tuning.
To this end we can play with several hyperparameters, first of all
the learning rate $\alpha$,
the weight $\beta$ that balances the spectral and spatial terms in the loss,
and the number of iterations $N_b$.
To gain insight into the problem, we carried out some preliminary experiments
using various $(\alpha,\beta)$ pairs and studying their impact on the two loss components, $\L_S$ and $\L_\lambda$
(these losses depend also on the band, $b$, but we avoid a further subscript for the sake of clarity).
To avoid initialization issues, we trained the network from scratch and used a very large number of iterations, $N_b=1000$.
In Fig.~\ref{fig:trajectoryNoHyst}, for two different bands, one from the visible range (top) and one outside it (bottom),
we plot the evolution of the loss components in the $(\LE,\L_S$) plane as a function of the number of iterations.
For the reasons explained before, for the spectral loss we consider the normalized value $\LE=\L_\lambda/\L^{\rm EXP}_\lambda$.
Each curve starts from the top-right point and evolves over time towards lower values of the losses. %\L_{\lambda/E}

For the ``visible'' band everything works quite well.
At the end of the process, $\L_S$ is always very low, which is to be expected given the very high correlation with the PAN,
and also the spectral quality is quite good, with $\LE$ going largely below 1 for some $(\alpha,\beta)$ pairs,
thus improving significantly over the reference value provided by the EXP method.
For the ``outside'' band the story is completely different.
$\L_S$ never approaches 0, and this was expected considering the much lower correlation of this band with the PAN.
However, also $\L_\lambda$ remains rather large and only in some cases it approaches the reference $\L^{\rm EXP}_\lambda$ level.
The loss trajectories are especially informative.
In all cases, $\L_S$ improves regularly as the process goes on but this seems to affect negatively $\L_\lambda$ which, after an initial improvement, impairs markedly and never recovers.
Contrary to what happens for the visible band, $\LE$ never goes below 1.

This behavior gave us the idea to use a more sophisticated training schedule,
where the balance between spectral and spatial loss components is adjusted dynamically during the process.
In particular, we decided to set $\beta=0$ at the beginning and fine-tune the network based only on the spectral loss, so as to bring it down to a desired level, say $\LE=\gLOW$.
Then we turn the $\L_S$ loss component back on, but always keeping the spectral quality under control.
In fact, as soon as $\LE$ exceeds a second higher threshold, $\gHIGH$,
the spatial loss component is turned off again, and $\LE$ is brought again to the lower level.
Our intuition was that this optimization cycle could bring the curve to a new point in the trajectory, with the same spectral loss as before but a lower spatial loss.
Fig.~\ref{fig:trajectory} shows that this is actually the case.
In this figure, we plot the same curves as before but organized differently.
Each chart now refers to a different $(\alpha,\beta)$ pair and compares the $(\LE,\L_S)$ trajectories
obtained with flat training and with the proposed on-off schedule.
For the visible band, the proposed schedule does not really improve over the baseline (the mechanism could be even switched off).
On the contrary, for the outside band the gain is striking.
The desired level of spectral distortion is readily achieved with all $(\alpha,\beta)$ pairs.
Then $\L_S$ is gradually reduced while $\L_\lambda$ keeps oscillating in the desired range.
The final value of $\L_S$ is somewhat larger than with flat training,
but this is quite reasonable for an ``outside'' band, intrinsically less correlated with the PAN, and does not correspond to a worse spatial quality.
A further interesting property emerging from this analysis is the remarkable stability of the trajectories,
almost identical to one another as the hyperparameters change, which suggests a good robustness with respect to changing sources, namely, a good generalization ability.
Finally, note that most of the final iterations are used to gain only minor improvements in $\L_S$, and tuning could be safely stopped much earlier.
This last observation allowed us to define a computationally efficient stopping rule.
For the current band, $b$, we count both
the total number of iterations $N_b$ and
the number of iterations with active spatial loss ($\beta>0$), $N_{b|S}$.
Tuning stops when either one exceeds a predefined limit, whose setting is discussed in next Section.
Note that also the value of $\beta$ to be used when the spatial loss is active is not fixed once and for all but only after a suitable warm-up.
A large value, $\beta_0$, is used initially, but this is repeatedly halved if it causes an excessive increase in spectral loss, down to a minimum of $\beta_0/8$.
This mechanism relieves us from the burden of selecting a correct value for $\beta$ which
on the other hand can change from band to band.
Even so, a few parameters remain to be set, a problem that will be analyzed in some detail in the following Section.

\begin{figure}
\small
\centering
\begin{tabular}{r}
\includegraphics[width=0.94\columnwidth]{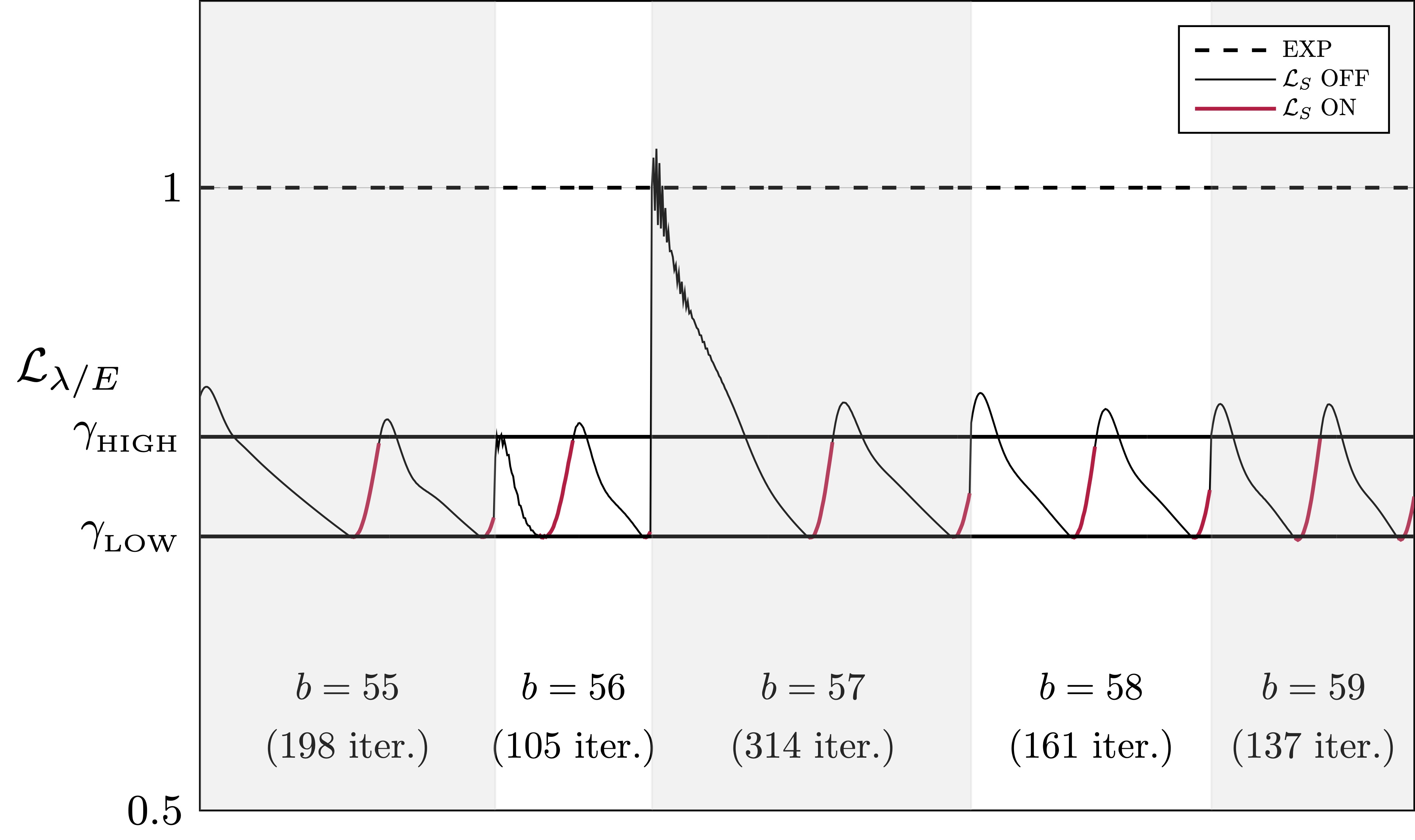}\\
\multicolumn{1}{c}{(a)}\\[2mm]
\includegraphics[width=0.91\columnwidth]{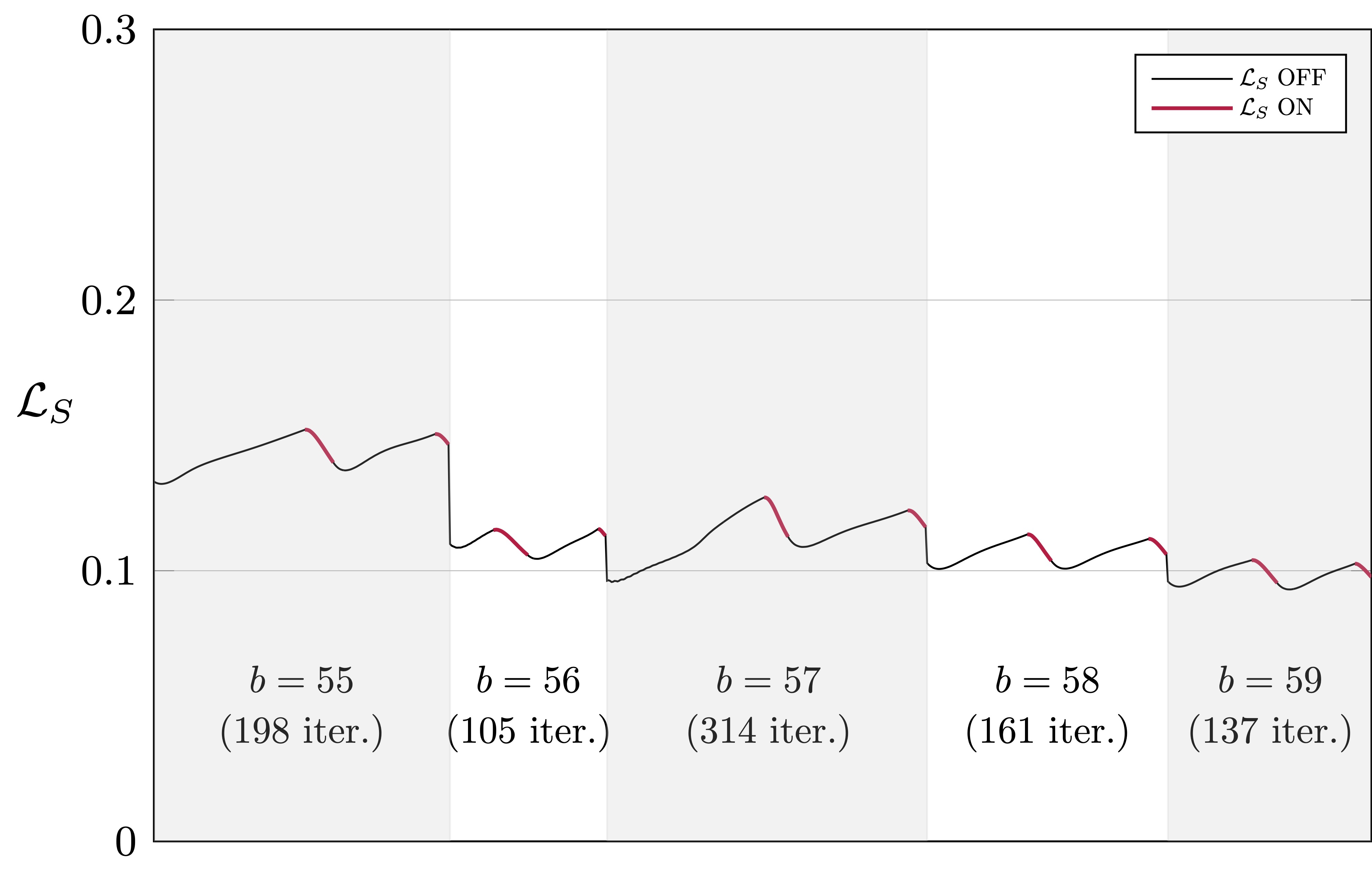}\\
\multicolumn{1}{c}{(b)}
\end{tabular}
\caption{Close-up (bands 55 to 59) of the evolution of spectral (a) and spatial (b) loss terms during the hysteresis-based tuning with model transfer.}
\label{fig:hysteresis}
\end{figure}

Fig.~\ref{fig:hysteresis} illustrates an example of tuning with hysteresis for a few bands of a sample image.
Parts (a) and (b) show the evolution of the (normalized) spectral and, respectively, spatial loss components as the tuning proceeds.
Red lines highlight ON phases, where both loss terms are active.
In this example we have fixed $\gLOW=0.72$ and $\gHIGH=0.8$.
Several interesting observations can be made:
{\it   i)}  most of the times, the model optimized for the previous band works very well on the new one, {\em e.g.} 55$\to$56 or 58$\to$59.
            In some cases, {\em e.g.} 56$\to$57, the transfer is less effective which impacts significantly on quality
            and calls for a tuning overhead to bring the spectral loss within the hysteresis limits.
            This latter case, however, occurs rarely, typically when going from the last band of a correlation block to the first band of another one.
{\it  ii)}  different average levels of spatial loss are observed for different bands, reflecting their different relationship with the PAN.
{\it iii)}  boosting sharpness ($\L_S$ ON) causes an increase of the spectral loss $\L_\lambda$
            and the need for multiple hysteresis cycles to keep it within the desired limits.

Given the oscillatory behavior of both losses, and some overshooting due to momentum,
the last model provided by the fine-tuning procedure is not necessarily the best performing one.
Therefore, in our implementation,
we keep memory of all of them and output the model with the lowest spatial loss among those that satisfy the constraints on the spectral loss.

Algorithm \ref{alg:rho} describes the proposed method more formally by means of a high-level pseudo-code.

\begin{center}
\begin{minipage}[c]{0.95\linewidth}
\begin{algorithm}[H]%[ht!]
\setstretch{1.15}
\footnotesize
\textbf{Input}: $\P$, $\H$, net (network structure)\\
\textbf{Hyperparameters}: $\gHIGH$, $\gLOW$, $\beta_0$, $\eta$, $N_0$, $N_S^{\rm max}$, $\epsilon$\\
\textbf{Output}: $\wh{\H}$
\begin{algorithmic}[1]
\State $B \gets {\rm size}(\H, 3)$                                                                  \Comment{\# of bands}
\State $\mathbf{c} \gets {\rm correlations}(\H)$                                                    \Comment{$\mathbf{c}_b = {\rm corrcoef}(\H_b, \H_{b-1})$; $\mathbf{c}_1 = 0$}
\State $\phi \gets {\rm weightInit}({\rm net})$                                                     \Comment{net weights init.}
\For{$b\gets 1:B$}
    \State $N_b^{\rm max} \gets {\rm iterationUpperBound}(\mathbf{c}, N_0, B, \eta)$                \Comment{Eqs. (\ref{eqn:Nbm}), (\ref{eqn:deltaN})}
    \State $\wt{H}_b \gets {\rm interpolator}(\H_b)$
    \State $\mathcal{L}_{\lambda}^{\rm EXP} \gets {\rm getSpectralLoss}(\wt{\H}_b,\H_b)$
    \State $N_b, N_{b|S} \gets 0$                                                                   \Comment{counter init.}
    \State $\beta \gets {\rm warmUp}(\H_b, \P, {\rm net}, \phi, \beta_0, \epsilon, \alpha)$         \Comment{$\beta$ warm-up}
    \State ${\rm ON} \gets \mathbf{\rm false}$                                                      \Comment{$\L_S$ flag init.}
    \While{ $ \{N_{b|S} < N_{S}^{\rm max}\} \;\mbox{\bf and}\; \{N_b < N^{\rm max}_b$\} }           \Comment{Eq. (\ref{eq:max_iterations}) }
        \State $\phi, \mathcal{L}_{\lambda}, \mathcal{L}_S \gets {\rm optimizationStep}
                                                    (\H_b,\P,{\rm net},\phi,\beta,{\rm ON})$
        \State $N_b\gets N_b+1$
        \If{ON}
            \State $N_{b|S} \gets N_{b|S} + 1$
        \EndIf
        \If{$\LE > \gHIGH$}
            \State ON $\gets \mathbf{\rm false}$                                                    \Comment{switch-off}
        \ElsIf {$\LE < \gLOW$}
            \State ON $\gets \mathbf{\rm true}$                                                     \Comment{switch-on}
        \EndIf
    \EndWhile
    \State $\wh{H}_b \gets {\rm predict}(\H_b, \P, {\rm net}, \phi)$                                \Comment{inference on band $b$}
\EndFor
\State {\bf return} $\wh{H}$
\end{algorithmic}
\caption{$\rho$-PNN pseudo-code.}
\label{alg:rho}
\end{algorithm}
\end{minipage}
\end{center}

\section{Experimental Analysis}
\label{sec:results}
\renewcommand{\top}[1]{\textbf{{\color{green!60!black} #1}}}
\newcommand{\best}[1]{{\color{green!60!black} #1}}
\newcommand{\worst}[1]{{\color{red!90!black} #1}}

In this Section, we perform the comparative performance assessment of the proposed method.
First, we set some key parameters of the method based on preliminary experiments on the validation set of the benchmarking toolbox \cite{Ciotola2024}.
Then, after discussing the assessment methodology,
we carry out an objective performance analysis on reduced-resolution and full-resolution test data of the same toolbox,
and finally perform visual analysis of results.

\subsection{Parameter Setting}

\begin{figure}
\centering
\includegraphics[width=0.48\textwidth]{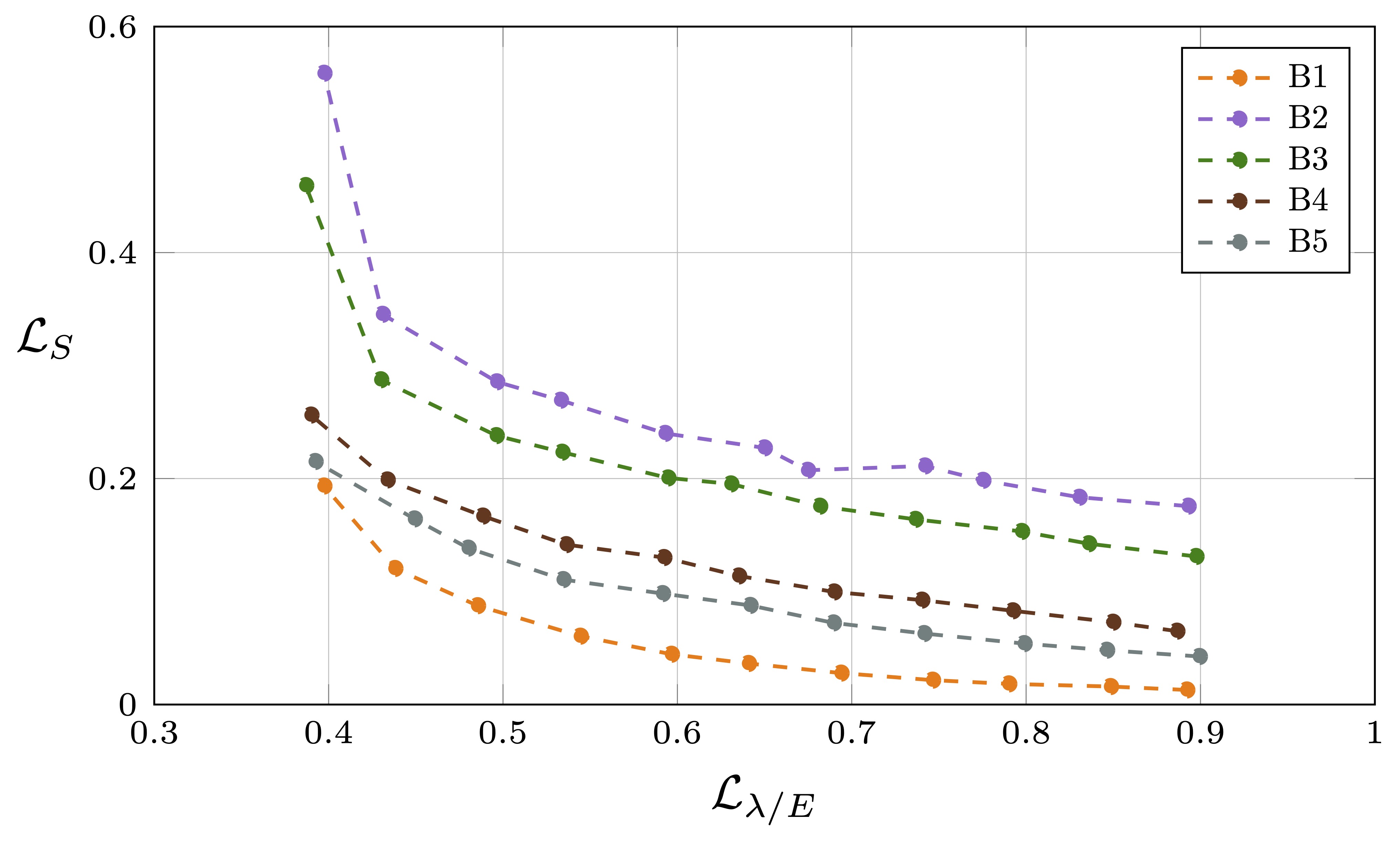}
\caption{Loss points in the $(\LE,\L_S)$-plane for various bands after thorough hysteresis-based tuning as a function of the $\gHIGH$ and $\gLOW$ parameters.}
\label{fig:gamma}
\end{figure}

\subsubsection{Setting the Target for Spectral Loss}
a key question for our approach is: to what extent can the spectral distortion be reduced without causing appreciable spatial distortion?
%More concretely, how low can $\gLOW$ be?
To answer this question, we carried out a set of experiments on five bands \{B1, \dots, B5\} of the validation dataset,
sampled uniformly from the whole spectral range.
For each band, we trained the model from scratch (no cross-band model transfer) with our hysteresis-based schedule, for virtually infinite iterations,
using always $\alpha=10^{-5}$ and $\beta=0.5$, but varying $\gHIGH$ in $\{0.40, 0.45, \ldots, 0.90\}$, with $\gLOW=0.9\gHIGH$.

In Fig.~\ref{fig:gamma} we show the final loss values achieved as points in the $(\LE,\L_S)$ plane.
For all bands we observe a similar behavior.
The lowest spatial loss is obtained for $\gHIGH=0.9$ and differs from band to band
depending on its noise level and especially its similarity with the PAN.
For bands in the visible spectrum, {\em e.g.} B1, with very high local correlation with the PAN, $\L_S$ goes almost to 0 (0.03),
while it is much higher for other bands, {\em e.g.} B2, from the near-IR range, with $\L_S$ close to 0.2.
In all cases,
the spatial loss increases as $\gHIGH$ decreases, as expected, but just slowly at the beginning,
and a steep increase is observed only when $\LE$ falls below 0.5, that is, $\L_\lambda$ is 50\% of the reference $\L^{\rm EXP}_\lambda$ value.
Based on this consistent behavior, we decided to set $\gHIGH=0.65$ and $\gLOW=0.59$ in the upcoming experiments.

\subsubsection{Limiting the Number of Iterations}

\begin{figure}
\centering
\includegraphics[width=0.45\textwidth]{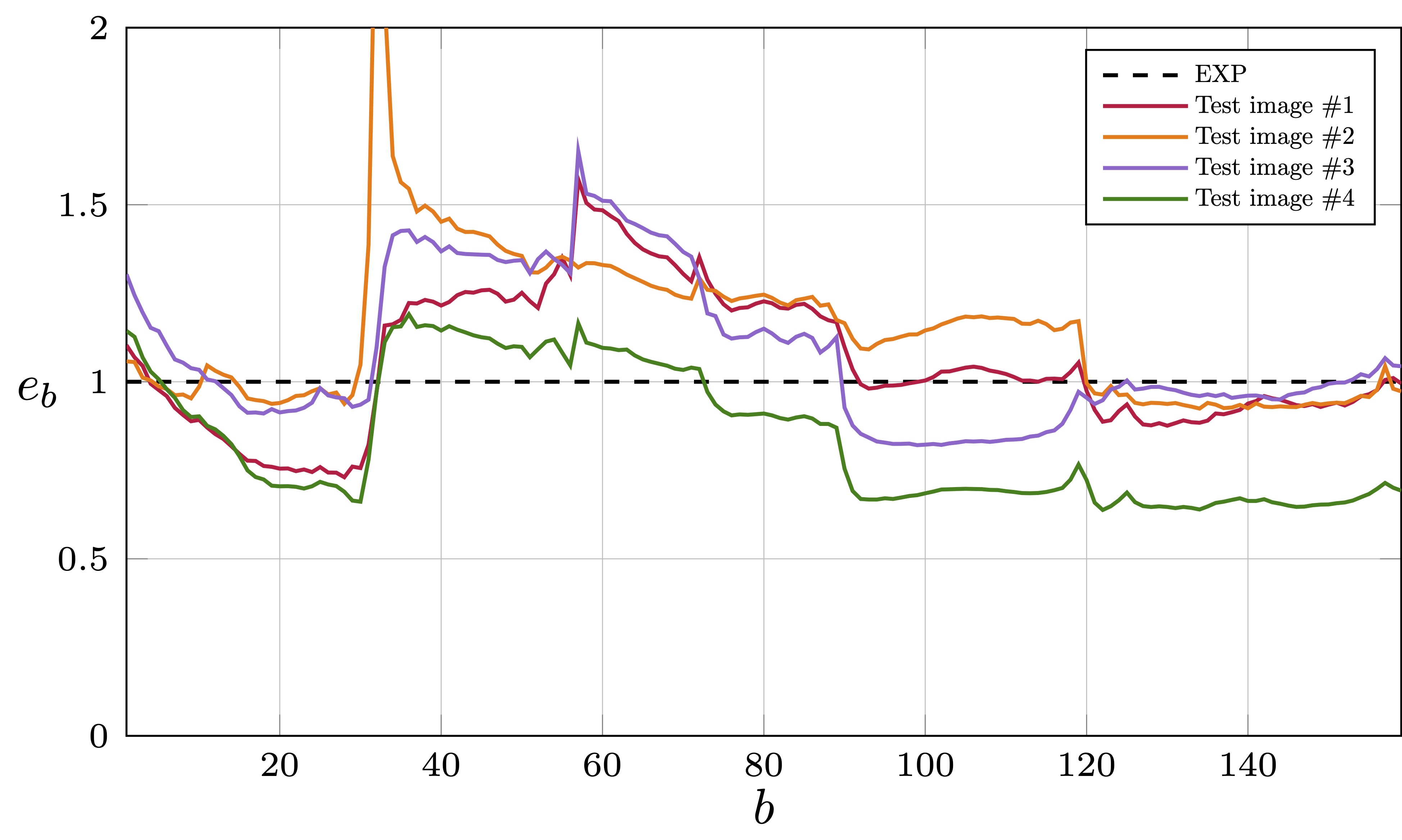}
\caption{R-PNN normalized reprojection error over four PRISMA images.}
\label{fig:mae norm rpnn}
\end{figure}

besides ensuring high and consistent quality, we want to keep complexity under control by limiting the number of tuning iterations.
However, as also shown by Fig.~\ref{fig:hysteresis},
the previous-band model is not always a good starting point, in which case many iterations are required to bring the loss down to acceptable values.
This appears clearly in Fig.~\ref{fig:mae norm rpnn} which shows the normalized reprojection error of R-PNN over four PRISMA images.
The error spikes over the first band of a correlation block, then stabilizes at lower values in subsequent bands due to the high inter-band similarity.
Based on this observation, we decided to stop tuning when either one of two conditions holds:
\begin{equation}
    \{ N_{b|S} = N_{b|S}^{\rm max} \} \;\; \mbox{OR}\;\; \{ N_b =  N^{\rm max}_b \},
    \label{eq:max_iterations}
\end{equation}
Most of the times, the spectral loss reaches quickly the desired value, and most iterations involve also the spatial loss.
In these cases, the active constraint is on $N_{b|S}$, and we set the maximum to a fixed value for all bands, $N_{b|S}^{\rm max}=N_{S}^{\rm max}$.
More rarely, when subsequent bands are weakly correlated, reducing the spectral loss is more difficult,
and a limit on the total number of iterations $N^{\rm max}_b$ becomes necessary to prevent an inordinate number of them.
This latter limit, however, must be carefully set to make sure that a reasonable number of iterations involve also the spatial loss.
Therefore, we decided to make this limit depend on inter-band correlation, that is
\renewcommand{\c}{\mathbf{c}}
\begin{equation}
	\label{eqn:Nbm}
    N^{\rm max}_b = N_0 + \Delta N (1-\c_b)
\end{equation}
with $c_b$ the correlation coefficient between bands $b$ and $b-1$.
Therefore, the maximum number of iterations is
$N_0\geq N_{S}^{\rm max}$
%(as a rule of thumb it can be set $N_0\in [2N^{\rm max}_S, 5N^{\rm max}_S]$)
in case of perfect correlation ($\c_b=1$) and $N_0+\Delta N$ in case of orthogonal bands ($\c_b=0$).
With this rule,
we can easily set an overall bound to the total number of iterations $N$ as follows
%the total number of iterations cannot exceed the bound
\begin{equation}
    N=\sum_b N_b \leq N^{\rm max} = \sum_b N^{\rm max}_b = (1+\eta)BN_0
%    N^{\rm max} = \sum_b N^{\rm max}_b = B[N_0+\Delta N (1-\langle c_b\rangle)]
\end{equation}
being $\eta$ a suitable overhead multiplier to be set by the user,
%where $\langle c_b\rangle$ is the average inter-band correlation,
and $B$ the number of bands.
Therefore, one can set $N^{\rm max}$ in advance based on complexity issues, and $N_0$ and $\eta$ accordingly.
Then, $\Delta N$ will be given by the (easy to proof) formula
\begin{equation}
\label{eqn:deltaN}
\Delta N = \frac{\eta B N_0}{B-\sum_b \c_b}
\end{equation}

\begin{figure}
\centering
\includegraphics[width=0.45\textwidth]{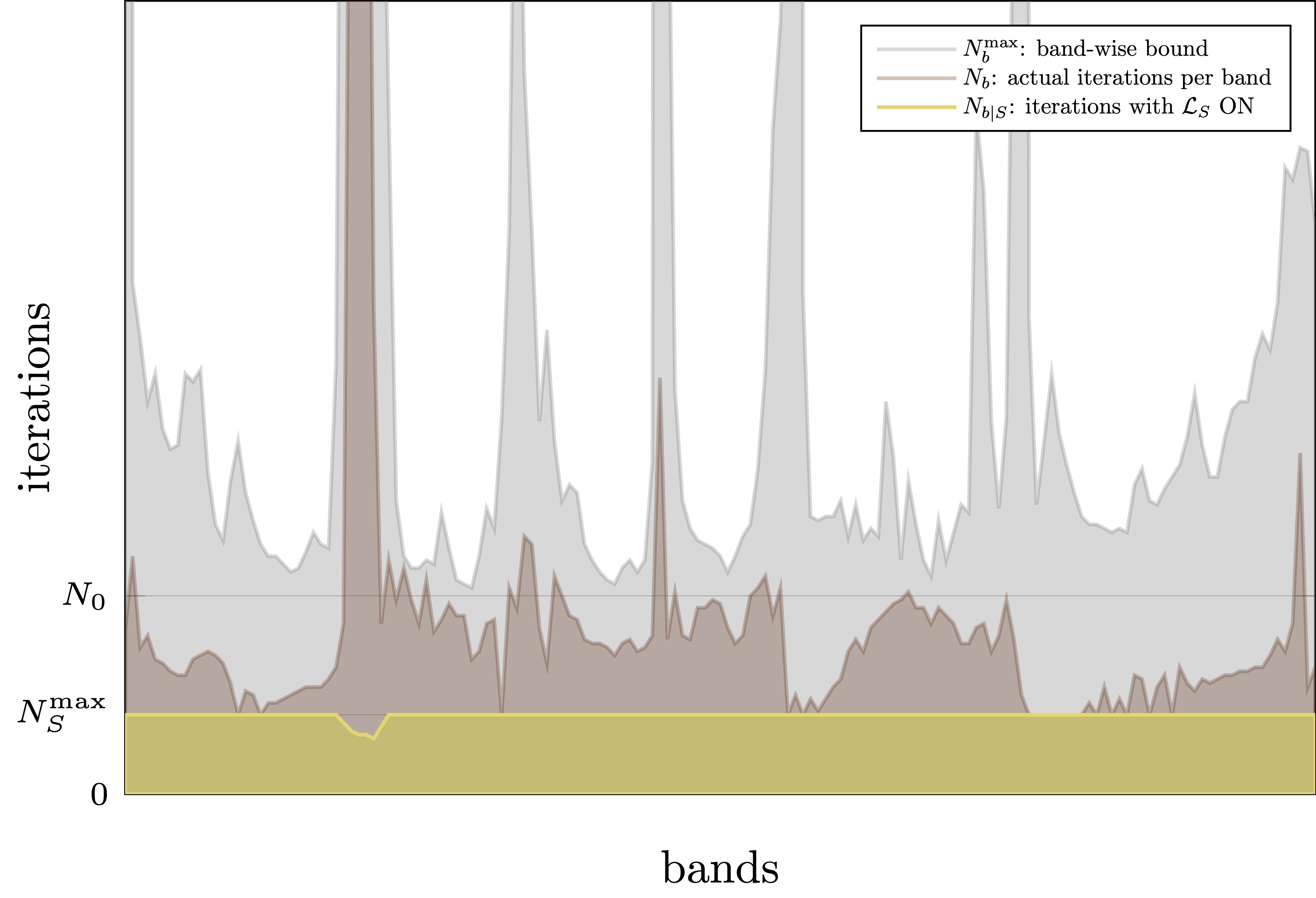}
\caption{
Distribution of tuning iterations for a test PRISMA image.
On average, 47 iterations/band are used, almost always 20 of them with spatial loss ON.}
\label{fig:iterations}
\end{figure}

Fig.~\ref{fig:iterations} shows an example of how iterations distribute over the various bands of a PRISMA image.
Here we set $N_{S}^{\rm max}=20, N_0=50$ and $\eta=8$.
$N_{b|S}$ is shown in yellow and $N_b$ in brown, with its upper bound $N^{\rm max}_b$ in light gray.
Although a maximum of $(1+\eta)N_0=450$ iterations per band is allowed on average, only 47 are used on the average.
Most of the times, tuning stops because $N_{b|S}$ reaches its maximum, $N_{S}^{\rm max}=20$.
The total number of iterations rarely exceeds $N_0=50$
and only for a group of bands, on the boundary between the visible and NIR ranges, the $N_b=N_b^{\rm max}$ exit condition is activated.

Tab.~\ref{tab:hyperparameters} summarizes
the most relevant hyperparameters of the proposed method, their selected values, and their meaning.

\begin{table}
\caption{Main hyperparameters of the proposed method.}
\footnotesize
\centering
\setlength{\tabcolsep}{2pt}
\label{tab:hyperparameters}
\begin{tabular}{ccp{6cm}}
\hline
\ru \bf Param           & \bf Value & \bf Meaning \\ \hline
\ru     $\gHIGH$        &  0.65 & Upper limit for $\LE$ \\
        $\gLOW$         &  0.59 & Lower limit for $\LE$       \\
        $\beta_0$       &     2 & Initial $\beta$, weight of the spatial loss term \\
        $\epsilon$      & 0.007 & Growth factor of $\L_\lambda$ for $\beta$ adaptation \\
        $N_S^{\rm max}$ &    20 & Target \# of tuning iter.s with active spatial loss \\
        $N_0$           &    80 & Max    \# of tuning iter.s when $\c_b=1$ \\
        $\eta$          &    30 & Global param. controlling total \# of tuning iter.s \\ \hline
\end{tabular}
\end{table}

\subsection{Assessment Methodology}
To assess the quality of the proposed solution we take advantage of the Benchmarking toolbox \cite{Ciotola2024}
which provides
a publicly available set of rich and diverse PAN-HS PRISMA images,
a large set of SotA comparative methods,
and several performance evaluation tools based on the most credited assessment indexes.

\begin{table}
\caption{PAN-MS PRISMA benchmark datasets for testing.}
\centering
%\scriptsize
\footnotesize
\setlength{\tabcolsep}{1mm}
\begin{tabular}{lll}
%\hline
\hline
Date & Location & Land covers\\
\hline
05/09/22 & Cagliari, Italy &  Roofs, Streets, Crops, Hills, Water\\
24/08/23 & Udine, Italy &  Roofs, Streets, Crops, Water\\
08/09/23 & Ford County,  Kansas & Streets, Crops\\
20/11/23 & Macuspana, Tabasco & Streets, Crops \\
\hline
\multicolumn{3}{l}{RR test image size: 600$\times$600 (from a 3600$\times$3600 real FR image)}\\
\multicolumn{3}{l}{Real FR test image size: 1200$\times$1200}\\
\multicolumn{3}{l}{Resolution ratio $R=6$}\\
\multicolumn{3}{l}{Bands $B=159/239$}\\
\multicolumn{3}{l}{Resolution: 30 m (RR) or 5 m (FR)}\\
\hline
\end{tabular}
\label{tab:prisma}
\end{table}

PRISMA satellites are equipped with
a sensor operating in the 400--700~nm spectral range that provides a panchromatic image with 5~m spatial resolution,
and two hyperspectral sensors that acquire 66 bands at VNIR wavelengths, 400--1010~nm, and 173 bands at SWIR wavelengths, 920--2505~nm,
with 30~m resolution.
The dataset released by \cite{Ciotola2024} includes 16 images (12 for training and validation, 4 for testing)
comprising 159 spectral bands after removing noisy and uninformative channels.
For our experiments we use the four test images of the Toolbox, see Tab.~\ref{tab:prisma}.
From each original PRISMA PAN-HS pair (PAN/HS size: 3600$\times$3600/600$\times$600)
two test datasets were created:
a FR one (PAN/HS size: 1200$\times$1200/200$\times$200) obtained by cropping;
a RR one (PAN/HS/GT size 600$\times$600/100$\times$100/600$\times$600) obtained by resolution downgrading of the whole PAN-HS pair, with the original FR HS playing the role of GT.

Tab.~\ref{tab:methods} gathers the comparative methods
which can be roughly grouped in four categories: component substitution (CS),
multiresolution analysis (MRA), model-based optimization (MBO), and deep learning (DL).
The interested Reader is referred to \cite{Ciotola2024} for details on these methods.

\begin{table}
\caption{HS pansharpening methods.}
\footnotesize
%\scriptsize
\centering
\setlength{\tabcolsep}{2pt}
\begin{tabular}{lcp{5.5cm}}%{p{2cm}p{3cm}p{1.5cm}}%{p{8.4cm}}
\hline
\bf \ru Name & \bf Ref  & \bf Summary\\
\hline
\hline
\ru EXP                 &    & Approximation of the ideal interpolator \\
\hline
\multicolumn{3}{c}{\bf \ru Component Substitution (CS) methods} \\[1mm]
%GS                 & \cite{Laben2000}    & Gram-Schmidt \\
GSA                 & \cite{Aiazzi2007}   & Gram-Schmidt adaptive component substitution \\
BT-H                & \cite{Lolli2017}    & Brovey transform with haze correction \\
%BDSD               & \cite{Garzelli2008} & Band-dependent spatial detail injection\\
BDSD-PC             & \cite{Vivone2019}   & Band-dependent spatial detail injection with physical constraint\\
PRACS               & \cite{Choi2011}     & Partial replacement adaptive CS \\ \hline
\multicolumn{3}{c}{\bf \ru Multiresolution Analysis (MRA) methods} \\[1mm]
%MTF-GLP            & \cite{Aiazzi2006}   & MTF-matched Generalized Laplacian Pyramid \\
MTF-GLP-FS          & \cite{Vivone2018a}  & Modulation Transfer Function (MTF)-matched Generalized Laplacian Pyramid (MTF-GLP) with fusion rule at full scale \\
MTF-GLP-HPM         & \cite{Alparone2017} & MTF-GLP with high pass modulation\\
% MTF-GLP-HPM-H     &  \cite{Lolli2017}   & MTF-GLP-HPM with haze correction\\
MTF-GLP-HPM-R       & \cite{Vivone2017}   & MTF-GLP-HPM with regression-based spectral matching\\
AWLP                & \cite{Otazu2005}    & Additive wavelet luminance proportional \\
MF                  & \cite{Restaino2016} & Nonlinear decomposition with morphological filters \\ \hline
\multicolumn{3}{c}{\bf \ru Model-Based Optimization (MBO) methods}\\[1mm]
% Bayesian Naive    & \cite{Wei2015b}     & Bayesian estimation with naive independent Gaussian prior \\ %\cite{Wei201},
% Bayesian Sparse   & \cite{Wei2015a}     & Bayesian estimation with sparse representations-based prior \\
HySURE              & \cite{Simoes2015}   & Bayesian estimation with vector total variation prior \\
SR-D                & \cite{Vicinanza2015}& Sparse representations-based detail injection \\
TV                  & \cite{Palsson2014}  & Total variation-based pansharpening \\ \hline
\multicolumn{3}{c}{\bf \ru Deep Learning (DL) methods} \\[1mm]
HyperPNN            & \cite{He2019a}      & 7-layer net with spectral encoder-decoder structure\\
HSpeNet             & \cite{He2020}       & Advanced version of HyperPNN, with deeper architecture and Spectral Angle Mapper (SAM) loss term\\
DHP-DARN            & \cite{Zheng2020}    & Deep residual channel-spatial attention net with Deep Image Prior (DIP) for HS resizing\\
DIP-HyperKite       &  \cite{Bandara2022} & Overcomplete net with DIP for HS resizing\\
HyperDSNet          & \cite{Zhuo2022}     & Spectral attention-based detail injection from deep-shallow features \\
R-PNN (unsup.)      & \cite{Guarino2023}  & Band-wise pansharpening using modified {Z-PNN} \cite{Ciotola2022} with tuning propagation \\
PCA-Z-PNN (unsup.)  & \cite{Guarino2023a} & Z-PNN model with PCA-based input reduction\\
\hline
\end{tabular}
\label{tab:methods}
\end{table}

The assessment follows the well-established Wald's protocol \cite{Wald1997} which considers two contexts:
a reduced resolution framework, where suitably degraded (in resolution) datasets are used for assessment {\em with} GT (where the ground truth is given by the original full resolution HS image),
and a full resolution framework for assessment {\em without} GT but using the original data as input.
Accordingly, the quality indexes, gathered in Tab.~\ref{tab:indexes},
are split in two groups for reduced resolution (RR) and full resolution (FR) assessment.

As well known, the assessment of spatial quality in the full resolution context (hence, without a GT) is still an open and challenging problem.
Therefore, to obtain a more comprehensive evaluation of performance,
besides the $\DS$ index used in the toolbox \cite{Ciotola2024} we provide also the correlation-based index $\DR$ proposed in \cite{Scarpa2022}.
Even so, no numerical index or combination of indexes provides a fully reliable measure of spatial quality,
and we will resort to visual inspection of sample results to get further indications on spatial quality.

\begin{table}
\caption{HS pansharpening quality assessment indexes.}
\centering
\footnotesize
\setlength{\tabcolsep}{2pt}
\begin{tabular}{rl} \hline
\multicolumn{2}{c}{\bf \ru RR assessment}\\ \hline
\ru
ERGAS                  & {\em Erreur Relative Globale Adimensionnelle de Synth{\'e}se} \cite{Wald2002} \\
SAM                   & Spectral Angle Mapper \cite{Yuhas1992}  \\ %: average divergence (cosine) between predicted and reference spectral response.\\
%$Q$ \cite{Wang2002}                   & Universal Image Quality Index\\
$Q2^n$             & Multiband extension \cite{Garzelli2009} of Universal Image Quality Index~\cite{Wang2002}\\ \hline \hline
\multicolumn{2}{c}{\bf \ru  FR assessment}\\ \hline
\ru
$\DL$      & Khan's spectral distortion index \cite{Arienzo2022, Khan2009}\\
%$\DS$ \cite{Alparone2008}             & Spatial distortion index\\
$\DS$ & Spatial distortion index \cite{Arienzo2022, Alparone2018}\\
%$\DSR$ \cite{Arienzo2022}             & A variant of $\DS$\\
$\DR$   & Correlation-based spatial distortion index \cite{Scarpa2022}\\
%RQNR     & Regression-based QNR index \cite{Arienzo2022, Vivone2023}\\
\hline
\end{tabular}
\label{tab:indexes}
\end{table}

\subsection{Reduced-resolution Quality Assessment}
The numerical results obtained for the four RR datasets are summarized in Tab.~\ref{tab:RR}.
For each index, for each dataset and on average (Avg.), the top 5 and the worst 5 methods are highlighted in green and red, respectively.
The best one is in bold green.
Before analyzing the results, focusing on DL methods,
it is important to recall that supervised models are optimized for the RR case
while unsupervised ones (R-PNN, PCA-Z-PNN, and proposed) are optimized for FR real data.
Therefore, it is not surprising that several supervised techniques ({\em e.g.}, HyperPNN, HSpeNet and especially Hyper-DSNet) are on top.
However, despite its unsupervised approach, the proposed solution also performs quite well.
More specifically,
it presents a good spectral behavior (SAM is the most appropriate index in this respect)
and also a very good capacity to preserve spatial structures ($Q2^n$ is an index oriented to assess spatial correlation and contrast).
It looks less competitive in terms of ERGAS, a band-wise normalized variant of the root mean square error, but still close to the top 5.

\begin{table*}
\caption{Results at Reduced Resolution. \top{Best score} is in \top{bold green}, \best{Top-5} in \best{green},  and \worst{Worst-5} in \worst{red}.
%The average score is in the rightmost column.
}
\footnotesize
\centering
\setlength{\tabcolsep}{2pt}
\begin{tabular}{lc@{\rule{2mm}{0mm}}cccccc@{\rule{2mm}{0mm}}cccccc@{\rule{2mm}{0mm}}cccccc} \hline
\ru Method	&	&		\multicolumn{5}{c}{ERGAS}																		&	&		\multicolumn{5}{c}{SAM}																		&	&		\multicolumn{5}{c}{Q$2^n$}																		 \\	\hline
\ru	&	&		Cagliari		&		Udine		&		Ford		&		Tabasco		&		Avg.		&	&		Cagliari		&		Udine		&		Ford		&		Tabasco		&		Avg.		&	&		Cagliari		&		Udine		&		Ford		&		Tabasco		&		Avg.		 \\	\cline{1-1} \cline{3-7} \cline{9-13} \cline{15-19}
\ru (Ideal)	&	&		0		&		0		&		0		&		0	&		0	&	&		0		&		0		&		0		&		0		&		0	&	&		1		&		1		&		1		&		1		&		1		 \\	\cline{1-1} \cline{3-7} \cline{9-13} \cline{15-19}
\ru EXP	&	&		1.7716		&		3.9587		&		1.6921		&		4.3650		&		2.9468		&	&		2.3073		&		4.6288		&		2.8528		&		6.3524		&		4.0353		&	&		0.5971		&		0.6453		&		0.7984		&		0.5935		&		0.6586		 \\	\cline{1-1} \cline{3-7} \cline{9-13} \cline{15-19}
\ru GSA	&	&	\best{0.9389}	&		2.4243		&	\best{1.0008}	&		2.8065		&		1.7926		&	&	\best{1.8191}	&		3.5730		&	\best{2.2570}	&		5.7455		&	\best{3.3486}	&	&		0.8739		&		0.8110		&		0.9201		&		0.7955		&		0.8501		 \\	
\ru BT-H	&	&	\worst{1.2784}	&		3.4898		&		1.2462		&		3.1015		&		2.2790		&	&		2.3857		&		4.6059		&		2.5563		&		6.3532		&		3.9753		&	&	\worst{0.8180}	&		0.7166		&		0.8941		&		0.7837		&	\worst{0.8031}	 \\	
\ru BDSD-PC	&	&		1.1664		&		2.7627		&		1.2637		&	\worst{3.9316}	&		2.2811		&	&		1.9439		&		5.0116		&		2.6646		&	\worst{7.8908}	&		4.3777		&	&		0.8524		&		0.7904		&		0.9048		&	\worst{0.7144}	&		0.8155		 \\	
\ru PRACS	&	&		1.1803		&		3.4688		&		1.2773		&		3.8578		&		2.4461		&	&		1.9742		&		4.2221		&		2.4571		&		6.2161		&		3.7174		&	&	\worst{0.8166}	&	\worst{0.7144}	&	\worst{0.8804}	&	\worst{0.6778}	&	\worst{0.7723}	 \\	
\cline{1-1} \cline{3-7} \cline{9-13} \cline{15-19}
\ru MTF-GLP-FS	&	&	\best{0.9239}	&		2.3660		&	\best{0.9808}	&	\best{2.7356}	&	\best{1.7516}	&	&	\best{1.8123}	&	\best{3.4496}	&	\best{2.2297}	&	\best{5.7208}	&	\top{3.3031}	&	&	\best{0.8792}	&		0.8218		&	\best{0.9238}	&		0.8026		&		0.8568		 \\	
\ru MTF-GLP-HPM	&	&	\best{0.9292}	&		4.0148		&		1.2231		&		3.0609		&		2.3070		&	&	\best{1.8555}	&	\worst{5.1824}	&		2.6120		&		6.4104		&		4.0151		&	&	\best{0.8833}	&	\worst{0.6698}	&		0.9000		&		0.7890		&		0.8105		 \\	
\ru MTF-GLP-HPM-R	&	&	\top{0.9114}	&		2.4661		&	\top{0.9743}	&	\best{2.6967}	&	\best{1.7621}	&	&	\top{1.8121}	&	\best{3.3953}	&	\best{2.2124}	&		6.0986		&	\best{3.3796}	&	&	\best{0.8800}	&		0.8134		&	\best{0.9239}	&		0.8023		&		0.8549		 \\	
\ru AWLP	&	&		1.1846		&	\worst{4.6926}	&		1.3424		&	\worst{3.8592}	&	\worst{2.7697}	&	&		2.3323		&	\worst{7.9280}	&	\worst{2.9252}	&		7.7668		&	\worst{5.2381}	&	&		0.8516		&	\worst{0.6290}	&		0.8832		&	\worst{0.7496}	&	\worst{0.7783}	 \\	
\ru MF	&	&		1.1392		&	\worst{	4.0155	}	&	\worst{1.5506}	&		3.3944		&	\worst{2.5249}	&	&		1.9213		&		4.5587		&		2.7173		&		6.2918		&		3.8723		&	&		0.8638		&		0.7263		&	\worst{0.8804}	&		0.7807		&		0.8128		 \\	\cline{1-1} \cline{3-7} \cline{9-13} \cline{15-19}
\ru HySURE	&	&	\worst{1.4664}	&	\worst{4.3476}	&	\worst{2.0749}	&	\worst{5.1119}	&	\worst{3.2502}	&	&	\worst{2.9264}	&	\worst{6.6950}	&	\worst{4.4153}	&	\worst{9.0468}	&	\worst{5.7709}	&	&	\worst{0.7943}	&	\worst{0.5761}	&	\worst{0.7475}	&	\worst{0.4951}	&	\worst{0.6532}	 \\	
\ru SR-D	&	&	\worst{1.8476}	&	\worst{4.2618}	&	\worst{1.8825}	&	\worst{4.8011}	&	\worst{3.1982}	&	&	\worst{2.4242}	&	\worst{6.1149}	&	\worst{3.0551}	&	\worst{7.9927}	&	\worst{4.8967}	&	&	\worst{0.5734}	&	\worst{0.5846}	&	\worst{0.7313}	&	\worst{0.5440}	&	\worst{0.6083}	 \\	
\ru TV	&	&		1.2624	&		2.8864		&		1.3163		&		3.1771		&		2.1605		&	&		2.2212		&		4.0400		&		2.5705		&	\best{5.6672}	&		3.6247		&	&	\worst{0.8291}	&		0.7833		&		0.8911		&		0.7692		&		0.8182		 \\	\cline{1-1} \cline{3-7} \cline{9-13} \cline{15-19}
\ru HyperPNN	&	&		1.2245		&	\best{2.2237}	&		1.1864		&		3.3557		&		1.9976		&	&	\worst{2.8925}	&		4.3058		&		2.6461		&		7.0296		&		4.2185		&	&		0.8777		&	\best{0.8539}	&	\best{0.9247}	&	\best{0.8229}	&	\best{0.8698}	 \\	
\ru HSpeNet	&	&		1.0354		&	\best{2.0527}	&		1.0674		&		2.7964		&	\best{1.7379}	&	&		2.0644		&	\best{3.2157}	&		2.2838		&		5.8462		&	\best{3.3525}	&	&		0.8674		&	\best{0.8588}	&		0.9136		&	\best{0.8245}	&	\best{0.8660}	 \\	
\ru DHP-DARN	&	&	\worst{1.9137}	&	\worst{4.4280}	&	\worst{1.7389}	&	\worst{4.7617}	&	\worst{3.2106}	&	&	\worst{3.4673}	&		5.1740		&	\worst{3.3833}	&	\worst{9.1288}	&	\worst{5.2883}	&	&		0.8489		&		0.8234		&		0.9190		&		0.7934		&		0.8462		 \\	
\ru DIP-HyperKite	&	&	\worst{1.3510}	&	\best{2.0206}	&		1.0995		&		2.8092		&		1.8201		&	&	\worst{3.6543}	&	\worst{5.8268}	&		2.7089		&	\worst{9.1932}	&	\worst{5.3458}	&	&		0.8419		&	\best{0.8497}	&		0.9205		&	\best{0.8201}	&	\best{0.8581}	 \\	
\ru Hyper-DSNet	&	&	\best{0.9725}	&	\top{1.8198}	&	\best{0.9820}	&	\top{2.4237}	&	\top{1.5495}	&	&	\best{1.9162}	&	\top{2.9638}	&	\top{2.1906}	&		7.4854		&		3.6390		&	&	\top{0.8833}	&	\top{0.8767}	&	\top{0.9303}	&	\top{0.8426}	&	\top{0.8832}	 \\	
\ru R-PNN	&	&		0.9884		&	\best{2.2256}	&		1.0581		&	\best{2.6704}	&	\best{1.7356}	&	&		1.9720		&	3.5728 	&		2.3860		&	\best{5.6797}	&		3.4026		&	&		0.8712		&		0.8225		&		0.9172		&		0.8112		&		0.8555		 \\	
\ru PCA-Z-PNN	&	&		1.0824		&		2.9062		&	\worst{1.5575}	&	2.7813	&		2.0819		&	&		2.1124		&		4.6125		&	\worst{3.1154}	&	\best{5.5297}	&		3.8425		&	&		0.8507		&		0.7798		&	\worst{0.8647}	&		0.8116		&		0.8267		 \\	\cline{1-1} \cline{3-7} \cline{9-13} \cline{15-19}
\ru $\rho$-PNN	&	&		1.0109		&		2.3268		&	\best{1.0205}	&		\best{2.7308}		&		1.7722		&	&		1.9625		&		\best{3.5609}		&	\best{2.2439}	&	\top{5.3396}	&	\best{3.2767}	&	&	\best{0.8782}	&	\best{0.8279}	&	\best{0.9246}	&	\best{0.8177}	&	\best{0.8621}	 \\	\hline
\end{tabular}
\label{tab:RR}
\end{table*}

\begin{table*}
\caption{Results at Full Resolution. \top{Best score} is in \top{bold green}, \best{Top-5} in \best{green},  and \worst{Worst-5} in \worst{red}}.
\footnotesize
\centering
\setlength{\tabcolsep}{2pt}
\begin{tabular}{lc@{\rule{2mm}{0mm}}cccccc@{\rule{2mm}{0mm}}cccccc@{\rule{2mm}{0mm}}cccccc} \hline
\ru	Method	&	&		\multicolumn{5}{c}{$D_{\lambda}$}																		&	&		\multicolumn{5}{c}{$D_{S}$}																		&	&		\multicolumn{5}{c}{$D_{\rho}$}																		 \\	\hline
\ru		&	&		Cagliari		&		Udine		&		Ford		&		Tabasco		&		Avg.		&	&		Cagliari		&		Udine		&		Ford		&		Tabasco		&		Avg.		&	&		Cagliari		&		Udine		&		Ford		&		Tabasco		&		Avg.		 \\	\cline{1-1} \cline{3-7} \cline{9-13} \cline{15-19}
\ru	(Ideal)	&	&		0		&		0		&		0		&		0	&		0	&	&		0		&		0		&		0		&		0		&		0		&	&		0		&		0		&		0		&		0		&		0		 \\	\cline{1-1} \cline{3-7} \cline{9-13} \cline{15-19}	
\ru	EXP	&	&		0.0126		&		0.0077		&		0.0054		&		0.0106		&		0.0091		&	&		0.1306		&		0.1273		&		0.0704		&		0.1947		&		0.1308		&	&		0.7973		&		0.8839		&		0.7963		&		0.7972		&		0.8187		 \\	\cline{1-1} \cline{3-7} \cline{9-13} \cline{15-19}
\ru	GSA	&	&		0.0227		&		0.0134		&		0.0078		&		0.0122		&		0.0140		&	&		0.0056		&		0.0074		&		0.0077		&		0.0184		&		0.0098		&	&	\best{0.0593}	&	\worst{0.6009}	&		0.2075		&		0.0791		&		0.2367		 \\	
\ru	BT-H	&	&	\worst{0.0633}	&	\worst{0.0936}	&		0.0195		&	\worst{0.0288}	&	\worst{0.0513}	&	&	\top{0.0000}	&	\top{0.0000}	&	\top{0.0000}	&	\top{0.0002}	&	\top{0.0001}	&	&	\best{0.0579}	&	\best{0.0954}	&	\best{0.0544}	&	\best{0.0676}	&	\best{0.0688}	 \\	
\ru	BDSD-PC	&	&	\worst{0.0439}	&	\worst{0.0256}	&		0.0220		&	\worst{0.0487}	&	\worst{0.0351}	&	&	\best{0.0024}	&	\best{0.0000}	&	\best{0.0001}	&	\best{0.0003}	&	\best{0.0007}	&	&		0.1280		&		0.3878		&		0.1726		&		0.1192		&		0.2019		 \\	
\ru	PRACS	&	&	\best{0.0118}	&		0.0155		&		0.0047		&		0.0088		&		0.0102		&	&		0.0083		&		0.0163		&		0.0114		&		0.0245		&		0.0151		&	&	\worst{0.1775}	&		0.3252		&	\worst{0.3409}	&		0.1903		&		0.2585		 \\	
\cline{1-1} \cline{3-7} \cline{9-13} \cline{15-19}
\ru	MTF-GLP-FS	&	&	\best{0.0101}	&	\best{0.0054}	&	\best{0.0033}	&	\best{0.0065}	&	\best{0.0063}	&	&		0.0214		&		0.0277		&		0.0174		&		0.0322		&		0.0247		&	&		0.0618		&	\worst{0.6000}	&		0.2123		&		0.0831		&		0.2393		 \\	
\ru	MTF-GLP-HPM	&	&		0.0147		&		0.0087		&		0.0067		&		0.0093		&		0.0098		&	&		0.0215		&	\worst{0.0284}	&		0.0179		&	\worst{0.0389}	&	\worst{0.0267}	&	&	\best{0.0479}	&	\best{0.1022}	&	\best{0.0438}	&	\best{0.0538}	&	\best{0.0619}	 \\	
\ru	MTF-GLP-HPM-R	&	&	\best{0.0100}	&		0.0057		&	\best{0.0033}	&		0.0074		&	\best{0.0066}	&	&	\worst{0.0217}	&	\worst{0.0285}	&		0.0176		&		0.0374		&		0.0263		&	&		0.0669		&		0.6050		&		0.2164		&		0.0950		&		0.2458		 \\	
\ru	AWLP	&	&		0.0132		&		0.0097		&		0.0055		&		0.0092		&		0.0094		&	&	\worst{0.0251}	&	\worst{0.0318}	&	\worst{0.0206}	&	\worst{	0.0407	}	&	\worst{0.0295}	&	&		0.0695		&		0.1778		&	0.0754	&		0.0933		&		0.1040		 \\	
\ru	MF	&	&	\worst{0.0655}	&	\worst{0.0392}	&	\worst{0.0390}	&	\worst{0.0331}	&	\worst{0.0442}	&	&	\worst{0.0389}	&		0.0269		&	\worst{0.0250}	&	\worst{0.0664}	&	\worst{0.0393}	&	&	\top{0.0476}	&	\top{0.0904}	&	\top{0.0391}	&	\best{0.0585}	&	\top{0.0589}	 \\	\cline{1-1} \cline{3-7} \cline{9-13} \cline{15-19}
\ru	HySURE	&	&	\worst{0.1126}	&	\worst{0.0618}	&	\worst{0.0510}	&	\worst{0.1158}	&	\worst{0.0853}	&	&	\best{0.0022}	&	\best{0.0011}	&	\best{0.0016}	&	\best{0.0022}	&	\best{0.0018}	&	&		0.1414		&		0.2276		&		0.0990		&	\worst{0.2688}	&		0.1842		 \\	
\ru	SR-D	&	&		0.0128		&		0.0087		&		0.0055		&		0.0115		&		0.0096		&	&	\worst{0.1316}	&	\worst{0.1292}	&	\worst{0.0720}	&	\worst{0.1981}	&	\worst{0.1327}	&	&	\worst{0.8234}	&	\worst{0.9061}	&	\worst{0.8531}	&	\worst{0.8560}	&	\worst{0.8596}	 \\	
\ru	TV	&	&	\top{0.0042}	&	\best{0.0028}	&	\best{0.0021}	&	\top{0.0036}	&	\top{0.0032}	&	&	\worst{0.0586}	&	\worst{0.0524}	&	\worst{0.0358}	&	\worst{0.0813}	&	\worst{0.0570}	&	&	\worst{0.2590}	&		0.3294		&	\worst{0.2791}	&		0.2001		&		0.2669		 \\	\cline{1-1} \cline{3-7} \cline{9-13} \cline{15-19}
\ru	HyperPNN	&	&		0.0415		&		0.0139		&	\worst{0.0256}	&		0.0248		&		0.0264		&	&	\best{0.0036}	&		0.0063		&	\best{0.0040}	&	\best{0.0067}	&	\best{0.0051}	&	&		0.1310		&		0.5080		&	\worst{0.2419}	&	\worst{0.2625}	&		0.2859		 \\	
\ru	HSpeNet	&	&		0.0174		&		0.0084		&	\worst{0.0240}	&		0.0154		&		0.0163		&	&		0.0101		&		0.0160		&		0.0141		&		0.0236		&		0.0159		&	&		0.1439		&		0.5338		&		0.2348		&		0.2558		&	\worst{0.2921}	 \\	
\ru	DHP-DARN	&	&	\worst{0.0832}	&	\worst{0.0615}	&	\worst{0.0467}	&	\worst{0.0513}	&	\worst{0.0607}	&	&	\best{0.0030}	&	\best{0.0053}	&	\best{0.0031}	&		0.0130		&		0.0061		&	&		0.1494		&		0.4713		&		0.2106		&		0.2315		&		0.2657		 \\	
\ru	DIP-HyperKite	&	&		0.0208		&		0.0125		&		0.0148		&		0.0146		&		0.0157		&	&		0.0042		&	\best{0.0050}	&		0.0043		&	\best{0.0059}	&	\best{0.0048}	&	&	\worst{0.2952}	&		0.5270		&		0.2217		&	\worst{0.2765}	&	\worst{0.3301}	 \\	
\ru	Hyper-DSNet	&	&		0.0119		&	\best{0.0050}	&		0.0082		&	\best{0.0068}	&		0.0080		&	&		0.0200		&		0.0246		&	\worst{0.0195}	&		0.0346		&		0.0247		&	&	\worst{0.1957}	&		0.5336		&		0.2276		&	\worst{0.3353}	&	\worst{0.3230}	 \\	
\ru	R-PNN	&	&		0.0131		&		0.0091		&		0.0063		&	\best{0.0074}	&		0.0090		&	&		0.0168		&		0.0144		&		0.0129		&		0.0341		&		0.0195		&	&	\best{0.0511}	&	\best{0.1366}	&	\best{0.0532}	&	\top{0.0504}	&	\best{0.0728}	 \\	
\ru	PCA-Z-PNN	&	&		0.0129		&	\best{0.0041}	&	\best{0.0046}	&		0.0080		&	\best{0.0074}	&	&		0.0152		&		0.0116		&		0.0088		&		0.0257		&		0.0153		&	&		0.0959		&	\worst{0.6828}	&	\worst{0.7935}	&		0.0792		&	\worst{0.4128}	 \\	\cline{1-1} \cline{3-7} \cline{9-13} \cline{15-19}	
\ru	$\rho$-PNN	&	&	\best{0.0043}	&	\top{0.0027}	&	\top{0.0018}	&	\best{0.0043}	&	\best{0.0033}	&	&		0.0215		&		0.0168		&		0.0173		&		0.0380		&		0.0234		&	&		0.0790		&	\best{0.1302}	&		\best{0.0743}		&	\best{0.0532}	&	\best{0.0842}	 \\	\hline
\end{tabular}
\label{tab:FR}
\end{table*}

\subsection{Full-resolution quality assessment}
FR numerical results are gathered in Tab.~\ref{tab:FR} using the same highlighting scheme of Tab.~\ref{tab:RR}.
The first five columns give the spectral quality index, $\DL$, individually on the four test images and averaged on all of them.
The next two groups of columns, instead, measure spatial quality by means of two alternative indices, $\DS$ and $\DR$.
In terms of spectral quality, the proposed method provides top scores, on par with TV, and much better than all other methods, including its ancestor R-PNN.
This was our aim from the beginning, together with that of uniform quality across the whole spectrum, not catched by this index.

$\DL$ is a solid indicator of quality, as it is computed against a well defined and unambiguous reference.
Measuring spatial quality is much more challenging, which is why we consider two alternative indices.
In fact, neither of them is without flaws.
For example, $\DS$ measures the distance between the PAN and the best (closest to the PAN) linear combination of the pansharpened spectral bands \cite{Alparone2018}.
Therefore,
if one band (among hundreds) is an exact copy of the PAN, then $\DS=0$, no matter what the other bands look like.
This happens for several entries in Tab.~\ref{tab:FR} which, far from signaling perfect pansharpening, correspond typically to degenerate solutions.
While such cases are rarely seen with MS images, they occur more easily with HS images.
Similar warnings apply to the second indicator.
In fact, $\DR$ is defined as the average of the local correlation coefficient between PAN and the pansharpened HS bands \cite{Scarpa2022}.
From a physical perspective, this metric evaluates the extent to which a $\sigma{\times}\sigma$ patch from any band of $\wh{\H}$
can be linearly predicted from the corresponding patch in the PAN image.
Consequently, $\DR$ directly reflects the degree of spatial alignment between the pansharpened product $\wh{\H}$ and the PAN image $P$ at fine spatial scales.
Therefore, it vanishes only when there is perfect correlation between all bands and the PAN.
However, this is not desirable for bands that have no spectral overlap with the PAN and may differ significantly from it.

Keeping this in mind, we comment on the numerical results with some caution
and await confirmation from the visual analysis to draw firm conclusions.
In particular, the proposed method performs quite well in terms of spatial quality indices, especially when considering $\DR$.
Together with the excellent results of $\DL$, this speaks of a reliable and well-balanced method.
This does not apply to TV, however, whose performance drops dramatically, suggesting questionable spatial quality.
On the contrary, MF is characterized by excellent spatial quality indices but a very bad $\DL$.
Both these methods, and many others, seem unable to achieve a good balance between spectral and spatial fidelity.
Overall, the proposed method compares quite favorably with all DL-based references.
Only R-PNN provides a better $\DR$ on average, but this is also due to the perfect alignment between the metric and the spatial loss of R-PNN,
a property that does not hold with the new spatial loss used in the proposal.

\subsection{Visual results}
The numerical results of Tab.~\ref{tab:RR} provide indications on RR performances that are clear and reliable, although only proxies of the actual performances in the FR space.
These latter are quantified directly in Tab.~\ref{tab:FR}.
However, while $\DL$ captures spectral quality unambiguously,
$D_S$ and $D_{\rho}$ provide more controversial and often contrasting indications on spatial quality.
Therefore, we rely on the visual inspection of some sample results to gain a more solid insight on this latter point.

In Fig.~\ref{fig:FR1} and Fig.~\ref{fig:FR2} we show results for two closeups from the FR Cagliari and the FR Tabasco datasets.
Following \cite{Ciotola2024}, we show two different spectral triplets of bands for false-RGB composition, from the visible and NIR-SWIR spectral ranges.
Given the large number of reference methods,
we limit attention to the top-5 $\DS$ and top-5 $\DR$ methods (based on averages from Tab.~\ref{tab:FR}) plus TV, top-1 $\DL$,
for a total of 10 methods including the proposed one.\footnote{The full set of results is available on the web repository at \\ \git.}
Together with the images, we also report the spectral ($\DL$) and spatial ($\DS$ and $\DR$) quality scores computed over all bands of the full scene.

\newcommand{\D}{\footnotesize}
\newcommand{\image}[1]{\includegraphics[width=0.162\linewidth]{./prisma_img/#1.jpeg}}
\newcommand{\imageS}[1]{\includegraphics[width=0.139\linewidth]{./prisma_img/#1.jpeg}}

\newcommand{\scores}[3]{{\D ($#1$), ($#2$, $#3$)}}
\newcommand{\scoresR}[3]{{\D ($#1$, $#2$, $#3$)}}

\begin{figure*}
    \centering
    \setlength\tabcolsep{1pt}
    {\D
    \begin{tabular}{cccccc}
    PAN                            & HS & BT-H & BDSD-PC & HySURE & DIP-HyperKite \\
                                   &                                 & \scores{\worst{0.063}}{\top{0.000}}{\best{0.058}} & \scores{\worst{0.044}}{\best{0.002}}{0.128} & \scores{\worst{0.113}}{\best{0.002}}{0.141} & \scores{0.021}{0.004}{\worst{0.295}}       \\
    \image{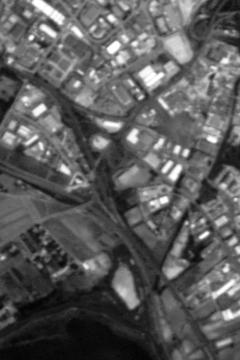} & \image{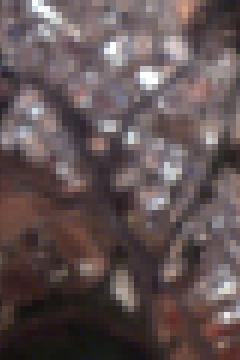} & \image{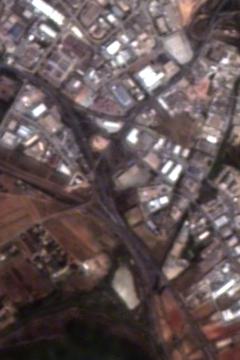} & \image{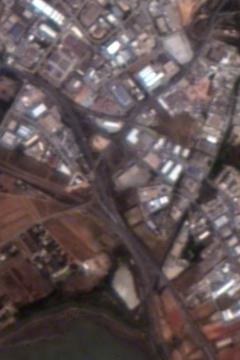} & \image{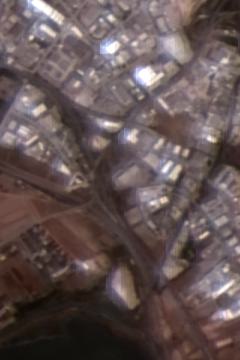} & \image{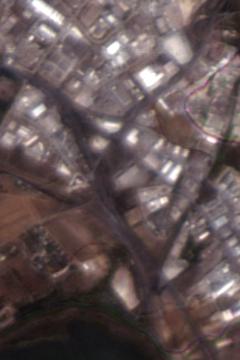} \\%[1mm]
                                   & \image{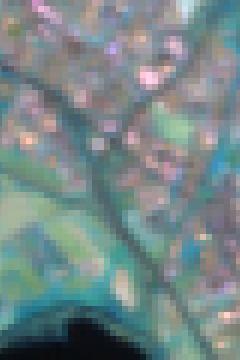}  & \image{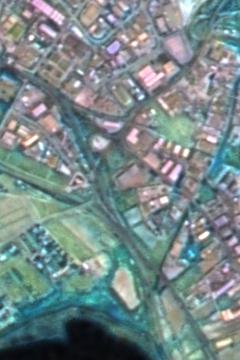}  & \image{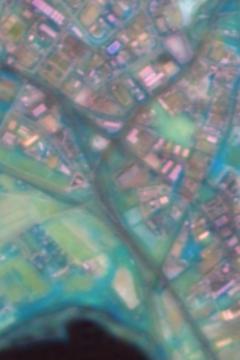}  & \image{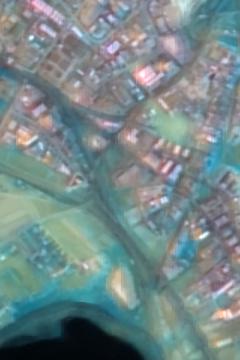}  & \image{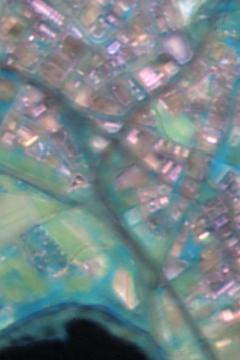}  \\%[1mm]
    \end{tabular}

    \vspace{3mm}

    \begin{tabular}{cccccc}
         HyperPNN & TV & MF &  MTF-GLP-HPM & R-PNN & $\rho$-PNN \\
         \scores{0.041}{\best{0.004}}{0.131}    & \scores{\top{0.004}}{\worst{0.059}}{\worst{0.259}} & \scores{\worst{0.065}}{\worst{0.039}}{\top{0.048}}
            & \scores{0.015}{0.021}{\best{0.048}}& \scores{0.013}{0.017}{\best{0.051}} & \scores{\best{0.004}}{0.022}{0.079}   \\
        \image{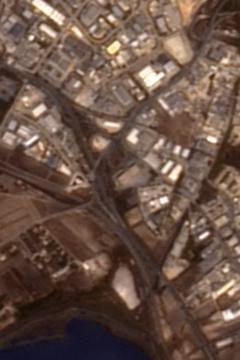} & \image{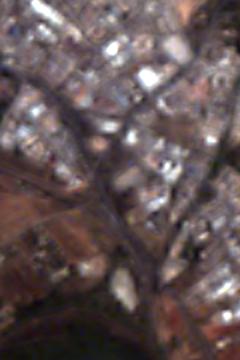} & \image{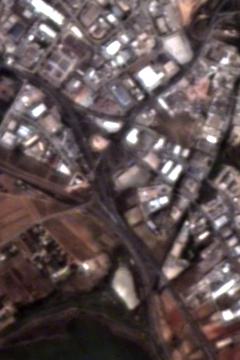}
        & \image{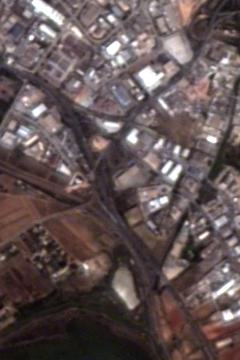} & \image{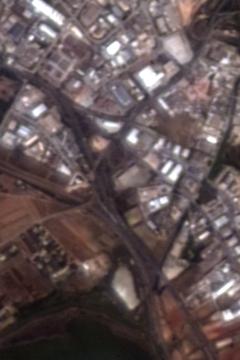} & \image{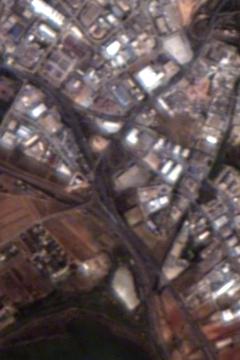} \\%[1mm]
        \image{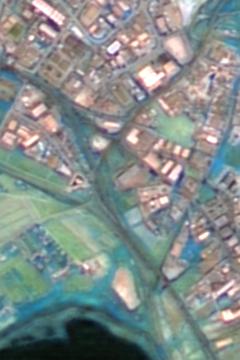} & \image{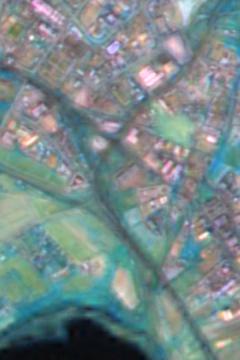} & \image{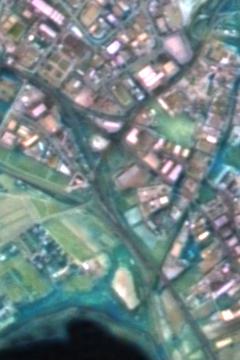}
        & \image{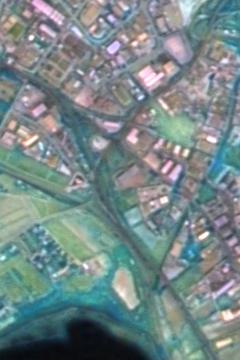} & \image{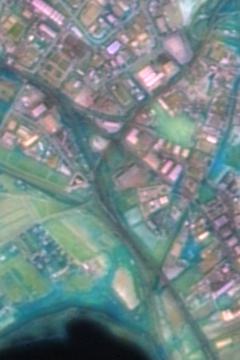} & \image{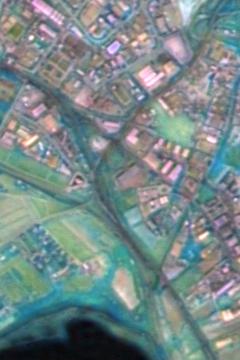}    \\[1mm]
    \end{tabular}
    }
    \caption{Pansharpening results on the {\bf FR Cagliari dataset} (240$\times$360 close-up):
	PAN image (top left) with paired lower resolution HS input
    (bands: 663, 560 and 466 nm)
    followed by a selection of pansharpening results.
    The input HS and all results are also displayed, on even rows,
    using a selection of bands from the NIR-SWIR range (1943, 1261 and 832 nm).
    For each result the corresponding scores ``\scores{\DL}{\DS}{\DR}'' are reported (from Tab.~\ref{tab:FR}).
    }
    \label{fig:FR1}
\end{figure*}

\begin{figure*}
    \centering
    \setlength\tabcolsep{1pt}
    {\D
    \begin{tabular}{cccccc}
        PAN & HS & BT-H & BDSD-PC & HySURE & DIP-HyperKite
       \\
        	    &       & \scores{\worst{0.029}}{\top{0.000}}{\best{0.068}} & \scores{\worst{0.049}}{\best{0.000}}{0.119}
        	    	& \scores{\worst{0.116}}{\best{0.002}}{\worst{0.269}} & \scores{0.015}{\best{0.006}}{\worst{0.276}}
       	\\
        \image{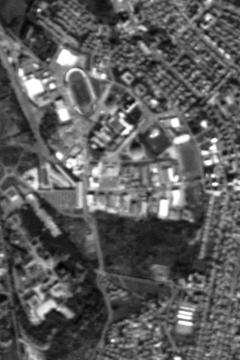} & \image{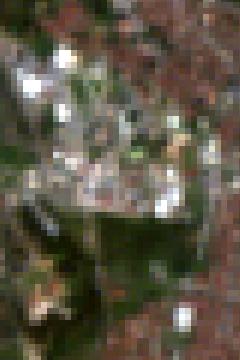} & \image{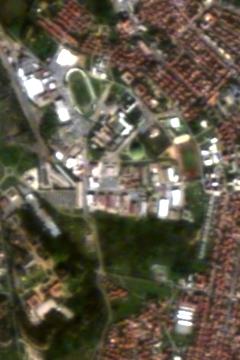}
        & \image{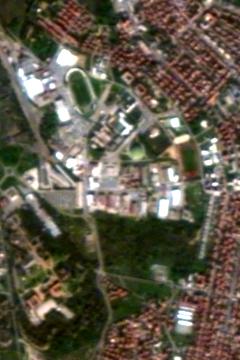} & \image{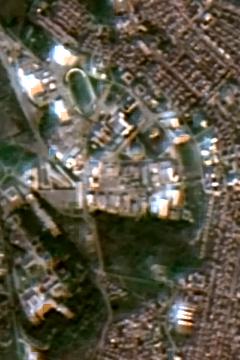} & \image{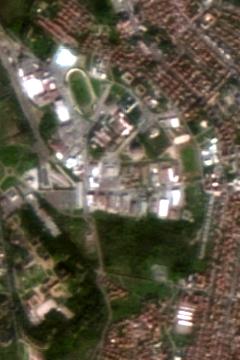}
        \\%[1mm]
       																	& \image{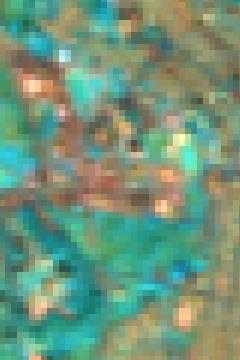} & \image{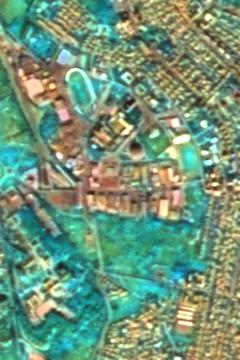}
       	& \image{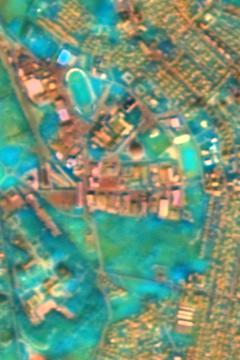} & \image{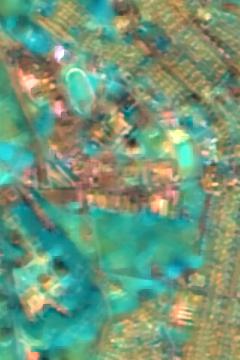} & \image{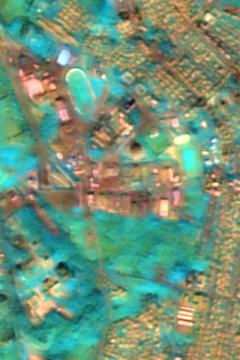}
       	\\%[1mm]
    \end{tabular}

    \vspace{3mm}

    \begin{tabular}{cccccc}
         HyperPNN & TV & MF &  MTF-GLP-HPM & R-PNN & $\rho$-PNN
         \\
         \scores{0.025}{\best{0.007}}{\red{0.262}} & \scores{\top{0.004}}{\worst{0.081}}{0.200} & \scores{\worst{0.033}}{\worst{0.066}}{\best{0.058}}
         	& \scores{0.009}{\worst{0.039}}{\best{0.054}} & \scores{\best{0.007}}{0.034}{\top{0.050}} & \scores{\best{0.004}}{0.038}{0.053}
        \\
        \image{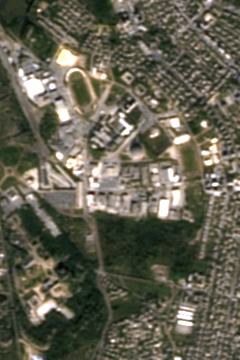} & \image{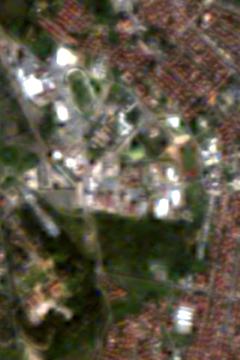} & \image{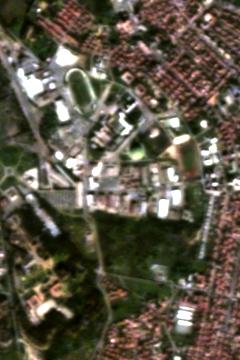}
        & \image{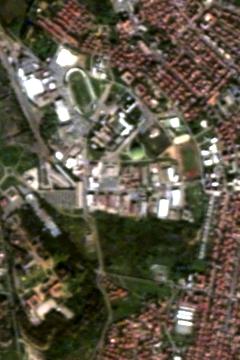} & \image{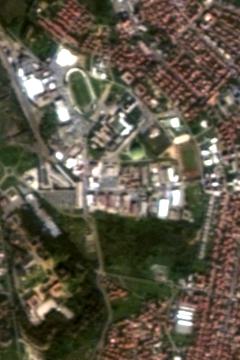} & \image{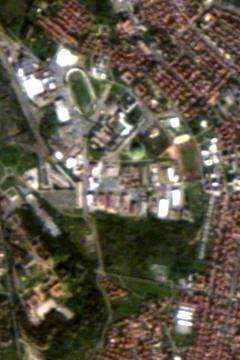}
        \\%[1mm]
        \image{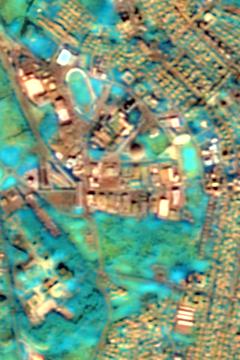} & \image{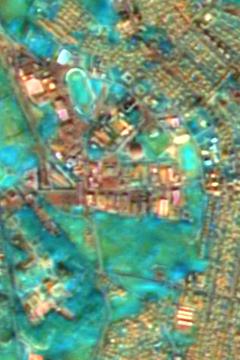} & \image{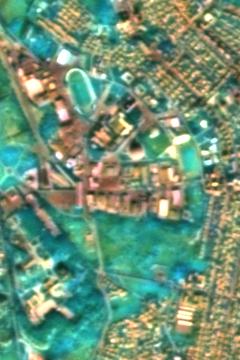}
        & \image{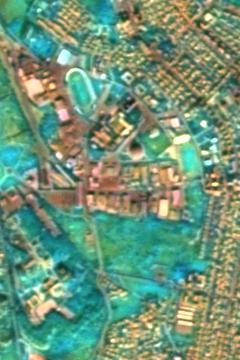} & \image{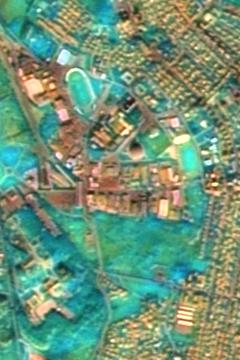} & \image{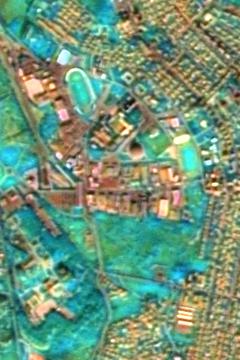}
        \\[1mm]
    \end{tabular}
    }
    \caption{Pansharpening results on the {\bf FR Tabasco dataset} (240$\times$360 close-up):
	PAN image (top left) with paired lower resolution HS input
    (bands: 663, 560 and 466 nm)
    followed by a selection of pansharpening results.
    The input HS and all results are also displayed, on even rows,
    using a selection of bands from the NIR-SWIR range (1943, 1261 and 832 nm).
    For each result the corresponding scores ``\scores{\DL}{\DS}{\DR}'' are reported (from Tab.~\ref{tab:FR}).
    }
    \label{fig:FR2}
\end{figure*}

\begin{figure*}
   \centering
    \setlength\tabcolsep{1pt}
    {\D
    \begin{tabular}{cccccc}
        GT & HyperPNN & HSpeNet & DIP-HyperKite & Hyper-DSNet & $\rho$-PNN \\
         & \scoresR{\best{2.224}}{4.306}{\best{0.854}} & \scoresR{\best{2.053}}{\best{3.216}}{\best{0.859}} & \scoresR{\best{2.021}}{\worst{5.827}}{\best{0.850}} & \scoresR{\top{1.820}}{\top{2.964}}{\top{0.877}} & \scoresR{2.327}{\best{3.561}}{\best{0.828}} \\
        \image{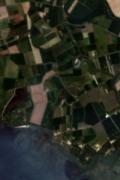} & \image{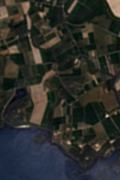} & \image{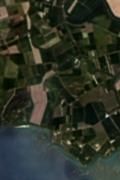} & \image{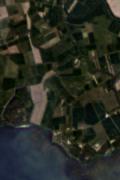} & \image{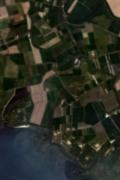} & \image{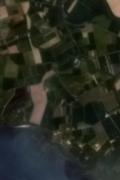} \\%[1mm]
        \image{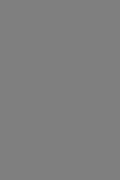} & \image{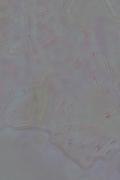} & \image{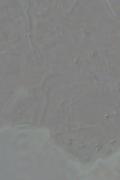} & \image{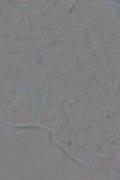} & \image{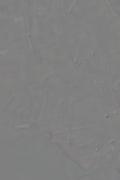} & \image{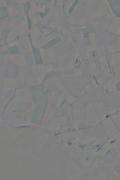}\\ [1mm]
        \multicolumn{6}{c}{(a)} \\ [2mm]
        \image{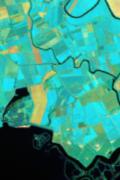} & \image{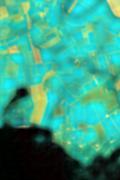} & \image{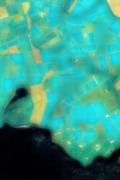} & \image{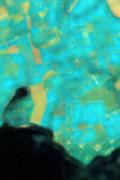} & \image{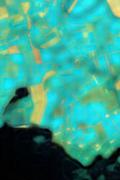} & \image{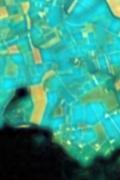}
        \\%[1mm]
       \image{20230824100356RR_ERR_GT} & \image{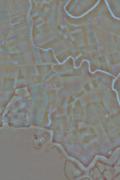} & \image{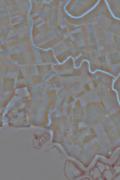} & \image{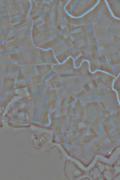} & \image{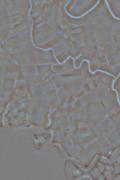} & \image{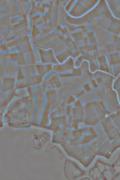}
        \\[1mm]
       \multicolumn{6}{c}{(b)} \\[2mm]
    \end{tabular}
    }
    \caption{Pansharpening results on the {\bf RR Udine dataset} (120$\times$180 close-up) for visible bands (a) and NIR-SWIR bands (b). 
    In both cases, ground truth and pansharpening results (top row) with error maps (bottom row). 
    For each result the corresponding scores ``(ERGAS, SAM, $Q2^n$)'' are reported.
    }
   \label{fig:RR1}
\end{figure*}

From the visual inspection of images we can make the following observations:

\begin{itemize}[noitemsep,leftmargin=*]
\item   Results with low $\DR$ are generally sharper, resembling closely the spatial structures of the PAN.
		This is true for scenes and in both the visible and NIR-SWIR spectral ranges.
		In particular, this is the case of BT-H, MF, MTF-GLP-HPM, R-PNN and proposed.
\item   Results with low $\DS$ reproduce the PAN spatial structures faithfully in the visible spectrum, but seem to fail outside that range.
        This is the case of BDSD-PC, HySURE and DIP-HyperKite.
		BT-H seems an exception, characterized also by a low $\DR$, however it presents a large spectral distortion.
\item   The top-$\DL$ method, TV, whose spectral quality level is nearly the same as the proposed one,
		does not provide effective sharpening, providing blurred images.
\end{itemize}

Based on the above observations, we single out three methods, MTF-GLP-HPM, R-PNN and our proposal, that appear to be both robust and effective,
performing uniformly well on both scenes and all displayed spectral bands in terms of both spectral and spatial quality.
Samples of these methods are shown close to each other
but the differences are so small that further examination seems like a futile exercise that could lead to claims related only to the choice of scene/bands/clips.
On the other hand, all three methods have very similar spatial quality indicators.
However, we point out that the proposed method has a much smaller average spectral distortion (0.0033, see Tab.\ref{tab:FR}) than these competitors (0.0098 and 0.0090, respectively).
Furthermore, only the proposed method ensures uniform quality across all bands, a feature not captured by the global indicators.

We complete this analysis by showing, in Fig.~\ref{fig:RR1},
some error maps obtained by subtracting the ground truth (GT) from the pansharpened image.
In this case, we use the Udine dataset, where some interesting phenomena are visible,
and work at reduced resolution, where the original HS component can be used as GT.
To save space only 5 methods are considered, the proposed one and the best four competitors according to the Q2$^{n}$ metric.
For bands in the visible range (top) only small errors occur, and the error maps appear almost uniformly gray,
with the exception of HyperPNN where some color aberrations are visible.
As expected, larger errors appear for the NIR–SWIR bands (bottom) that do not overlap the PAN.
Interestingly, some sharp structures visible in the GT (probably, waterways) are almost completely lost in the fused images, uniformly for all methods.
On the other hand, such structures are hardly distinguishable in the visible bands and therefore in the PAN,
suggesting that the relevant information may be truly missing in this case.

\subsection{Generalization Ability}

As already said, $\rho$-PNN does not require any preliminary training phase but adapts on the fly to the target data.
Therefore, we expect it to generalize smoothly to new sources, a key requirement of data-driven methods.
To investigate this issue experimentally we would need an alternative dataset, but we could find none available online,
so we built a new one {\it ad hoc}, taking advantage of the good coverage and wide availability of Copernicus Sentinel-2 imagery.
We found a Sentinel-2 image covering the same area as our PRISMA Kansas dataset with a temporal shift of one day.
We used the RGB (665, 560 and 490~nm) high-resolution bands of the Sentinel-2 image to synthesize a 10-meter resolution PAN,
and co-registered it with the PRISMA HS dataset.
Finally, to preserve the resolution ratio of 6 used in the toolbox [42] for all methods,
the hyperspectral data were downscaled by a factor of 2, resulting in a spatial resolution of 60 meters.

On this heterogeneous dataset we repeated the same experiments carried out on the homogeneous PRISMA datasets
working, however, only at full resolution, for lack of sufficient data.
The Results are reported in Table~\ref{tab:FRS} and Figure~\ref{fig:FR3} with the by now usual format,
where comparison methods are selected based on the results of Table~\ref{tab:FRS}.
Both numerical and visual results confirm our expectations:
$\rho$-PNN keeps working very well and is the only method that achieves top-5 ranking for both spectral and spatial quality.
On the contrary, several deep learning-based methods appear to struggle with these new data, especially in terms of spectral quality.
This is especially obvious for HyperPNN, with pansharpened images that are severely corrupted, see Figure~\ref{fig:FR3}.
Model-based methods, on the other hand, are not significantly affected by the data mismatch.
In extreme synthesis, these results show that the proposed method keeps ensuring a very good and stable performance
also on this new dataset, confirming the expected good generalization ability.

\begin{table}
\caption{Results at Full Resolution on \textbf{Sentinel-2 / PRISMA dataset}.  \top{Best score} is in \top{bold green}, \best{Top-5} in \best{green},  and \worst{Worst-5} in \worst{red}}.
\footnotesize
\centering
\setlength{\tabcolsep}{2pt}
    %\begin{tabular}{lcccc}
    \begin{tabular}{lc@{\rule{2mm}{0mm}}ccc}
        \hline
        \ru Method	&	&		$D_{\lambda}$		&		$D_{S}$		&		$D_{\rho}$		 \\
        \cline{1-1} \cline{3-5}
        \ru (Ideal) & & 0 & 0 & 0 \\
        \cline{1-1} \cline{3-5}
       \ru  EXP	&	&	0.0039 &	0.0985	&	0.7919 \\
        \cline{1-1} \cline{3-5}
        \ru GSA	&	&	0.0068 & \best{0.0002} & \worst{0.2833}  \\
        \ru BT-H	&	&	\worst{0.0338} & \top{0.0000} &	\top{0.0298} \\
        \ru BDSD-PC	&	&	0.0095 & \best{0.0011} & 0.1236 \\
        \ru PRACS	&	&	0.0047	&	\best{	0.0023}	&	0.2629 \\
        \cline{1-1} \cline{3-5}
        \ru MTF-GLP-FS	&	&	\best{0.0019} &	0.0134	&	0.2820	\\
        \ru MTF-GLP-HPM	&	&	0.0039	&	\worst{0.0140} & \best{0.0384}  \\
        \ru MTF-GLP-HPM-R	&	&	\best{0.0019}	&	0.0139	&	\worst{0.2861} \\
        \ru AWLP	&	&	0.0034	&	\worst{0.0154} & 0.0626	 \\
        \ru MF	&	&	0.0278	&	\worst{0.0340} &	\best{0.0439} \\
        \cline{1-1} \cline{3-5}
        \ru HySURE	&	&	\worst{0.0553}	&	0.0068	&	0.1323 \\
        \ru SR-D	&	&	0.0043	&	\worst{0.1015}	&	\worst{0.8624} \\
        \ru TV	&	&	\best{0.0018}	&	\worst{0.0536} &	0.2117 \\
        \cline{1-1} \cline{3-5}
        \ru HyperPNN &	&	\worst{0.0824}	&	\best{0.0027}	&	0.1967 \\
        \ru HSpeNet	&	&	\worst{0.0920} & 0.0033 & 0.0799	 \\
        \ru DHP-DARN	&	&	\worst{0.1136}	&	0.0035 & \worst{0.3505}  \\
        \ru DIP-HyperKite	&	&	0.0044	&	0.0032 &	0.1657  \\
        \ru Hyper-DSNet	&	&	0.0284	&	0.0121 & 0.1894	 \\
        \ru R-PNN	&	&	0.0032 &	0.0079	&	\best{0.0516}	 \\
        \ru PCA-Z-PNN	&	&	\best{0.0013} & 0.0067 & \worst{0.8200} \\
        \cline{1-1} \cline{3-5}
        \ru $\rho$-PNN	&	&	\top{0.0013} & 0.0071	&	\best{0.0521} \\
        \hline
        \end{tabular}
\label{tab:FRS}
\end{table}

\begin{figure*}
    \centering
    \setlength\tabcolsep{1pt}
    {\D
    \begin{tabular}{ccccccc}
        PAN & HS & GSA & BT-H & BDSD-PC & PRACS \\
         & & \scores{0.007}{\best{0.000}}{\worst{0.283}} & \scores{\worst{0.034}}{\top{0.000}}{\top{0.030}} & \scores{0.010}{\best{0.001}}{0.124}   & \scores{0.005}{\best{0.02}}{0.263} \\
        \image{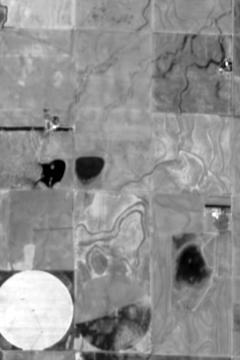} & \image{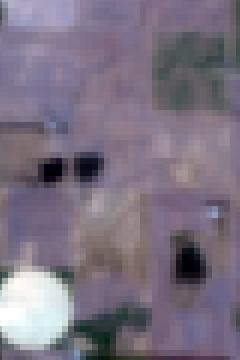} & \image{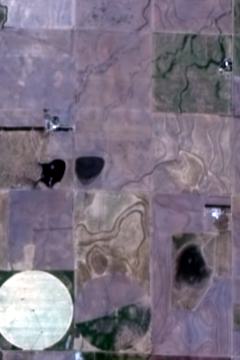} &  \image{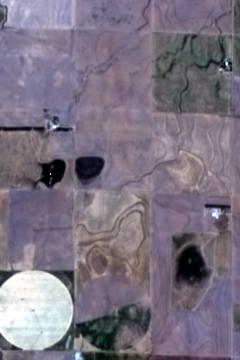} & \image{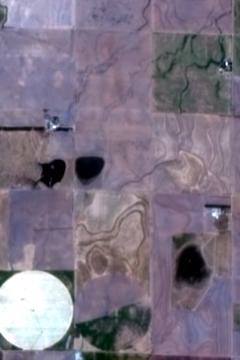}  & \image{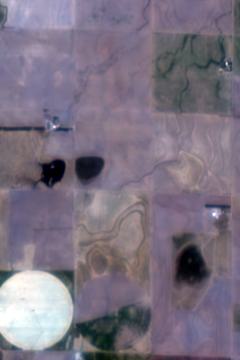} \\%[1mm]
         & \image{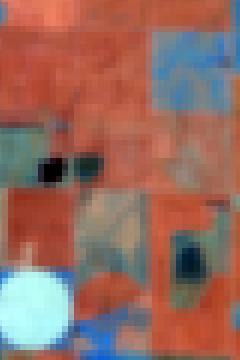} & \image{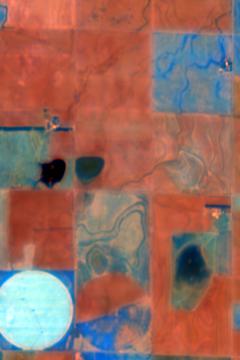} &  \image{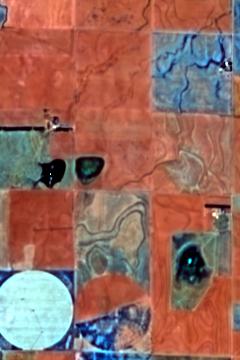} & \image{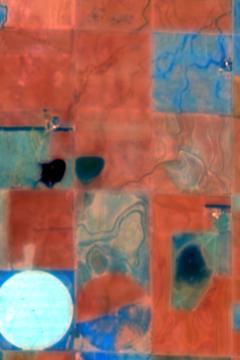}  & \image{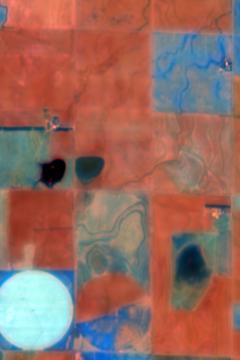} \\%[1mm]

       HyperPNN & MTF-GLP-HPM & MF & R-PNN & PCA-Z-PNN& $\rho$-PNN \\
       \scores{\worst{0.082}}{\best{0.003}}{0.197} & \scores{0.004}{\worst{0.014}}{\best{0.038}} & \scores{0.028}{\worst{0.034}}{\best{0.044}} &  \scores{0.003}{0.008}{\best{0.052}} & \scores{\best{0.001}}{0.007}{\worst{0.820}} & \scores{\top{0.001}}{0.007}{\best{0.052}} \\
       \image{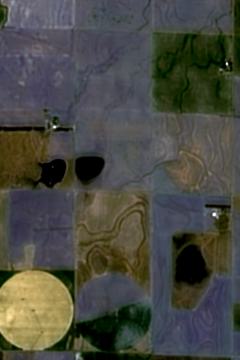} & \image{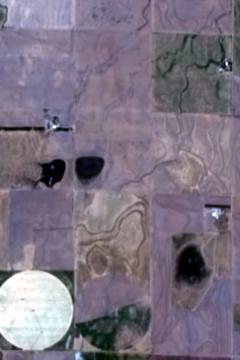} & \image{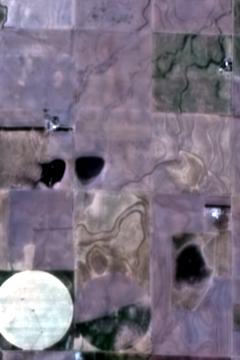} & \image{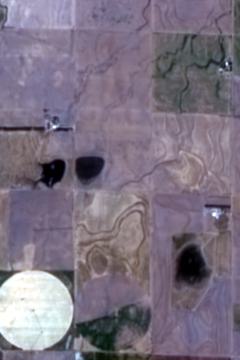} & \image{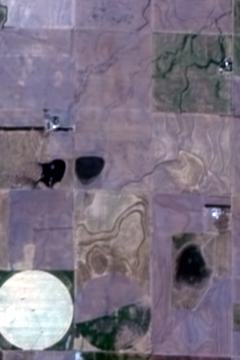} & \image{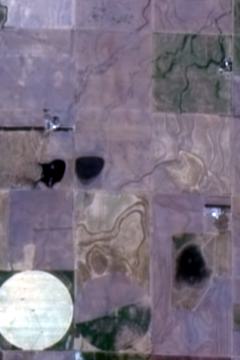} \\
       \image{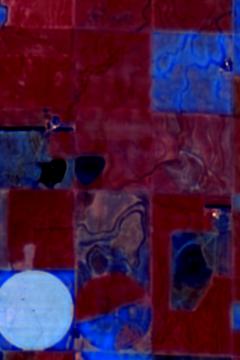} & \image{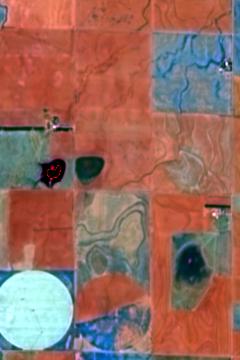} & \image{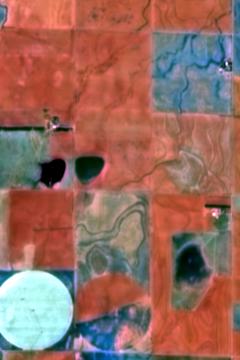} & \image{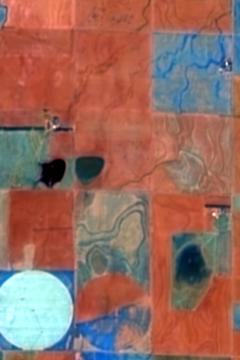} & \image{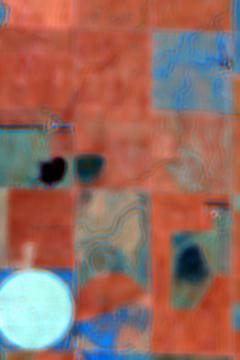} & \image{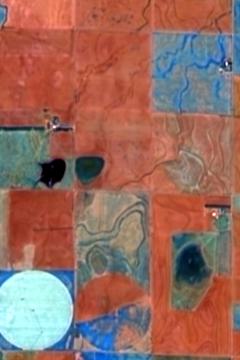} \\
    \end{tabular}
    }
    \caption{Pansharpening results on the synthetic {\bf Sentinel-2/PRISMA dataset} (240$\times$360 close-up):
	Simulated PAN image obtained from Sentinel-2 RGB bands (top left) with paired reduced resolution PRISMA HS input
    (bands: 663, 560 and 466 nm)
    followed by a selection of pansharpening results.
    The input HS and all results are also displayed, on even rows,
    using a selection of bands from the NIR-SWIR range (1943, 1261 and 832 nm).
    For each result the corresponding scores ``\scores{\DL}{\DS}{\DR}'' are reported (from Tab.~\ref{tab:FRS}).
    }
    \label{fig:FR3}
\end{figure*}

\subsection{Discussion}

This large body of experimental results fully confirms our expectations about the proposed method.
For the sake of clarity, we summarize here its pros and cons, also with reference to competing methods.

From a structural point of view, $\rho$-PNN presents several qualifying points, some of which are shared only with its ancestor R-PNN, and with PCA-Z-PNN:
\begin{itemize}
\item it runs from scratch, without the need for prior training: this is obtained by means of a longer tuning run on the first band that affects only marginally the overall computational cost due to the very large number of bands;
\item it works with an arbitrary number of bands: as the algorithm works band-wise with no joint processing of the HS stack, the number of spectral bands does not need to be fixed to build the model;
\item it operates at full resolution with no resolution downgrading or simulation operations: as the proposed loss is fully unsupervised and based of input-output self-consistency, traditional resolution downgrade protocols ({\em e.g.}, \cite{Wald1997}) to give rise to labeled, but synthetic, dataset are avoided.
\end{itemize}
Furthermore, in terms of performance
\begin{itemize}
\item it is on par with the best SotA methods, both model-based and data-driven;
\item it guarantees a uniformly high spectral quality at all wavelengths, as only the TV method does;
\item it shows excellent generalization ability.
\end{itemize}
On the down side,
$\rho$-PNN has a relatively high computational cost compared to model-based techniques and even several data-driven techniques.
This cost can be significantly reduced, however, by slightly relaxing the spectral quality constraints.

\section{Conclusions}
\label{sec:conclusions}

In this work
we have proposed a new pansharpening method aimed at ensuring a uniformly high fidelity over the whole spectral range.
This is an important feature for hyperspectral images, where most bands are acquired at spectral frequencies not covered by the PAN, and may differ significantly from it.
To this end, we leverage the rolling PNN architecture where pansharpening is carried out band-wise using cross-band transfer learning and a suitable zero-shot unsupervised tuning.
From this base, we improve with two key innovations:
a new loss, better adapted to the characteristics of HS images, that takes into account the correlation inversion phenomena that often occur in bands outside the visible range;
and a hysteresis-based tuning scheme where the balance of spatial and spectral loss components changes dynamically to overcome unwanted low-quality solutions.

\begin{comment}
The proposed method inherits the precious features of R-PNN, no need for pre-training, very good generalization ability, flexibility with respect to number of bands.
To these, it adds the performance improvements granted by the HS-tailored spatial loss, the improved optimization scheme, and the uniformly high spectral quality.
\end{comment}

Extensive experimental assessment over a SotA benchmarking toolbox proves the proposed method
to ensure excellent results both in terms of visual quality and according to the most credited quality indicators.

However, this latter point deserves certainly further research.
To date, no spatial quality metric can be considered a fully reliable indicator of image quality at full-resolution.
Lacking a solid unambiguous reference, numerical assessment of performance remains inconclusive.
This is the most urgent research topic in the pansharpening context, in our opinion,
as it is intimately linked to the definition of an effective loss for unsupervised learning.
%We conclude, recalling that all the material related to this work is shared on GitHub to ensure full reproducibility of our results further encouraging the future research on this topic.

\bibliographystyle{IEEEtran}
\bibliography{refs}

\begin{thebibliography}{10}
\providecommand{\url}[1]{#1}
\csname url@samestyle\endcsname
\providecommand{\newblock}{\relax}
\providecommand{\bibinfo}[2]{#2}
\providecommand{\BIBentrySTDinterwordspacing}{\spaceskip=0pt\relax}
\providecommand{\BIBentryALTinterwordstretchfactor}{4}
\providecommand{\BIBentryALTinterwordspacing}{\spaceskip=\fontdimen2\font plus
\BIBentryALTinterwordstretchfactor\fontdimen3\font minus
  \fontdimen4\font\relax}
\providecommand{\BIBforeignlanguage}[2]{{%
\expandafter\ifx\csname l@#1\endcsname\relax
\typeout{** WARNING: IEEEtran.bst: No hyphenation pattern has been}%
\typeout{** loaded for the language `#1'. Using the pattern for}%
\typeout{** the default language instead.}%
\else
\language=\csname l@#1\endcsname
\fi
#2}}
\providecommand{\BIBdecl}{\relax}
\BIBdecl

\bibitem{Goodenough2003}
D.~Goodenough, A.~Dyk, K.~Niemann, J.~Pearlman, H.~Chen, T.~Han, M.~Murdoch,
  and C.~West, ``Processing {Hyperion} and {ALI} for forest classification,''
  \emph{IEEE Transactions on Geoscience and Remote Sensing}, vol.~41, no.~6,
  pp. 1321--1331, 2003.

\bibitem{Dalponte2013}
M.~Dalponte, H.~O. Ørka, T.~Gobakken, D.~Gianelle, and E.~Næsset, ``Tree
  species classification in boreal forests with hyperspectral data,''
  \emph{IEEE Transactions on Geoscience and Remote Sensing}, vol.~51, no.~5,
  pp. 2632--2645, 2013.

\bibitem{Jia2022}
J.~Jia, J.~Chen, X.~Zheng, Y.~Wang, S.~Guo, H.~Sun, C.~Jiang, M.~Karjalainen,
  K.~Karila, Z.~Duan, T.~Wang, C.~Xu, J.~Hyyppä, and Y.~Chen, ``Tradeoffs in
  the spatial and spectral resolution of airborne hyperspectral imaging
  systems: A crop identification case study,'' \emph{IEEE Transactions on
  Geoscience and Remote Sensing}, vol.~60, pp. 1--18, 2022.

\bibitem{Reddy2024}
C.~Kishor Kumar~Reddy, A.~Daduvy, R.~Madana~Mohana, B.~Assiri, M.~Shuaib,
  S.~Alam, and M.~Abdullah~Sheneamer, ``Enhancing precision agriculture and
  land cover classification: A self-attention {3D} convolutional neural network
  approach for hyperspectral image analysis,'' \emph{IEEE Access}, vol.~12, pp.
  125\,592--125\,608, 2024.

\bibitem{Dumke2019}
I.~Dumke, M.~Ludvigsen, S.~L. Ellefmo, F.~Søreide, G.~Johnsen, and B.~J.
  Murton, ``Underwater hyperspectral imaging using a stationary platform in the
  trans-atlantic geotraverse hydrothermal field,'' \emph{IEEE Transactions on
  Geoscience and Remote Sensing}, vol.~57, no.~5, pp. 2947--2962, 2019.

\bibitem{Kang2023}
X.~Kang, B.~Deng, P.~Duan, X.~Wei, and S.~Li, ``Self-supervised
  spectral–spatial transformer network for hyperspectral oil spill mapping,''
  \emph{IEEE Transactions on Geoscience and Remote Sensing}, vol.~61, pp.
  1--10, 2023.

\bibitem{Brando2003}
V.~Brando and A.~Dekker, ``Satellite hyperspectral remote sensing for
  estimating estuarine and coastal water quality,'' \emph{IEEE Transactions on
  Geoscience and Remote Sensing}, vol.~41, no.~6, pp. 1378--1387, 2003.

\bibitem{Gao2022}
Y.~Gao, W.~Li, M.~Zhang, J.~Wang, W.~Sun, R.~Tao, and Q.~Du, ``Hyperspectral
  and multispectral classification for coastal wetland using depthwise feature
  interaction network,'' \emph{IEEE Transactions on Geoscience and Remote
  Sensing}, vol.~60, pp. 1--15, 2022.

\bibitem{Hestir2015}
E.~L. Hestir, V.~E. Brando, M.~Bresciani, C.~Giardino, E.~Matta, P.~Villa, and
  A.~G. Dekker, ``Measuring freshwater aquatic ecosystems: The need for a
  hyperspectral global mapping satellite mission,'' \emph{Remote Sensing of
  Environment}, vol. 167, pp. 181--195, 2015, special Issue on the
  Hyperspectral Infrared Imager (HyspIRI).

\bibitem{Su2023}
H.~Su, F.~Shao, Y.~Gao, H.~Zhang, W.~Sun, and Q.~Du, ``Probabilistic
  collaborative representation based ensemble learning for classification of
  wetland hyperspectral imagery,'' \emph{IEEE Transactions on Geoscience and
  Remote Sensing}, vol.~61, pp. 1--17, 2023.

\bibitem{Nolin2000}
A.~W. Nolin and J.~Dozier, ``A hyperspectral method for remotely sensing the
  grain size of snow,'' \emph{Remote Sensing of Environment}, vol.~74, no.~2,
  pp. 207--216, 2000.

\bibitem{Han2021}
\BIBentryALTinterwordspacing
Y.~Han, X.~Shi, S.~Yang, Y.~Zhang, Z.~Hong, and R.~Zhou, ``Hyperspectral sea
  ice image classification based on the spectral-spatial-joint feature with the
  {PCA} network,'' \emph{Remote Sensing}, vol.~13, no.~12, 2021. [Online].
  Available: \url{https://www.mdpi.com/2072-4292/13/12/2253}
\BIBentrySTDinterwordspacing

\bibitem{Di2022}
D.~Di, J.~Li, W.~Han, and R.~Yin, ``Geostationary hyperspectral infrared
  sounder channel selection for capturing fast-changing atmospheric
  information,'' \emph{IEEE Transactions on Geoscience and Remote Sensing},
  vol.~60, pp. 1--10, 2022.

\bibitem{Wang2024}
W.~Wang, C.~Shi, H.~Shang, S.~Yin, J.~Xu, N.~Xu, L.~Chen, and H.~Letu,
  ``Development of an algorithm for the simultaneous retrieval of cloud-top
  height and cloud optical thickness combining radiative transfer and
  multisource satellite information from {O}$_4$ hyperspectral measurements,''
  \emph{IEEE Transactions on Geoscience and Remote Sensing}, vol.~62, pp.
  1--11, 2024.

\bibitem{Siebels2020}
K.~Siebels, K.~Goïta, and M.~Germain, ``Estimation of mineral abundance from
  hyperspectral data using a new supervised neighbor-band ratio unmixing
  approach,'' \emph{IEEE Transactions on Geoscience and Remote Sensing},
  vol.~58, no.~10, pp. 6754--6766, 2020.

\bibitem{Lorenz2021}
S.~Lorenz, P.~Ghamisi, M.~Kirsch, R.~Jackisch, B.~Rasti, and R.~Gloaguen,
  ``Feature extraction for hyperspectral mineral domain mapping: A test of
  conventional and innovative methods,'' \emph{Remote Sensing of Environment},
  vol. 252, p. 112129, 2021.

\bibitem{Simoes2014}
M.~Simoes, J.~Bioucas-Dias, L.~B. Almeida, and J.~Chanussot, ``A convex
  formulation for hyperspectral image superresolution via subspace-based
  regularization,'' \emph{IEEE Transactions on Geoscience and Remote Sensing},
  vol.~53, no.~6, pp. 3373--3388, 2014.

\bibitem{Roessner2001}
S.~Roessner, K.~Segl, U.~Heiden, and H.~Kaufmann, ``Automated differentiation
  of urban surfaces based on airborne hyperspectral imagery,'' \emph{IEEE
  Transactions on Geoscience and Remote Sensing}, vol.~39, no.~7, pp.
  1525--1532, 2001.

\bibitem{Huang2008}
X.~Huang and L.~Zhang, ``An adaptive mean-shift analysis approach for object
  extraction and classification from urban hyperspectral imagery,'' \emph{IEEE
  Transactions on Geoscience and Remote Sensing}, vol.~46, no.~12, pp.
  4173--4185, 2008.

\bibitem{Wu2024}
X.~Wu, J.~Feng, R.~Shang, J.~Wu, X.~Zhang, L.~Jiao, and P.~Gamba, ``Multi-task
  multi-objective evolutionary network for hyperspectral image classification
  and pansharpening,'' \emph{Information Fusion}, vol. 108, p. 102383, 2024.

\bibitem{Lin2015}
C.~Lin, C.-C. Wu, K.~Tsogt, Y.-C. Ouyang, and C.-I. Chang, ``Effects of
  atmospheric correction and pansharpening on lulc classification accuracy
  using worldview-2 imagery,'' \emph{Information Processing in Agriculture},
  vol.~2, no.~1, pp. 25--36, 2015.

\bibitem{Vivone2020}
G.~Vivone, M.~Dalla~Mura, A.~Garzelli, R.~Restaino, G.~Scarpa, M.~O. Ulfarsson,
  L.~Alparone, and J.~Chanussot, ``A new benchmark based on recent advances in
  multispectral pansharpening: Revisiting pansharpening with classical and
  emerging pansharpening methods,'' \emph{IEEE Geoscience and Remote Sensing
  Magazine}, vol.~9, no.~1, pp. 53--81, 2021.

\bibitem{Deng2022}
L.-j. Deng, G.~Vivone, M.~E. Paoletti, G.~Scarpa, J.~He, Y.~Zhang,
  J.~Chanussot, and A.~Plaza, ``Machine learning in pansharpening: A benchmark,
  from shallow to deep networks,'' \emph{IEEE Geoscience and Remote Sensing
  Magazine}, vol.~10, no.~3, pp. 279--315, 2022.

\bibitem{Aiazzi2007}
B.~Aiazzi, S.~Baronti, and M.~Selva, ``Improving component substitution
  pansharpening through multivariate regression of {MS+Pan} data,'' \emph{IEEE
  Transactions on Geoscience and Remote Sensing}, vol.~45, no.~10, pp.
  3230--3239, 2007.

\bibitem{Garzelli2008}
A.~Garzelli, F.~Nencini, and L.~Capobianco, ``{Optimal MMSE pan sharpening of
  very high resolution multispectral images},'' \emph{IEEE Transactions on
  Geoscience and Remote Sensing}, vol.~46, no.~1, pp. 228--236, 2008.

\bibitem{Choi2011}
J.~Choi, K.~Yu, and Y.~Kim, ``A new adaptive component-substitution-based
  satellite image fusion by using partial replacement,'' \emph{IEEE
  Transactions on Geoscience and Remote Sensing}, vol.~49, no.~1, pp. 295--309,
  2011.

\bibitem{Lolli2017}
S.~{Lolli}, L.~{Alparone}, A.~{Garzelli}, and G.~{Vivone}, ``Haze correction
  for contrast-based multispectral pansharpening,'' \emph{IEEE Geoscience and
  Remote Sensing Letters}, vol.~14, no.~12, pp. 2255--2259, 2017.

\bibitem{Otazu2005}
X.~Otazu, M.~Gonz\'alez-Aud\'icana, O.~Fors, and J.~N\'u{\~n}ez, ``Introduction
  of sensor spectral response into image fusion methods. {A}pplication to
  wavelet-based methods,'' \emph{IEEE Transactions on Geoscience and Remote
  Sensing}, vol.~43, no.~10, pp. 2376--2385, 2005.

\bibitem{Aiazzi2006}
B.~Aiazzi, L.~Alparone, S.~Baronti, A.~Garzelli, and M.~Selva, ``{MTF}-tailored
  multiscale fusion of high-resolution {MS} and {P}an imagery,''
  \emph{Photogrammetric Engineering and Remote Sensing}, vol.~72, no.~5, pp.
  591--596, 2006.

\bibitem{Restaino2016}
R.~Restaino, G.~Vivone, M.~{Dalla Mura}, and J.~Chanussot, ``Fusion of
  multispectral and panchromatic images based on morphological operators,''
  \emph{IEEE Transactions on Image Processing}, vol.~25, no.~6, pp. 2882--2895,
  2016.

\bibitem{Palsson2014}
F.~Palsson, J.~R. Sveinsson, and M.~O. Ulfarsson, ``A new pansharpening
  algorithm based on total variation,'' \emph{IEEE Geoscience and Remote
  Sensing Letters}, vol.~11, no.~1, pp. 318--322, 2014.

\bibitem{Vivone2015}
G.~{Vivone}, M.~{Simões}, M.~{Dalla Mura}, R.~{Restaino}, J.~M.
  {Bioucas-Dias}, G.~A. {Licciardi}, and J.~{Chanussot}, ``Pansharpening based
  on semiblind deconvolution,'' \emph{IEEE Transactions on Geoscience and
  Remote Sensing}, vol.~53, no.~4, pp. 1997--2010, 2015.

\bibitem{Palsson2020}
F.~Palsson, M.~O. Ulfarsson, and J.~R. Sveinsson, ``Model-based reduced-rank
  pansharpening,'' \emph{IEEE Geoscience and Remote Sensing Letters}, vol.~17,
  no.~4, pp. 656--660, 2020.

\bibitem{Yu2021}
L.~Yu, D.~Liu, H.~Mansour, and P.~T. Boufounos, ``Fast and high-quality blind
  multi-spectral image pansharpening,'' \emph{IEEE Transactions on Geoscience
  and Remote Sensing}, vol.~60, pp. 1--17, 2021.

\bibitem{Masi2016}
\BIBentryALTinterwordspacing
G.~Masi, D.~Cozzolino, L.~Verdoliva, and G.~Scarpa, ``Pansharpening by
  convolutional neural networks,'' \emph{Remote Sensing}, vol.~8, no.~7, p.
  594, 2016. [Online]. Available: \url{http://www.mdpi.com/2072-4292/8/7/594}
\BIBentrySTDinterwordspacing

\bibitem{Yang2017}
J.~Yang, X.~Fu, Y.~Hu, Y.~Huang, X.~Ding, and J.~Paisley, ``{PanNet}: A deep
  network architecture for pan-sharpening,'' in \emph{ICCV}, 2017.

\bibitem{Scarpa2018}
G.~Scarpa, S.~Vitale, and D.~Cozzolino, ``Target-adaptive {CNN}-based
  pansharpening,'' \emph{IEEE Transactions on Geoscience and Remote Sensing},
  vol.~56, no.~9, pp. 5443--5457, 2018.

\bibitem{Ciotola2023a}
\BIBentryALTinterwordspacing
M.~Ciotola and G.~Scarpa, ``Fast full-resolution target-adaptive {CNN}-based
  pansharpening framework,'' \emph{Remote Sensing}, vol.~15, no.~2, 2023.
  [Online]. Available: \url{https://www.mdpi.com/2072-4292/15/2/319}
\BIBentrySTDinterwordspacing

\bibitem{Ciotola2024a}
\BIBentryALTinterwordspacing
M.~Ciotola, G.~Guarino, and G.~Scarpa, ``An unsupervised {CNN}-based
  pansharpening framework with spectral-spatial fidelity balance,''
  \emph{Remote Sensing}, vol.~16, no.~16, 2024. [Online]. Available:
  \url{https://www.mdpi.com/2072-4292/16/16/3014}
\BIBentrySTDinterwordspacing

\bibitem{Ciotola2024}
M.~Ciotola, G.~Guarino, G.~Vivone, G.~Poggi, J.~Chanussot, A.~Plaza, and
  G.~Scarpa, ``Hyperspectral pansharpening: Critical review, tools, and future
  perspectives,'' \emph{IEEE Geoscience and Remote Sensing Magazine}, vol.~13,
  no.~1, pp. 311--338, 2025.

\bibitem{Vivone2023}
G.~Vivone, A.~Garzelli, Y.~Xu, W.~Liao, and J.~Chanussot, ``Panchromatic and
  hyperspectral image fusion: Outcome of the 2022 {WHISPERS} hyperspectral
  pansharpening challenge,'' \emph{IEEE Journal of Selected Topics in Applied
  Earth Observations and Remote Sensing}, vol.~16, pp. 166--179, 2023.

\bibitem{Capo07}
L.~Capobianco, A.~Garzelli, F.~Nencini, L.~Alparone, and S.~Baronti, ``Spatial
  enhancement of {Hyperion} hyperspectral data through {ALI} panchromatic
  image,'' in \emph{IEEE International Geoscience and Remote Sensing Symposium
  IGARSS}, 2007, pp. 5158--5161.

\bibitem{Zhang2009}
Y.~Zhang, S.~De~Backer, and P.~Scheunders, ``Noise-resistant wavelet-based
  bayesian fusion of multispectral and hyperspectral images,'' \emph{IEEE
  Transactions on Geoscience and Remote Sensing}, vol.~47, no.~11, pp.
  3834--3843, 2009.

\bibitem{Simoes2015}
M.~Simoes, J.~Bioucas-Dias, L.~Almeida, and J.~Chanussot, ``A convex
  formulation for hyperspectral image superresolution via subspace-based
  regularization,'' \emph{IEEE Transactions on Geoscience and Remote Sensing},
  vol.~53, no.~6, pp. 3373--3388, 2015.

\bibitem{Wei2015}
Q.~Wei, N.~Dobigeon, and J.-Y. Tourneret, ``Bayesian fusion of multi-band
  images,'' \emph{IEEE Journal of Selected Topics in Signal Processing},
  vol.~9, no.~6, pp. 1117--1127, 2015.

\bibitem{Wei2015a}
Q.~Wei, J.~Bioucas-Dias, N.~Dobigeon, and J.-Y. Tourneret, ``Hyperspectral and
  multispectral image fusion based on a sparse representation,'' \emph{IEEE
  Transactions on Geoscience and Remote Sensing}, vol.~53, no.~7, pp.
  3658--3668, 2015.

\bibitem{Berne2010}
O.~Bern{\'e}, A.~Helens, P.~Pilleri, and C.~Joblin, ``Non-negative matrix
  factorization pansharpening of hyperspectral data: An application to
  mid-infrared astronomy,'' in \emph{2010 2nd Workshop on Hyperspectral Image
  and Signal Processing: Evolution in Remote Sensing}.\hskip 1em plus 0.5em
  minus 0.4em\relax IEEE, 2010, pp. 1--4.

\bibitem{Moel09}
M.~Moeller, T.~Wittman, and A.~L. Bertozzi, ``A variational approach to
  hyperspectral image fusion,'' in \emph{Algorithms and Technologies for
  Multispectral, Hyperspectral, and Ultraspectral Imagery XV}, vol. 7334.\hskip
  1em plus 0.5em minus 0.4em\relax SPIE, 2009, pp. 502--511.

\bibitem{Huang2013}
B.~Huang, H.~Song, H.~Cui, J.~Peng, and Z.~Xu, ``Spatial and spectral image
  fusion using sparse matrix factorization,'' \emph{IEEE Transactions on
  Geoscience and Remote Sensing}, vol.~52, no.~3, pp. 1693--1704, 2013.

\bibitem{Kawakami2011}
R.~Kawakami, Y.~Matsushita, J.~Wright, M.~Ben-Ezra, Y.-W. Tai, and K.~Ikeuchi,
  ``High-resolution hyperspectral imaging via matrix factorization,'' in
  \emph{The IEEE Conference on Computer Vision and Pattern Recognition
  (CVPR)}.\hskip 1em plus 0.5em minus 0.4em\relax IEEE, 2011, pp. 2329--2336.

\bibitem{Vivone2014b}
G.~Vivone, R.~Restaino, G.~Licciardi, M.~{Dalla Mura}, and J.~Chanussot,
  ``Multiresolution analysis and component substitution techniques for
  hyperspectral pansharpening,'' in \emph{IEEE International Geoscience and
  Remote Sensing Symposium IGARSS}, 2014, pp. 2649--2652.

\bibitem{Loncan2015}
L.~Loncan, L.~De~Almeida, J.~Bioucas-Dias, X.~Briottet, J.~Chanussot,
  N.~Dobigeon, S.~Fabre, W.~Liao, G.~Licciardi, M.~Simoes \emph{et~al.},
  ``Hyperspectral pansharpening: A review,'' \emph{IEEE Geoscience and Remote
  Sensing Magazine}, vol.~3, no.~3, pp. 27--46, 2015.

\bibitem{Qu17}
J.~Qu, Y.~Li, and W.~Dong, ``\BIBforeignlanguage{en}{Hyperspectral
  pansharpening with guided filter},'' \emph{\BIBforeignlanguage{en}{IEEE
  Geoscience and Remote Sensing Letters}}, vol.~14, no.~11, pp. 2152--2156,
  2017.

\bibitem{Adde17}
P.~Addesso, M.~Dalla~Mura, L.~Condat, R.~Restaino, G.~Vivone, D.~Picone, and
  J.~Chanussot, ``Collaborative total variation for hyperspectral
  pansharpening,'' in \emph{IEEE International Geoscience and Remote Sensing
  Symposium IGARSS}, 2017, pp. 2597--2600.

\bibitem{Huan17}
Z.~Huang, Q.~Chen, Y.~Shen, Q.~Chen, and X.~Liu, ``\BIBforeignlanguage{en}{An
  improved variational method for hyperspectral image pansharpening with the
  constraint of spectral difference minimization},''
  \emph{\BIBforeignlanguage{en}{The International Archives of the
  Photogrammetry, Remote Sensing and Spatial Information Sciences}}, vol.
  XLII-2/W7, pp. 753--760, 2017.

\bibitem{Dong20}
W.~Dong, J.~Liang, and S.~Xiao, ``\BIBforeignlanguage{en}{Saliency analysis and
  gaussian mixture model-based detail extraction algorithm for hyperspectral
  pansharpening},'' \emph{\BIBforeignlanguage{en}{IEEE Transactions on
  Geoscience and Remote Sensing}}, vol.~58, no.~8, pp. 5462--5476, 2020.

\bibitem{Zhu2017}
X.~X. Zhu, D.~Tuia, L.~Mou, G.-S. Xia, L.~Zhang, F.~Xu, and F.~Fraundorfer,
  ``Deep learning in remote sensing: A comprehensive review and list of
  resources,'' \emph{IEEE Geoscience and Remote Sensing Magazine}, vol.~5,
  no.~4, pp. 8--36, 2017.

\bibitem{He2019a}
L.~He, J.~Zhu, J.~Li, A.~Plaza, J.~Chanussot, and B.~Li,
  ``\BIBforeignlanguage{en}{{HyperPNN}: Hyperspectral pansharpening via
  spectrally predictive convolutional neural networks},''
  \emph{\BIBforeignlanguage{en}{IEEE Journal of Selected Topics in Applied
  Earth Observations and Remote Sensing}}, vol.~12, no.~8, pp. 3092--3100,
  2019.

\bibitem{He2020}
L.~He, J.~Zhu, J.~Li, D.~Meng, J.~Chanussot, and A.~Plaza,
  ``\BIBforeignlanguage{en}{Spectral-fidelity convolutional neural networks for
  hyperspectral pansharpening},'' \emph{\BIBforeignlanguage{en}{IEEE Journal of
  Selected Topics in Applied Earth Observations and Remote Sensing}}, vol.~13,
  pp. 5898--5914, 2020.

\bibitem{Zheng2020}
Y.~Zheng, J.~Li, Y.~Li, J.~Guo, X.~Wu, and J.~Chanussot, ``Hyperspectral
  pansharpening using deep prior and dual attention residual network,''
  \emph{IEEE Transactions on Geoscience and Remote Sensing}, vol.~58, no.~11,
  pp. 8059--8076, 2020.

\bibitem{He2022}
L.~He, J.~Zhu, J.~Li, A.~Plaza, J.~Chanussot, and Z.~Yu,
  ``\BIBforeignlanguage{en}{{CNN}-based hyperspectral pansharpening with
  arbitrary resolution},'' \emph{\BIBforeignlanguage{en}{IEEE Transactions on
  Geoscience and Remote Sensing}}, vol.~60, pp. 1--21, 2022.

\bibitem{He2022a}
L.~He, J.~Xie, J.~Li, A.~Plaza, J.~Chanussot, and J.~Zhu,
  ``\BIBforeignlanguage{en}{Variable subpixel convolution based
  arbitrary-resolution hyperspectral pansharpening},''
  \emph{\BIBforeignlanguage{en}{IEEE Transactions on Geoscience and Remote
  Sensing}}, vol.~60, pp. 1--19, 2022.

\bibitem{Bandara2022}
W.~G.~C. Bandara, J.~M.~J. Valanarasu, and V.~M. Patel, ``Hyperspectral
  pansharpening based on improved deep image prior and residual
  reconstruction,'' \emph{IEEE Transactions on Geoscience and Remote Sensing},
  vol.~60, pp. 1--16, 2022.

\bibitem{Guan2022}
P.~Guan and E.~Y. Lam, ``\BIBforeignlanguage{en}{Multistage dual-attention
  guided fusion network for hyperspectral pansharpening},''
  \emph{\BIBforeignlanguage{en}{IEEE Transactions on Geoscience and Remote
  Sensing}}, vol.~60, pp. 1--14, 2022.

\bibitem{Wu22}
X.~Wu, J.~Feng, R.~Shang, X.~Zhang, and L.~Jiao,
  ``\BIBforeignlanguage{en}{Multiobjective {Guided} {Divide}-and-{Conquer}
  {Network} for {Hyperspectral} {Pansharpening}},''
  \emph{\BIBforeignlanguage{en}{IEEE Transactions on Geoscience and Remote
  Sensing}}, vol.~60, pp. 1--17, 2022.

\bibitem{Qu22a}
J.~Qu, Y.~Shi, W.~Xie, Y.~Li, X.~Wu, and Q.~Du,
  ``\BIBforeignlanguage{en}{{MSSL}: Hyperspectral and panchromatic images
  fusion via multiresolution spatial--spectral feature learning networks},''
  \emph{\BIBforeignlanguage{en}{IEEE Transactions on Geoscience and Remote
  Sensing}}, vol.~60, pp. 1--13, 2022.

\bibitem{Dong22}
W.~Dong, Y.~Yang, J.~Qu, W.~Xie, and Y.~Li, ``\BIBforeignlanguage{en}{Fusion of
  hyperspectral and panchromatic images using generative adversarial network
  and image segmentation},'' \emph{\BIBforeignlanguage{en}{IEEE Transactions on
  Geoscience and Remote Sensing}}, vol.~60, pp. 1--13, 2022.

\bibitem{Dong22a}
W.~Dong, T.~Zhang, J.~Qu, S.~Xiao, J.~Liang, and Y.~Li,
  ``\BIBforeignlanguage{en}{Laplacian pyramid dense network for hyperspectral
  pansharpening},'' \emph{\BIBforeignlanguage{en}{IEEE Transactions on
  Geoscience and Remote Sensing}}, vol.~60, pp. 1--13, 2022.

\bibitem{Seo2020}
S.~Seo, J.-S. Choi, J.~Lee, H.-H. Kim, D.~Seo, J.~Jeong, and M.~Kim,
  ``{UPSNet}: Unsupervised pan-sharpening network with registration learning
  between panchromatic and multi-spectral images,'' \emph{IEEE Access}, vol.~8,
  pp. 201\,199--201\,217, 2020.

\bibitem{Ciotola2022}
M.~Ciotola, S.~Vitale, A.~Mazza, G.~Poggi, and G.~Scarpa, ``Pansharpening by
  convolutional neural networks in the full resolution framework,'' \emph{IEEE
  Transactions on Geoscience and Remote Sensing}, vol.~60, pp. 1--17, 2022.

\bibitem{Ciotola2023}
M.~Ciotola, G.~Poggi, and G.~Scarpa, ``Unsupervised deep learning-based
  pansharpening with jointly enhanced spectral and spatial fidelity,''
  \emph{IEEE Transactions on Geoscience and Remote Sensing}, vol.~61, pp.
  1--17, 2023.

\bibitem{Nie22}
J.~Nie, Q.~Xu, and J.~Pan, ``\BIBforeignlanguage{en}{Unsupervised hyperspectral
  pansharpening by ratio estimation and residual attention network},''
  \emph{\BIBforeignlanguage{en}{IEEE Geoscience and Remote Sensing Letters}},
  vol.~19, pp. 1--5, 2022.

\bibitem{Guarino2023}
G.~Guarino, M.~Ciotola, G.~Vivone, and G.~Scarpa, ``Band-wise hyperspectral
  image pansharpening using {CNN} model propagation,'' \emph{IEEE Transactions
  on Geoscience and Remote Sensing}, vol.~62, pp. 1--18, 2024.

\bibitem{Guarino2023a}
G.~Guarino, M.~Ciotola, G.~Vivone, G.~Poggi, and G.~Scarpa, ``{PCA}-{CNN}
  hybrid approach for hyperspectral pansharpening,'' \emph{IEEE Geoscience and
  Remote Sensing Letters}, vol.~20, pp. 1--5, 2023.

\bibitem{Thomas2008}
C.~Thomas, T.~Ranchin, L.~Wald, and J.~Chanussot, ``Synthesis of multispectral
  images to high spatial resolution: A critical review of fusion methods based
  on remote sensing physics,'' \emph{IEEE Transactions on Geoscience and Remote
  Sensing}, vol.~46, no.~5, pp. 1301--1312, 2008.

\bibitem{Vivone2019}
G.~Vivone, ``Robust band-dependent spatial-detail approaches for panchromatic
  sharpening,'' \emph{IEEE Transactions on Geoscience and Remote Sensing},
  vol.~57, no.~9, pp. 6421--6433, 2019.

\bibitem{Vivone2018a}
G.~{Vivone}, R.~{Restaino}, and J.~{Chanussot}, ``Full scale regression-based
  injection coefficients for panchromatic sharpening,'' \emph{IEEE Transactions
  on Image Processing}, vol.~27, no.~7, pp. 3418--3431, 2018.

\bibitem{Alparone2017}
L.~{Alparone}, A.~{Garzelli}, and G.~{Vivone}, ``Intersensor statistical
  matching for pansharpening: Theoretical issues and practical solutions,''
  \emph{IEEE Transactions on Geoscience and Remote Sensing}, vol.~55, no.~8,
  pp. 4682--4695, 2017.

\bibitem{Vivone2017}
G.~Vivone, R.~Restaino, and J.~Chanussot, ``A regression-based high-pass
  modulation pansharpening approach,'' \emph{IEEE Transactions on Geoscience
  and Remote Sensing}, vol.~56, no.~2, pp. 984--996, 2017.

\bibitem{Vicinanza2015}
M.~R. Vicinanza, R.~Restaino, G.~Vivone, M.~D. Mura, and J.~Chanussot, ``A
  pansharpening method based on the sparse representation of injected
  details,'' \emph{IEEE Geoscience and Remote Sensing Letters}, vol.~12, no.~1,
  pp. 180--184, 2015.

\bibitem{Zhuo2022}
Y.-W. Zhuo, T.-J. Zhang, J.-F. Hu, H.-X. Dou, T.-Z. Huang, and L.-J. Deng, ``A
  deep-shallow fusion network with multidetail extractor and spectral attention
  for hyperspectral pansharpening,'' \emph{IEEE Journal of Selected Topics in
  Applied Earth Observations and Remote Sensing}, vol.~15, pp. 7539--7555,
  2022.

\bibitem{Wald1997}
L.~Wald, T.~Ranchin, and M.~Mangolini, ``Fusion of satellite images of
  different spatial resolutions: Assessing the quality of resulting images,''
  \emph{Photogrammetric Engineering and Remote Sensing}, vol.~63, no.~6, pp.
  691--699, 1997.

\bibitem{Scarpa2022}
\BIBentryALTinterwordspacing
G.~Scarpa and M.~Ciotola, ``Full-resolution quality assessment for
  pansharpening,'' \emph{Remote Sensing}, vol.~14, no.~8, 2022. [Online].
  Available: \url{https://www.mdpi.com/2072-4292/14/8/1808}
\BIBentrySTDinterwordspacing

\bibitem{Wald2002}
L.~Wald, \emph{Data Fusion: Definitions and Architectures --- Fusion of images
  of different spatial resolutions}.\hskip 1em plus 0.5em minus 0.4em\relax
  Paris, France: Les Presses de l'\'Ecole des Mines, 2002.

\bibitem{Yuhas1992}
R.~H. Yuhas, A.~F.~H. Goetz, and J.~W. Boardman, ``Discrimination among
  semi-arid landscape endmembers using the {S}pectral {A}ngle {M}apper ({SAM})
  algorithm,'' in \emph{Proc. Summaries 3rd Annu. JPL Airborne Geosci.
  Workshop}, 1992, pp. 147--149.

\bibitem{Garzelli2009}
A.~Garzelli and F.~Nencini, ``Hypercomplex quality assessment of
  multi/hyperspectral images,'' \emph{IEEE Transactions on Geoscience and
  Remote Sensing}, vol.~6, no.~4, pp. 662--665, 2009.

\bibitem{Wang2002}
Z.~Wang and A.~C. Bovik, ``A universal image quality index,'' \emph{IEEE Signal
  Processing Letters}, vol.~9, no.~3, pp. 81--84, 2002.

\bibitem{Arienzo2022}
A.~Arienzo, G.~Vivone, A.~Garzelli, L.~Alparone, and J.~Chanussot, ``Full
  resolution quality assessment of pansharpening: Theoretical and hands-on
  approaches,'' \emph{IEEE Geoscience and Remote Sensing Magazine}, 2022.

\bibitem{Khan2009}
M.~M. Khan, L.~Alparone, and J.~Chanussot, ``Pansharpening quality assessment
  using the modulation transfer functions of instruments,'' \emph{IEEE
  Transactions on Geoscience and Remote Sensing}, vol.~47, no.~11, pp.
  3880--3891, 2009.

\bibitem{Alparone2018}
L.~Alparone, A.~Garzelli, and G.~Vivone, ``Spatial consistency for full-scale
  assessment of pansharpening,'' in \emph{IEEE International Geoscience and
  Remote Sensing Symposium IGARSS}.\hskip 1em plus 0.5em minus 0.4em\relax
  IEEE, 2018, pp. 5132--5134.

\end{thebibliography}
\end{document}